\documentclass[10pt,twocolumn,letterpaper]{article}

\usepackage[pagenumbers]{cvpr} %

\usepackage{graphicx}
\usepackage{amsmath}
\usepackage{amssymb}
\usepackage{booktabs}

\usepackage{algorithm}
\usepackage{algorithmic}
\usepackage{xspace}
\usepackage{textcomp}
\usepackage[export]{adjustbox}
\usepackage{rotating}
\usepackage{wrapfig}
\usepackage{float}
\usepackage{array}
\newcolumntype{P}[1]{>{\centering\arraybackslash}p{#1}}
\usepackage{bm}
\usepackage[accsupp]{axessibility} %

\newlength{\ww}

\usepackage[pagebackref,breaklinks,colorlinks]{hyperref}

\usepackage[capitalize]{cleveref}
\crefname{section}{Sec.}{Secs.}
\Crefname{section}{Section}{Sections}
\Crefname{table}{Table}{Tables}
\crefname{table}{Tab.}{Tabs.}

\def\cvprPaperID{5345} %
\def\confName{CVPR}
\def\confYear{2022}

\begin{document}

\title{Blended Diffusion for Text-driven Editing of Natural Images}

\author
{
    Omri Avrahami$^{1}$ 
    \hspace{6mm} Dani Lischinski$^{1}$ 
    \hspace{6mm} Ohad Fried$^{2}$
    \\
    $^{1}$The Hebrew University of Jerusalem  
    \hspace{10mm} $^{2}$Reichman University 
}

\maketitle

\def\ShowNotes{}
\newcommand{\ignorethis}[1]{}
\newcommand{\redund}[1]{#1}

\newcommand{\apriori    }     {\textit{a~priori}}
\newcommand{\aposteriori}     {\textit{a~posteriori}}
\newcommand{\perse      }     {\textit{per~se}}
\newcommand{\naive      }     {{na\"{\i}ve}}
\newcommand{\Naive      }     {{Na\"{\i}ve}}
\newcommand{\Identity   }     {\mat{I}}
\newcommand{\Zero       }     {\mathbf{0}}
\newcommand{\Reals      }     {{\textrm{I\kern-0.18em R}}}
\newcommand{\isdefined  }     {\mbox{\hspace{0.5ex}:=\hspace{0.5ex}}}
\newcommand{\texthalf   }     {\ensuremath{\textstyle\frac{1}{2}}}
\newcommand{\half       }     {\ensuremath{\frac{1}{2}}}
\newcommand{\third      }     {\ensuremath{\frac{1}{3}}}
\newcommand{\fourth     }     {\ensuremath{\frac{1}{4}}}

\newcommand{\Lone} {\ensuremath{L_1}}
\newcommand{\Ltwo} {\ensuremath{L_2}}

\newcommand{\degree} {\ensuremath{^{\circ}}}

\newcommand{\mat        } [1] {{\text{\boldmath $\mathbit{#1}$}}}
\newcommand{\Approx     } [1] {\widetilde{#1}}
\newcommand{\change     } [1] {\mbox{{\footnotesize $\Delta$} \kern-3pt}#1}

\newcommand{\Order      } [1] {O(#1)}
\newcommand{\set        } [1] {{\lbrace #1 \rbrace}}
\newcommand{\floor      } [1] {{\lfloor #1 \rfloor}}
\newcommand{\ceil       } [1] {{\lceil  #1 \rceil }}
\newcommand{\inverse    } [1] {{#1}^{-1}}
\newcommand{\transpose  } [1] {{#1}^\mathrm{T}}
\newcommand{\invtransp  } [1] {{#1}^{-\mathrm{T}}}
\newcommand{\relu       } [1] {{\lbrack #1 \rbrack_+}}

\newcommand{\abs        } [1] {{| #1 |}}
\newcommand{\Abs        } [1] {{\left| #1 \right|}}
\newcommand{\norm       } [1] {{\| #1 \|}}
\newcommand{\Norm       } [1] {{\left\| #1 \right\|}}
\newcommand{\pnorm      } [2] {\norm{#1}_{#2}}
\newcommand{\Pnorm      } [2] {\Norm{#1}_{#2}}
\newcommand{\inner      } [2] {{\langle {#1} \, | \, {#2} \rangle}}
\newcommand{\Inner      } [2] {{\left\langle \begin{array}{@{}c|c@{}}
                               \displaystyle {#1} & \displaystyle {#2}
                               \end{array} \right\rangle}}

\newcommand{\twopartdef}[4]
{
  \left\{
  \begin{array}{ll}
    #1 & \mbox{if } #2 \\
    #3 & \mbox{if } #4
  \end{array}
  \right.
}

\newcommand{\fourpartdef}[8]
{
  \left\{
  \begin{array}{ll}
    #1 & \mbox{if } #2 \\
    #3 & \mbox{if } #4 \\
    #5 & \mbox{if } #6 \\
    #7 & \mbox{if } #8
  \end{array}
  \right.
}

\newcommand{\len}[1]{\text{len}(#1)}

\newlength{\w}
\newlength{\h}
\newlength{\x}

\definecolor{darkred}{rgb}{0.7,0.1,0.1}
\definecolor{darkgreen}{rgb}{0.1,0.6,0.1}
\definecolor{cyan}{rgb}{0.7,0.0,0.7}
\definecolor{otherblue}{rgb}{0.1,0.4,0.8}
\definecolor{maroon}{rgb}{0.76,.13,.28}
\definecolor{burntorange}{rgb}{0.81,.33,0}

\ifdefined\ShowNotes
  \newcommand{\colornote}[3]{{\color{#1}\textbf{#2} #3\normalfont}}
\else
  \newcommand{\colornote}[3]{}
\fi

\newcommand {\note}[1]{\colornote{maroon}{}{#1}}
\newcommand {\todo}[1]{\colornote{cyan}{TODO}{#1}}
\newcommand {\ohad}[1]{\colornote{otherblue}{OF:}{#1}}
\newcommand {\dani}[1]{\colornote{darkgreen}{DL:}{#1}}
\newcommand {\omri}[1]{\colornote{burntorange}{OA:}{#1}}

\newcommand {\reqs}[1]{\colornote{red}{\tiny #1}}

\newcommand {\new}[1]{\colornote{red}{#1}}

\newcommand*\rot[1]{\rotatebox{90}{#1}}

\newcommand {\newstuff}[1]{#1}

\newcommand\todosilent[1]{}

\newcommand{\woBGmask}{{w/o~bg~\&~mask}}
\newcommand{\woMask}{{w/o~mask}}

\providecommand{\keywords}[1]
{
  \textbf{\textit{Keywords---}} #1
}

\newcommand {\shortcite}[1]{\cite{#1}}

\newcommand{\GAN}{\textit{GAN}}
\newcommand{\data}{\mathit{data}}
\newcommand{\unionGAN}{\textsc{UnionGAN}\xspace}
\newcommand {\ganArrow}[2]{\ensuremath{\GAN_{{#1} \rightarrow {#2}}}}
\newcommand {\gan}[1]{\ensuremath{\GAN_{#1}}}
\newcommand{\mclip}{\mathit{CLIP}}

\begin{abstract}
Natural language offers a highly intuitive interface for image editing. In this paper, we introduce the first solution for performing local (region-based) edits in generic natural images, based on a natural language description along with an ROI mask.
We achieve our goal by leveraging and combining a pretrained language-image model (CLIP), to steer the edit towards a user-provided text prompt, with a denoising diffusion probabilistic model (DDPM) to generate natural-looking results.
To seamlessly fuse the edited region with the unchanged parts of the image, we spatially blend noised versions of the input image with the local text-guided diffusion latent at a progression of noise levels.
In addition, we show that adding augmentations to the diffusion process mitigates adversarial results.
We compare against several baselines and related methods, both qualitatively and quantitatively, and show that our method outperforms these solutions in terms of overall realism, ability to preserve the background and matching the text. Finally, we show several text-driven editing applications, including adding a new object to an image, removing/replacing/altering existing objects, background replacement, and image extrapolation.
\end{abstract}
\section{Introduction}
\label{sec:intro}

It is said that ``a picture is worth a thousand words'', but recent research indicates that only a few words are often sufficient to describe one.
Recent works that leverage the tremendous progress in vision-language models and data-driven image generation have demonstrated that text-based interfaces for image creation and manipulation are now finally within reach \cite{ramesh2021zero, ding2021cogview, xia2021towards, tao2020df, li2019controllable, zhang2018photographic, qiao2019mirrorgan, li2019object, qiao2019learn, hinz2019semantic}.

The most impressive results in text-driven image manipulation leverage the strong generative capabilities of modern GANs \cite{goodfellow2014generative, karras2019style, karras2020analyzing, karras2021alias, brock2018large}.
However, GAN-based approaches are typically limited to images from a restricted domain, on which the GAN was trained. Furthermore, in order to manipulate real images, they must be first \emph{inverted} into the GAN's latent space. Although many GAN inversion techniques have recently emerged \cite{xia2021gan,abdal2019image2stylegan, zhu2020domain, abdal2020image2stylegan++, richardson2021encoding, alaluf2021restyle, tov2021designing}, it was also shown that there is a trade-off between the reconstruction accuracy and the editability of the inverted images \cite{tov2021designing}.
Restricting the image manipulation to a specific region in the image is another challenge for existing approaches \cite{bau2021paint}.

\begin{figure}[t]
	\centering
	\setlength{\tabcolsep}{0.5pt}
	\renewcommand{\arraystretch}{0.6}
	\setlength{\ww}{0.24\columnwidth}
	
	\begin{tabular}{cccc}
        \includegraphics[width=\ww,frame]{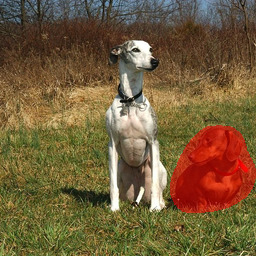} &
		\includegraphics[width=\ww,frame]{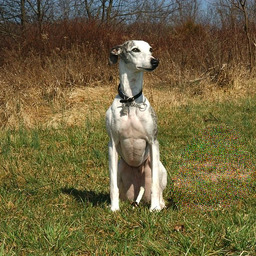} &
		\includegraphics[width=\ww,frame]{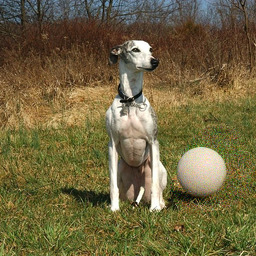} &
		\includegraphics[width=\ww,frame]{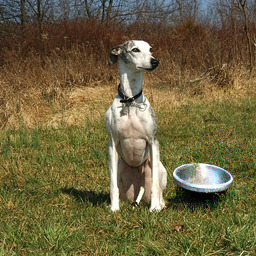} \\
		\small{input+mask} & \small{no prompt} & \small{``white ball''} & \small{``bowl of water''} \\
		& & & \\
		
		\includegraphics[width=\ww,frame]{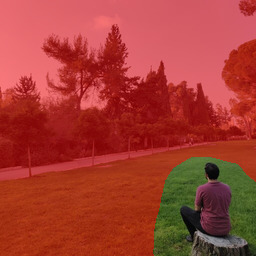} &
		\includegraphics[width=\ww,frame]{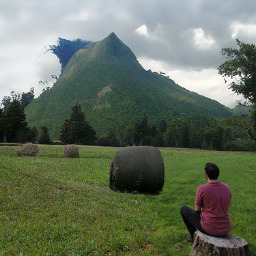} &
		\includegraphics[width=\ww,frame]{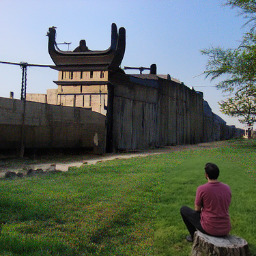} &
		\includegraphics[width=\ww,frame]{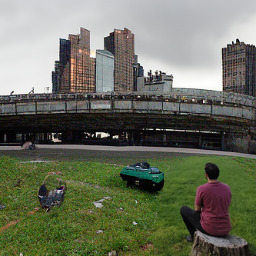} \\
		\small{input+mask} & \small{``big mountain''} & \small{``big wall''}  & \footnotesize{``New York City''}\\
	\end{tabular}
	\caption{\textbf{Text-driven object/background replacement:} Given an input image and a mask, we modify the masked area according to a guiding text prompt, without affecting the unmasked regions.}
	\label{fig:teaser}
	\vspace{-4mm}
\end{figure}

In this work, we present the first approach for region-based editing of \emph{generic} real-world natural images, using natural language text guidance\footnote{Code is available at: \url{https://omriavrahami.com/blended-diffusion-page/}}. Specifically, we aim at a text-driven method that (1) can operate on real images, rather than generated ones, (2) is not restricted to a specific domain, such as human faces or bedrooms, (3) modifies only a user-specified region, while preserving the rest of the image, (4) yields globally coherent (seamless) editing results, and (5) capable of generating multiple results for the same input, because of the one-to-many nature of the task. Several examples of such edits are shown in \Cref{fig:teaser}.

The demanding image editing scenario described above has not received much attention in the deep-learning era.
In fact, the most closely related works are classical approaches, such as seamless cloning \cite{perez2003poisson,farbman2009coords} and image completion \cite{hays2007completion}, none of which are text-driven.
A more recent related work is zero-shot semantic image painting \cite{bau2021paint} in which arbitrary simple textual descriptions can be attributed to the desired location within an image. However, this method does not operate on real images (requirement 1), does not preserve the background of the image (requirement 3), and does not generate multiple outputs for the same input (requirement 5).

To achieve our goals, we utilize two off-the-shelf pre-trained models: Denoising Diffusion Probabilistic Models (DDPM)~\cite{ho2020denoising, nichol2021improved, dhariwal2021diffusion} and Contrastive Language-Image Pre-training (CLIP) \cite{radford2021learning}. 
DDPM is a class of probabilistic generative models that has recently been shown to surpass the image generation quality of state-of-the-art GANs~\cite{dhariwal2021diffusion}. We use DPPM as our generative backbone in order to ensure natural-looking results.
The CLIP model is contrastively trained on a dataset of 400 million (image, text) pairs collected from the internet to learn a rich shared embedding space for images and text. We use CLIP in order to guide the manipulation to match the user-provided text prompt. 

We show that a \naive{} combination of DDPM and CLIP to perform text-driven local editing fails to preserve the image background, and in many cases, leads to a less natural result.
Instead, we propose a novel way to leverage the diffusion process, which blends the CLIP-guided diffusion latents with \emph{suitably noised versions of the input image}, at each diffusion step. We show that this scheme produces natural-looking results that are coherent with the unaltered parts of the input. We further show that using \emph{extending augmentations} at each step of the diffusion process reduces adversarial results. Our method utilizes pretrained DDPM and CLIP models, without requiring additional training.

In summary, our main contributions are: (1) We propose the first solution for general-purpose region-based image editing, using natural language guidance, applicable to real, diverse images. (2) Our background preservation technique guarantees that unaltered regions are perfectly preserved. (3) We demonstrate that a simple augmentation technique significantly reduces the risk of adversarial results, allowing us to use gradient-based diffusion guidance.
\section{Related Work}
\label{sec:related_work}

\paragraph{Text-to-image synthesis.} Recently, we've witnessed significant advances in text-to-image generation. Initial RNN-based works \cite{Mansimov2016GeneratingIF} were quickly superseded by generative adversarial approaches, such as the seminal work by Reed et al.~\cite{reed2016generative}. The latter was further improved by multi-stage architectures \cite{zhang2017stackgan, zhang2018stackgan++} and an attention mechanism \cite{xu2018attngan}.

DALL-E~\cite{ramesh2021zero} introduced a GAN-free two stage approach: first, a discrete VAE \cite{Oord2017NeuralDR, razavi2019generating} is trained to reduce the context for the transformer. Next, a transformer \cite{vaswani2017attention} is trained autoregressively to model the joint distribution over the text and image tokens.

Several recent projects~\cite{big_sleep, vqgan_clip, clip_guided_diffusion} utilize a pretrained generative model \cite{brock2018large, esser2021taming, dhariwal2021diffusion} using a pretrained CLIP model \cite{radford2021learning} to steer the generated result towards the desired target description. These methods are mainly used to create abstract artworks from text descriptions and lack the ability to edit parts of a real image, while preserving the rest.

While text-to-image is a challenging and interesting task, in this work we focus on text-driven image manipulation, where edits are restricted to a user-specified region.

\vspace{-3mm}
\paragraph{Text-driven image manipulation.} Several recent works utilize CLIP in order to manipulate real images. StyleCLIP \cite{patashnik2021styleclip} use pretrained StyleGAN2 \cite{karras2020analyzing} and CLIP models
to modify images based on text prompts. %
To manipulate real images (rather than generated ones), they must first be encoded to the latent space \cite{tov2021designing}. This approach cannot handle generic real images, and is restricted to domains for which high-quality generators are available.
In addition, StyleCLIP operates on images in a \emph{global} fashion, without providing spatial control over which areas should change.

More closely related to ours is the work of Bau et al.~\cite{bau2021paint}, %
where arbitrary simple textual descriptions can be attributed to a desired location within an image. Their GAN-based approach has several limitations: (1) although they attempt to preserve the background, it may still change, as can be seen in \Cref{fig:comparison_paint_by_word_original}; (2) their solution is mainly demonstrated in the restricted domain of bedrooms, and mainly for color and texture editing tasks. A few examples of general images are shown, but the results are less natural or lack background preservation (see \Cref{fig:comparison_paint_by_word_original}). (3) Their model is able to operate only on generated images and is not applicable out-of-the-box to arbitrary natural images. GAN-inversion techniques \cite{xia2021gan,abdal2019image2stylegan, zhu2020domain, abdal2020image2stylegan++, richardson2021encoding, alaluf2021restyle, tov2021designing} can be used to edit real images, but it was shown \cite{tov2021designing} that there is a trade-off between the edibility and the distortion of the reconstructed image.

Concurrently with our work, Liu et al.~\cite{liu2021more} and Kim et al.~\cite{kim2021diffusionclip} propose ways to utilize a diffusion model in order to perform \emph{global} text-guided image manipulations. In addition, GLIDE~\cite{nichol2021glide} is a concurrent work that utilizes the diffusion model for text-to-image synthesis, as well as local image editing using text guidance. In order to do so, they train a designated diffusion model for these tasks.

\begin{figure}[t]
    \centering
    \setlength{\ww}{\columnwidth}
    
    \includegraphics[width=\ww]{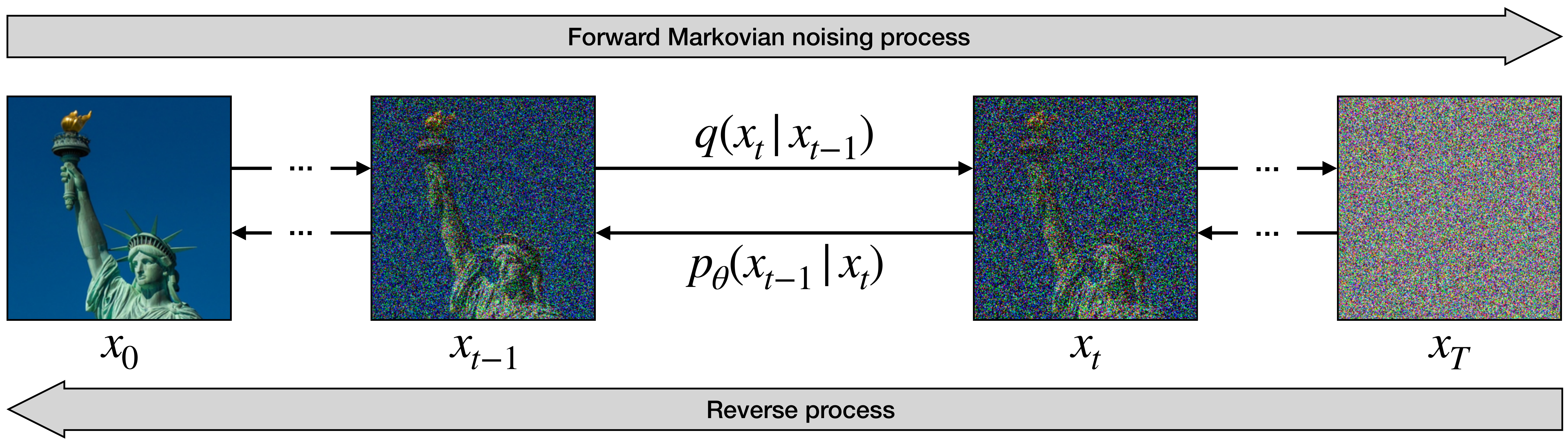}
    
    \caption{\textbf{Denoising diffusion.} Starting from a sample $x_0$, a forward Markovian noising process produces a series of noisy images by gradually adding Gaussian noise $q(x_t | x_{t-1})$, until obtaining a nearly isotropic Gaussian noise sample $x_T$. The reverse process transforms a Gaussian noise sample $x_T$ into $x_0$ by repeated denoising using a learned posterior $p_{\theta}(x_{t-1}|x_t)$.}
    \label{fig:diffusion_process_illustration_classic}
    \vspace{-4mm}
\end{figure}

\section{Denoising Diffusion Probabilistic Models}
\label{sec:preliminaries}

\newcommand{\train}{q(x_0)}

Denoising diffusion probabilistic models (DDPMs) learn to invert a parameterized Markovian image noising process. Starting from isotropic Gaussian noise samples, they transform them to samples from a training distribution, gradually removing the noise by an iterative diffusion process (\cref{fig:diffusion_process_illustration_classic}). DDPMs have recently been shown to generate high-quality images \cite{ho2020denoising, nichol2021improved, dhariwal2021diffusion}.
Below, we provide a brief overview of DDPMs, for more details please refer to \cite{sohl2015deep, ho2020denoising, nichol2021improved}. We follow the formulations and notations in \cite{nichol2021improved}.

Given a data distribution $x_0 \sim \train$, a forward noising process produces a series of latents $x_1, ..., x_T$ by adding Gaussian noise with variance $\beta_t \in (0,1)$ at time $t$:
\begin{align}
\label{eqn:forward_process}
    \begin{split}
         q(x_1, ..., x_T | x_0) &= \prod_{t=1}^{T} q(x_t | x_{t-1})
        \\
         q(x_t | x_{t-1}) &= \mathcal{N}(\sqrt{1-\beta_t} x_{t-1}, \beta_t \mathbf{I})
    \end{split}
\end{align}
When $T$ is large enough, the last latent $x_T$ is nearly an isotropic Gaussian distribution.

An important property of the forward noising process is that any step $x_t$ may be sampled directly from $x_0$, without the need to generate the intermediate steps,
\begin{align}
\label{eqn:fast_forward}
    \begin{split}
         q(x_t|x_0) &= \mathcal{N}(\sqrt{\bar{\alpha}_t} x_0, (1-\bar{\alpha}_t) \mathbf{I})
        \\
         x_t &=  \sqrt{\bar{\alpha}_t} x_0 + \sqrt{1-\bar{\alpha}_t} \epsilon,
    \end{split}
\end{align}
where $\epsilon \sim \mathcal{N}(0,\mathbf{I})$, $\alpha_t = 1 - \beta_t$ and $\bar{\alpha}_t = \prod_{s=0}^{t} \alpha_s$.

To draw a new sample from the distribution $\train$ the Markovian process is reversed. That is, starting from a Gaussian noise sample, $x_T \sim \mathcal{N}(0, \mathbf{I})$, a reverse sequence is generated by sampling the posteriors $q(x_{t-1}|x_t)$, which were shown to also be Gaussian distributions \cite{feller1949theory, sohl2015deep}.

However, $q(x_{t-1}|x_t)$ is unknown, as it depends on the unknown data distribution $\train$. In order to approximate this function, a deep neural network $p_\theta$ is trained to predict the mean and the covariance of $x_{t-1}$ given $x_t$ as input. Then $x_{t-1}$ may be sampled from the normal distribution defined by these parameters,
\begin{equation}
    p_{\theta}(x_{t-1}|x_t) = \mathcal{N}(\mu_{\theta}(x_t, t), \Sigma_{\theta}(x_t, t)).
\end{equation}

Rather than inferring $\mu_{\theta}(x_t, t)$ directly, Ho et al.~\cite{ho2020denoising} propose to predict the noise $\epsilon_{\theta}(x_t, t)$ that was added to $x_0$ in order to obtain $x_t$, according to \Cref{eqn:fast_forward}. Then $\mu_{\theta}(x_t, t)$ may be derived using Bayes' theorem:
\begin{equation}
\mu_{\theta}(x_t, t) = \frac{1}{\sqrt{\alpha_t}} \left( x_t - \frac{\beta_t}{\sqrt{1-\bar{\alpha}_t}} \epsilon_{\theta}(x_t, t) \right)
\end{equation}
For more details please see \cite{ho2020denoising}.

Ho et al.~\cite{ho2020denoising} kept $\Sigma_{\theta}(x_t, t)$ constant, but it was later shown \cite{nichol2021improved} that it is better to learn it by a neural network that interpolates between the upper and lower bounds for the fixed covariance proposed by Ho et al.

Dhariwal and Nichol~\cite{dhariwal2021diffusion} show that diffusion models can achieve image sample quality superior to the current state-of-the-art generative models. They improved the results of \cite{ho2020denoising}, in terms of FID score \cite{heusel2017gans}, by tuning the network architecture and by incorporating guidance using a classifier pretrained on noisy images. For more details please see the supplementary and the original paper \cite{dhariwal2021diffusion}.
\newcommand{\Ourmethod}{Text-driven blended diffusion\xspace}
\newcommand{\ourmethod}{text-driven blended diffusion\xspace}
\newcommand{\Our}{TBD\xspace}

\section{Method}
\label{sec:method}

Given an image $x$, a guiding text prompt $d$ and a binary mask $m$ that marks the region of interest in the image, our goal is to produce a modified image $\widehat{x}$, s.t.~the content $\widehat{x} \odot m$ is consistent with the text description $d$, while the complementary area remains as close as possible to the source image, i.e., $x \odot (1 - m) \approx \widehat{x} \odot (1 - m)$, where $\odot$ is element-wise multiplication. Furthermore, the transition between the two areas of $\widehat{x}$ should ideally appear seamless.

In \Cref{sec:local-clip} we start by adapting the DDPM approach described above to incorporate local text-driven editing by adding a guiding loss comprised of a masked CLIP loss and a background preservation term. The resulting method still falls short of satisfying our requirements, and we proceed to present a new \ourmethod method in \Cref{sec:tbd}, which guarantees background preservation and improves the coherence of the edited result.
\Cref{sec:mitigating_adversarial_examples} introduces an augmentation technique that we employ in order to avoid adversarial results. 

\subsection{Local CLIP-guided diffusion}
\label{sec:local-clip}

Dhariwal and Nichol \cite{dhariwal2021diffusion} use a classifier pretrained on noisy images to guide generation towards a target class. Similarly, a pretrained CLIP model may be used to guide diffusion towards 
a target prompt. Since CLIP is trained on clean images (and retraining it on noisy images is impractical), we need a way of estimating a clean image $x_0$ from each noisy latent $x_t$ during the denoising diffusion process.
Recall that the process estimates at each step the noise $\epsilon_{\theta}(x_t, t)$ that was added to $x_0$ to obtain $x_t$. Thus, $x_0$ may be obtained from $\epsilon_{\theta}(x_t, t)$ via \Cref{eqn:fast_forward}:
\begin{equation}
\label{eqn:clean_image_prediciton}
    \widehat{x}_0 = \frac{x_t}{\sqrt{\bar{\alpha}_t}} - \frac{\sqrt{1-\bar{\alpha}_t} \epsilon_{\theta}(x_t, t)}{\sqrt{\bar{\alpha}_t}}
\end{equation}

Now, a CLIP-based loss $\mathcal{D}_{\mclip}$ may be defined as the cosine distance between the CLIP embedding of the text prompt and the embedding of the estimated clean image $\widehat{x}_0$:
\begin{equation}
    \label{eqn:d_clip}
    \mathcal{D}_{\mclip}(x, d, m) = ~D_c(\mclip_{\mathit{img}}(x \odot m), \mclip_{\mathit{txt}}(d))
\end{equation}
where $D_c$ denotes cosine distance. A similar approach is used in CLIP-guided diffusion \cite{clip_guided_diffusion}, where a linear combination of $x_t$ and $\widehat{x}_0$ is used to provide global guidance for the diffusion. The guidance can be made local, by considering only the gradients of $\mathcal{D}_{\mclip}$ under the input mask. In this manner, we effectively adapt CLIP-guided diffusion \cite{clip_guided_diffusion} to the local (region-based) editing setting.

The above process starts from an isotropic Gaussian noise and has no background constraints. Thus, although $\mathcal{D}_{\mclip}$ is evaluated inside the masked region, it affects the entire image. In order to steer the surrounding region towards the input image, a background preservation loss $\mathcal{D}_{bg}$ is added to guide the diffusion outside the mask: 
\begin{align}
\label{eqn:loss_background_preservation}
    \begin{split}
         \mathcal{D}_{bg}(x_1, x_2, m) = d(x_1 \odot (1-m), x_2 \odot (1-m))
        \\
         d(x_1, x_2) = \frac{1}{2}(\textit{MSE}(x_1, x_2) + \textit{LPIPS}(x_1, x_2))
    \end{split}
\end{align}
where $\textit{MSE}$ is the $L_2$ norm of the pixel-wise difference between the images, and $\textit{LPIPS}$ is the Learned Perceptual Image Patch Similarity metric \cite{zhang2018unreasonable}.

The diffusion guidance loss is thus set to the weighted sum $\mathcal{D}_{\mclip}(\widehat{x}_0, d, m) + \lambda\mathcal{D}_{bg}(x, \widehat{x}_0, m)$, and the resulting method is summarized in \Cref{alg:local_clip_guided_diffusion}.

\begin{algorithm}[t]
    \footnotesize
    \caption{\footnotesize Local CLIP-guided diffusion, given a diffusion model $(\mu_{\theta}(x_t), \Sigma_{\theta}(x_t))$ and $\mclip$ model}
    \label{alg:local_clip_guided_diffusion}
    \begin{algorithmic}
        \STATE \textbf{Input:} source image $x$, target text description $d$, input mask $m$, diffusion steps $k$, background preservation coefficient $\lambda$
        \STATE \textbf{Output:} edited image $\widehat{x}$ that differs from input image $x$ inside area $m$ according to text description $d$ 
        \STATE $x_k \sim \mathcal{N}(\sqrt{\bar{\alpha}_k} x_0, (1-\bar{\alpha}_k) \mathbf{I})$
        \FORALL{$t$ from $k$ to 1}
            \STATE $\mu, \Sigma \gets \mu_{\theta}(x_t), \Sigma_{\theta}(x_t)$
            \STATE $\widehat{x}_0 \gets \frac{x_t}{\sqrt{\bar{\alpha}_t}} - \frac{\sqrt{1-\bar{\alpha}_t} \epsilon_{\theta}(x_t, t)}{\sqrt{\bar{\alpha}_t}}$
            \STATE $\widehat{x}_{0,{\textit{aug}}} \gets {\textit{ExtendingAugmentations}(\widehat{x}_0, N)}$
            \STATE $\mathcal{L} \gets \mathcal{D}_{\textit{CLIP}}(\widehat{x}_{0,{\textit{aug}}}, d, m) + \lambda \mathcal{D}_{\textit{bg}}(x, \widehat{x}_{0,{\textit{aug}}}, m)$
            \STATE $x_{t-1} \sim \mathcal{N}(\mu + \Sigma \nabla_{\widehat{x}_0}{\mathcal{L}}, \Sigma)$
        \ENDFOR
        \RETURN $x_0$
    \end{algorithmic}
\end{algorithm}
\begin{figure}[t]
    \centering
    \setlength{\tabcolsep}{0.5pt}
    \renewcommand{\arraystretch}{0.5}
    \setlength{\ww}{0.24\columnwidth}
  
    \begin{tabular}{cccc}
        \includegraphics[width=\ww,frame]{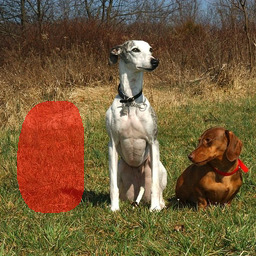} &
        \includegraphics[width=\ww,frame]{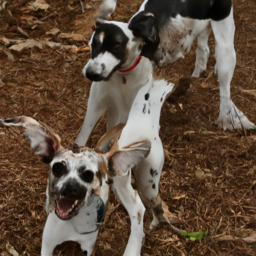} &
        \includegraphics[width=\ww,frame]{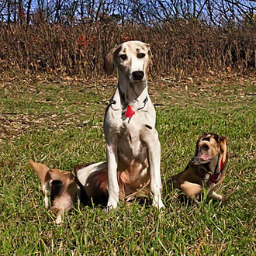} &
        \includegraphics[width=\ww,frame]{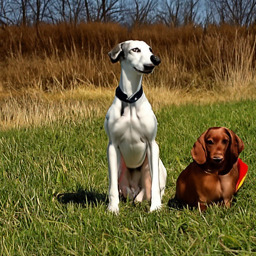} \\
        
        \scriptsize{Input + mask} & 
        \scriptsize{$\lambda=100$} & 
        \scriptsize{$\lambda=1000$} & 
        \scriptsize{$\lambda=10000$} \\
    \end{tabular}
    
    \caption{\textbf{Effect of $\lambda$ in local CLIP-guided diffusion.} Given an input image with a mask, and the prompt ``a dog'': with $\lambda$ set too low ($\lambda=100$), the entire image changes completely, while if $\lambda$ is too high ($\lambda=10000$), the model fails to change the foreground (and the background preservation is not perfect). Using an intermediate value ($\lambda=1000$) the model changes the foreground while resembling the original background (zoom for more details).}
    \label{fig:local_clip_guided_diffusion_lambda}
    \vspace{-4mm}
\end{figure}

In practice, we have found that an inherent trade-off exists between the two guidance terms above,
as demonstrated in \Cref{fig:local_clip_guided_diffusion_lambda}. Note that even in the intermediate case of $\lambda = 1000$ the result is far from perfect: the background is only roughly preserved and the foreground is severely limited. We overcome this issue in the next section.

\begin{algorithm}[t]
    \footnotesize
    \caption{\footnotesize \Ourmethod: given a diffusion model $(\mu_{\theta}(x_t), \Sigma_{\theta}(x_t))$, and $\mclip$ model}
    \label{alg:final}
    \begin{algorithmic}
        \STATE \textbf{Input:} source image $x$, target text description $d$, input mask $m$, diffusion steps $k$, number of extending augmentations $N$
        \STATE \textbf{Output:} edited image $\widehat{x}$ that differs from input image $x$ inside area $m$ according to text description $d$ 
        \STATE $x_k \sim \mathcal{N}(\sqrt{\bar{\alpha}_k} x_0, (1-\bar{\alpha}_k) \mathbf{I})$
        \FORALL{$t$ from $k$ to 0}
            \STATE $\mu, \Sigma \gets \mu_{\theta}(x_t), \Sigma_{\theta}(x_t)$
            \STATE $\widehat{x}_0 \gets \frac{x_t}{\sqrt{\bar{\alpha}_t}} - \frac{\sqrt{1-\bar{\alpha}_t} \epsilon_{\theta}(x_t, t)}{\sqrt{\bar{\alpha}_t}}$
            \STATE $\widehat{x}_{0,{\textit{aug}}} \gets {\textit{ExtendingAugmentations}(\widehat{x}_0, N)}$
            \STATE $\nabla_{\textit{text}} \gets \frac{1}{N} \sum_{i=1}^{N} \nabla_{\widehat{x}_{0,{\textit{aug}}}} \mathcal{D}_{\textit{CLIP}}(\widehat{x}_{0,{\textit{aug}}}, d, m)$
            \STATE $x_{\textit{fg}} \sim \mathcal{N}(\mu + \Sigma \nabla_{\textit{text}}, \Sigma)$
            \STATE $x_{\textit{bg}} \sim \mathcal{N}(\sqrt{\bar{\alpha}_t} x_0, (1-\bar{\alpha}_t) \mathbf{I})$
            \STATE $x_{t-1} \gets x_{\textit{fg}} \odot m + x_{\textit{bg}} \odot (1 - m)$
        \ENDFOR
        \RETURN $x_{-1}$
    \end{algorithmic}
\end{algorithm}
\begin{figure}[t]
    \centering
    \setlength{\ww}{\columnwidth}
    
    \includegraphics[width=\ww]{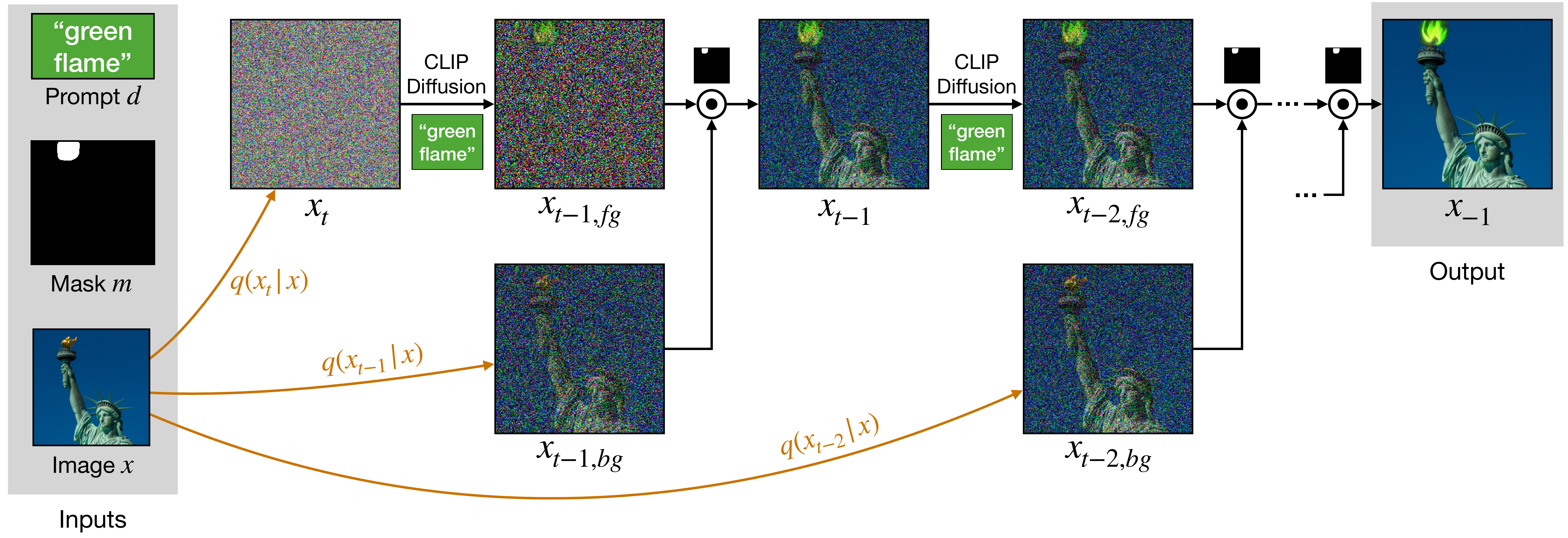}
    
    \caption{\textbf{\Ourmethod.} Given input image $x$, input mask $m$, and a text prompt $d$, we leverage the diffusion process to edit the image locally and coherently. We denote with $\odot$ the element-wise blending of two images using the input mask $m$.}
    \label{fig:diffusion_process_illustration_our}
    \vspace{-4mm}
\end{figure}

\subsection{\Ourmethod}
\label{sec:tbd}

The forward noising process implicitly defines a progression of image manifolds, where each manifold consists of noisier images. Each step of the reverse, denoising diffusion process, projects a noisy image onto the next, less noisy, manifold.
To create a seamless result where the masked region complies with a guiding prompt, while the rest of the image is identical to the original input, we spatially blend each of the noisy images progressively generated by the CLIP-guided process with the \emph{corresponding noisy version} of the input image. Our key insight is that, while in each step along the way, the result of blending the two noisy images is not guaranteed to be coherent, the denoising diffusion step that follows each blend, restores coherence by projecting onto the next manifold. This process is depicted in \Cref{fig:diffusion_process_illustration_our} and summarized in \Cref{alg:final}.

\subsubsection{Background preserving blending}
\label{sec:background_preservation_blending}

A \naive{} way to preserve the background is to let the CLIP-guided diffusion process generate an image $\widehat{x}$ without any background constraints (by setting $\lambda = 0$ in \Cref{alg:local_clip_guided_diffusion}). Next, replace the generated background with the original one, taken from the input image: $\widehat{x} \odot m + x \odot (1 - m)$. 
The obvious problem is that combining the two images in this manner fails to produce a coherent, seamless result.
See the supplementary for an example.

In their pioneering work, Burt and Adelson~\cite{burt1987laplacian} show that two images can be blended smoothly by separately blending each level of their Laplacian pyramids.
Inspired by this technique, we propose to perform the blending at different noise levels along the diffusion process.
Our key hypothesis is that at each step during the diffusion process, a noisy latent is projected onto a manifold of natural images noised to a certain level. While blending two noisy images (from the same level) yields a result that likely lies outside the manifold, the next diffusion step projects the result onto the next level manifold, thus ameliorating the incoherence.

Thus, at each stage, starting from a latent $x_t$, we perform a single CLIP-guided diffusion step, that denoises the latent in a direction dependent on the text prompt, yielding a latent denoted $x_{t-1, \textit{fg}}$. In addition, we obtain a noised version of the background $x_{t-1, \textit{bg}}$ from the input image using \Cref{eqn:fast_forward}.
The two latents are now blended using the mask: $x_{t-1} = x_{t-1, \mathit{fg}} \odot m + x_{t-1, \mathit{bg}} \odot (1 - m)$, and the process is repeated (see \Cref{fig:diffusion_process_illustration_our} and \Cref{alg:final}).

In the final step, the entire region outside the mask is replaced with the corresponding region from the input image, thus strictly preserving the background. 

\subsubsection{Extending augmentations}
\label{sec:mitigating_adversarial_examples}
Adversarial examples \cite{szegedy2014intriguing, goodfellow2014explaining} is a well known phenomenon that may occur when optimizing an image \emph{directly on its pixel values}. For example, a classifier can be easily fooled to classify an image incorrectly by slightly altering its pixels in the direction of their gradients with respect to some wrong class. Adding such adversarial noise will not be perceived by a human, but the classification will be wrong.

Similarly, gradual changes of pixel values by CLIP-guided diffusion, might result in reducing the CLIP loss without creating the desired high-level semantic change in the image. We find that this phenomenon frequently occurs in practice. Bau et al. \cite{bau2021paint} also experienced this issue and addressed it using a non-gradient method that is based on evolution strategy.

We hypothesized that this problem can be mitigated by performing several augmentations on the intermediate result estimated at each diffusion step, and calculating the gradients using CLIP on each of the augmentations separately. This way, in order to ``fool'' CLIP, the manipulation must do so on all the augmentations, which is harder to achieve without a high-level change in the image. Indeed, we find that a simple augmentation technique mitigates this problem: given the current estimated result $\widehat{x}_0$, instead of taking the gradients of the CLIP loss directly, we compute them with respect to several projectively transformed copies of this image. These gradients are then averaged together. We term this strategy as ``extending augmentation''. The effect of these augmentations is studied in \Cref{sec:ablation_study}. We've added extending augmentations to our method (\Cref{alg:final}) as well as to the Local CLIP GD baseline (\Cref{alg:local_clip_guided_diffusion}) for all the comparisons in this paper.

\subsubsection{Result ranking}
\label{sec:generation_ranking}

\Cref{alg:final} can generate multiple outputs for the same input; this is a desirable feature because our task is one-to-many by its nature. Similarly to \cite{razavi2019generating, ramesh2021zero}, we found it beneficial to generate multiple predictions, rank them and choose those with the higher scores. For the ranking, we utilize the CLIP model using the same $\mathcal{D}_{\mclip}$ from \Cref{eqn:d_clip} on the final results, without the extending augmentations.
\begin{figure*}[h]
    \centering
    \setlength{\tabcolsep}{0.5pt}
    \renewcommand{\arraystretch}{0.5}
    \setlength{\ww}{0.24\columnwidth}
  
    \begin{tabular}{cccccc}
        \rotatebox{90}{\scriptsize\phantom{AA}Input + mask}
        \includegraphics[width=\ww,frame]{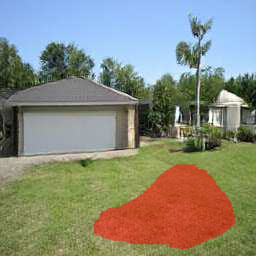} &
        \includegraphics[width=\ww,frame]{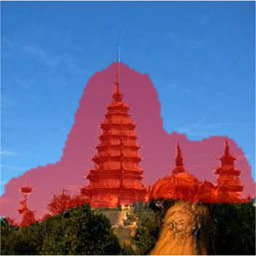} &
        \includegraphics[width=\ww,frame]{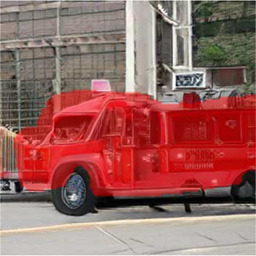} &
        \includegraphics[width=\ww,frame]{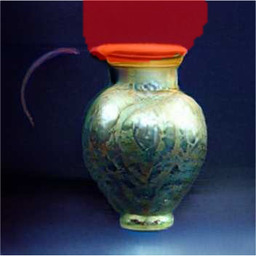} &
        \includegraphics[width=\ww,frame]{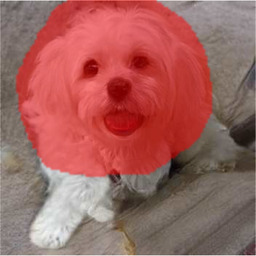} &
        \includegraphics[width=\ww,frame]{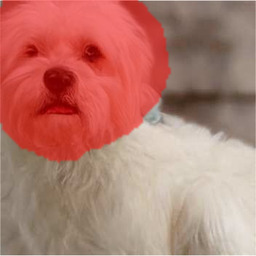} \\
        
        \rotatebox{90}{\scriptsize\phantom{AAAAA} (1)}
        \includegraphics[width=\ww,frame]{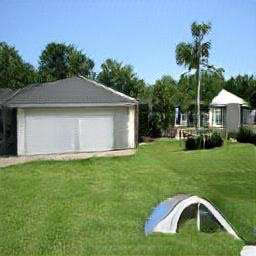} &
        \includegraphics[width=\ww,frame]{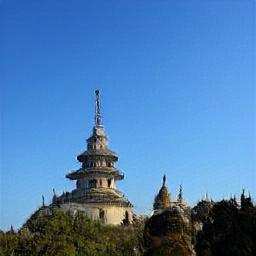} &
        \includegraphics[width=\ww,frame]{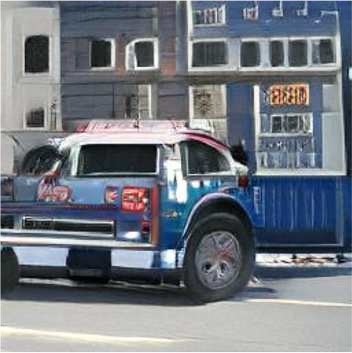} &
        \includegraphics[width=\ww,frame]{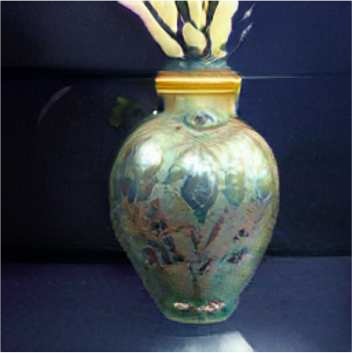} &
        \includegraphics[width=\ww,frame]{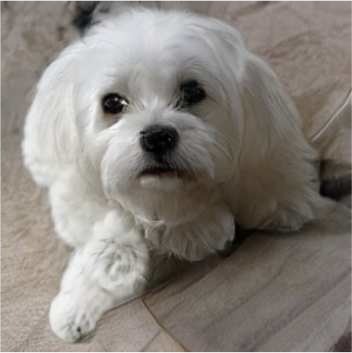} &
        \includegraphics[width=\ww,frame]{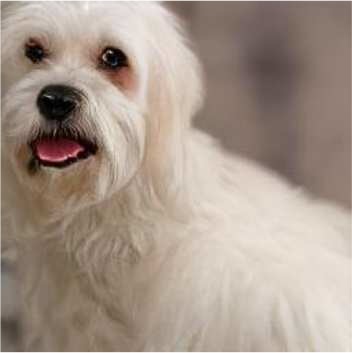} \\
        
        \rotatebox{90}{\scriptsize\phantom{AAAAA}{(2)}}
        \includegraphics[width=\ww,frame]{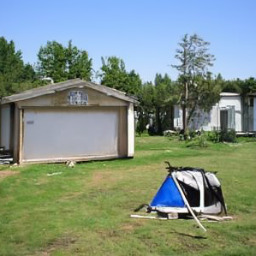} &
        \includegraphics[width=\ww,frame]{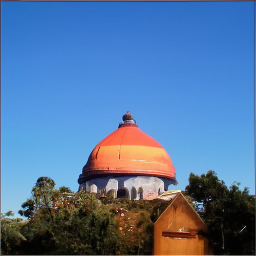} &
        \includegraphics[width=\ww,frame]{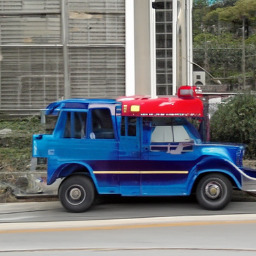} &
        \includegraphics[width=\ww,frame]{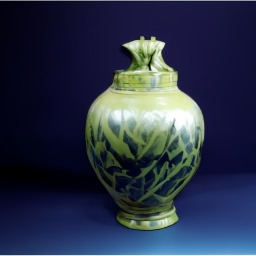} &
        \includegraphics[width=\ww,frame]{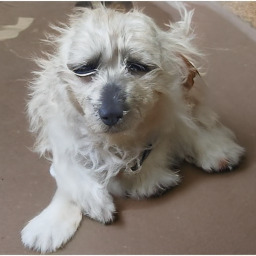} &
        \includegraphics[width=\ww,frame]{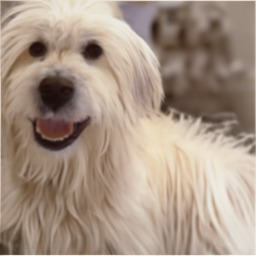} \\
        
        \rotatebox{90}{\scriptsize\phantom{AAAAA}{(3)}}
        \includegraphics[width=\ww,frame]{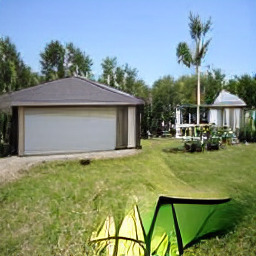} &
        \includegraphics[width=\ww,frame]{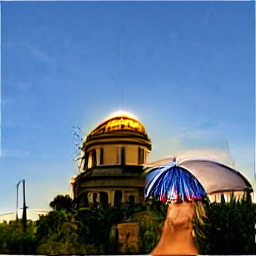} &
        \includegraphics[width=\ww,frame]{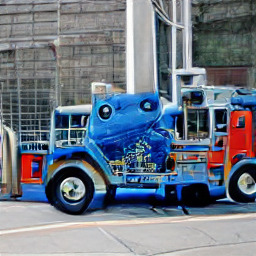} &
        \includegraphics[width=\ww,frame]{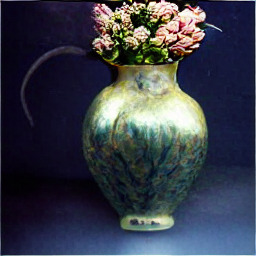} &
        \includegraphics[width=\ww,frame]{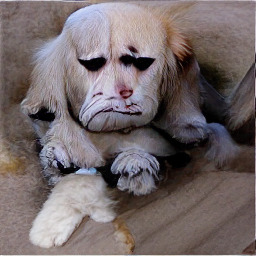} &
        \includegraphics[width=\ww,frame]{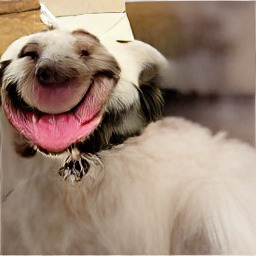} \\
        
        \rotatebox{90}{\scriptsize\phantom{AAAAA}Our}
        \includegraphics[width=\ww,frame]{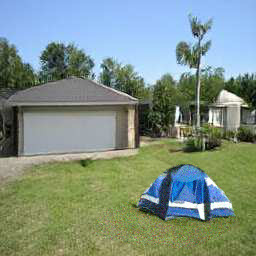} &
        \includegraphics[width=\ww,frame]{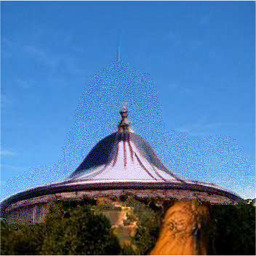} &
        \includegraphics[width=\ww,frame]{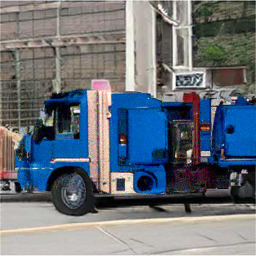} &
        \includegraphics[width=\ww,frame]{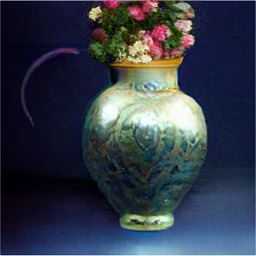} &
        \includegraphics[width=\ww,frame]{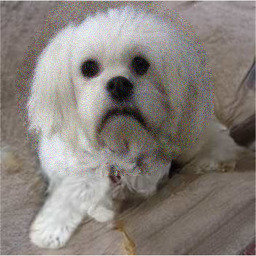} &
        \includegraphics[width=\ww,frame]{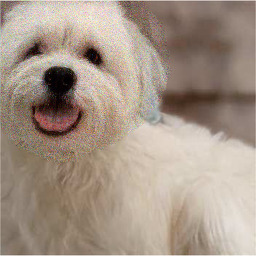} \\
        
        \scriptsize{``A photo of a} & 
        \scriptsize{``A photo of a} & 
        \scriptsize{``A photo of a} & 
        \scriptsize{``A photo of a} & 
        \scriptsize{``A photo of a} &
        \scriptsize{``A photo of a} \\
        
        \scriptsize{camping tent''} & 
        \scriptsize{magnificent dome''} &
        \scriptsize{blue firetruck''} &
        \scriptsize{bouquet of flowers''} & 
        \scriptsize{sad dog''} & 
        \scriptsize{happy dog''} \\
    \end{tabular}
    
    \caption{\textbf{Comparison using examples from \textsl{Paint By Word}~\cite{bau2021paint}.} We use the GAN-generated input images, and user-provided masks and text prompts from Bau et al.~\cite{bau2021paint}, as well as their results (1).
	In the next two rows, we show results of two other baselines: (2) Local CLIP GD \cite{clip_guided_diffusion} and (3) $\textit{PaintByWord++}$ \cite{vqgan_clip, bau2021paint}.
    Our results (bottom row) exhibit more realistic objects. Moreover, our method perfectly preserves the background region of the input image, while other methods change it.}

    \label{fig:comparison_paint_by_word_original}
    \vspace{-2mm}
\end{figure*}

\section{Results}
\label{sec:results}

We begin by comparing our method to previous methods and baselines both qualitatively and quantitatively. Next, we demonstrate the effect of our use of extending augmentations. Finally, we demonstrate several applications enabled by our method.

\subsection{Comparisons}

\begin{figure*}[h]
    \centering
    \setlength{\tabcolsep}{0.5pt}
    \renewcommand{\arraystretch}{0.5}
    \setlength{\ww}{0.23\columnwidth}
  
    \begin{tabular}{ccccccccc}
        \rotatebox{90}{\scriptsize\phantom{AA} Input + mask} &
        \includegraphics[width=\ww,frame]{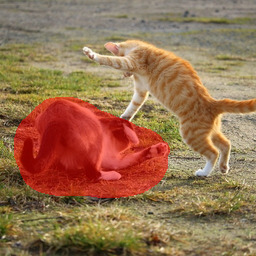} &
        \includegraphics[width=\ww,frame]{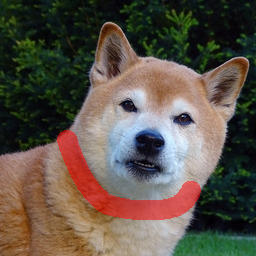} &
        \includegraphics[width=\ww,frame]{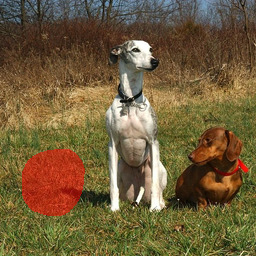} &
        \includegraphics[width=\ww,frame]{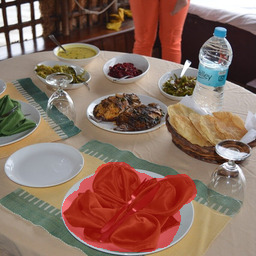} &
        \includegraphics[width=\ww,frame]{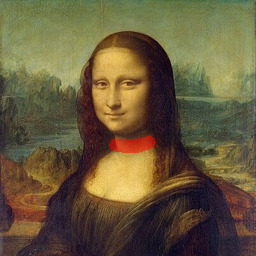} &
        \includegraphics[width=\ww,frame]{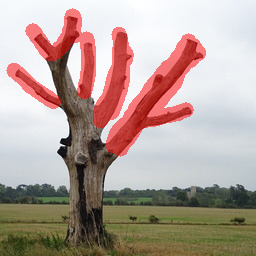} &
        \includegraphics[width=\ww,frame]{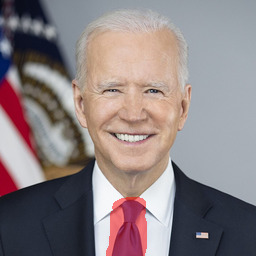} &
        \includegraphics[width=\ww,frame]{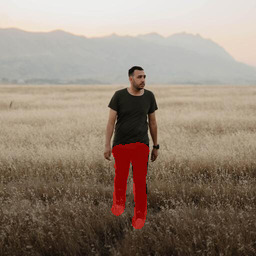} \\
        
        \rotatebox{90}{\scriptsize\phantom{AAAAA}{(1)}} &
        \includegraphics[width=\ww,frame]{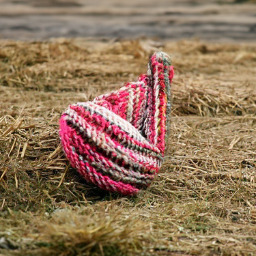} &
        \includegraphics[width=\ww,frame]{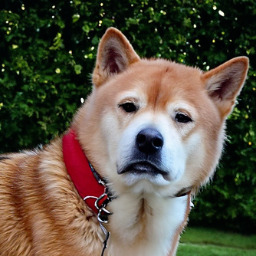} &
        \includegraphics[width=\ww,frame]{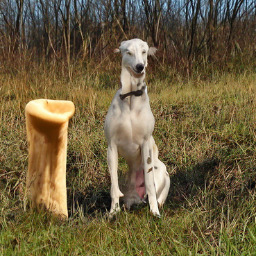} &
        \includegraphics[width=\ww,frame]{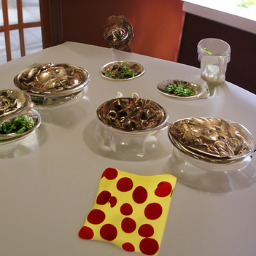} &
        \includegraphics[width=\ww,frame]{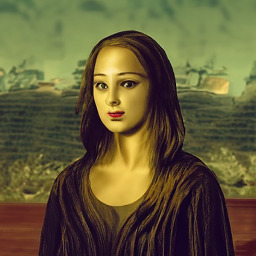} &
        \includegraphics[width=\ww,frame]{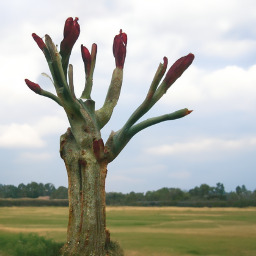} & 
        \includegraphics[width=\ww,frame]{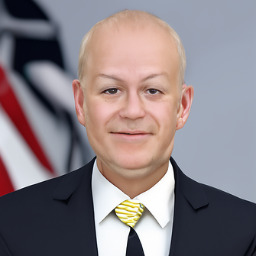} &
        \includegraphics[width=\ww,frame]{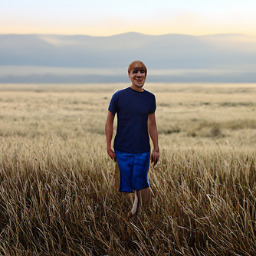} \\
        
        \rotatebox{90}{\scriptsize\phantom{AAAAAA}{(2)}} &
        \includegraphics[width=\ww,frame]{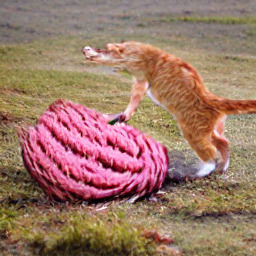} &
        \includegraphics[width=\ww,frame]{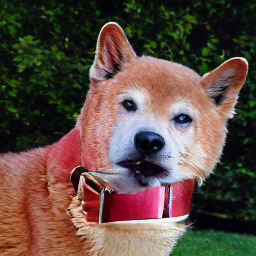} &
        \includegraphics[width=\ww,frame]{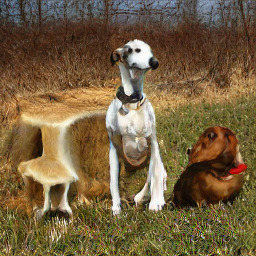} &
        \includegraphics[width=\ww,frame]{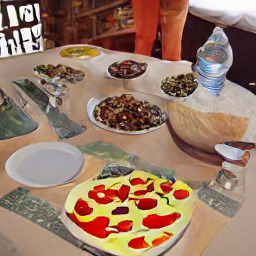} & 
        \includegraphics[width=\ww,frame]{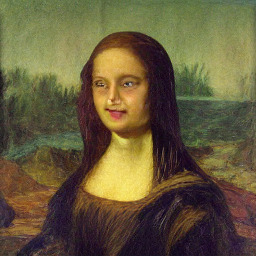} & 
        \includegraphics[width=\ww,frame]{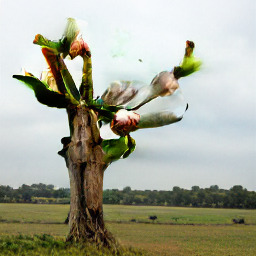} &
        \includegraphics[width=\ww,frame]{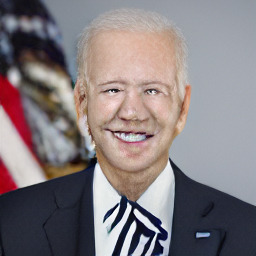} &
        \includegraphics[width=\ww,frame]{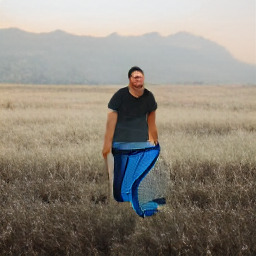} \\
        
        \rotatebox{90}{\scriptsize\phantom{AAAAA} Our} &
        \includegraphics[width=\ww,frame]{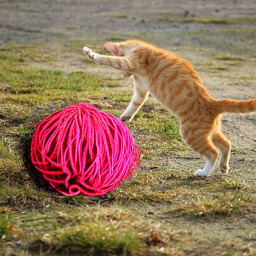} &
        \includegraphics[width=\ww,frame]{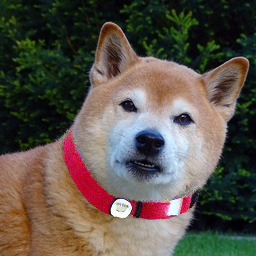} &
        \includegraphics[width=\ww,frame]{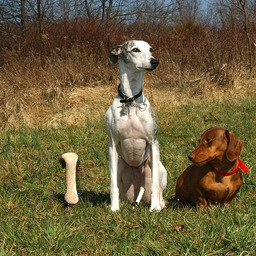} & 
        \includegraphics[width=\ww,frame]{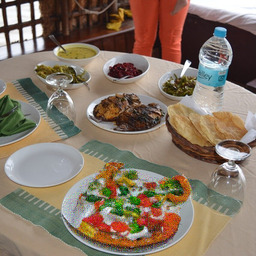} &
        \includegraphics[width=\ww,frame]{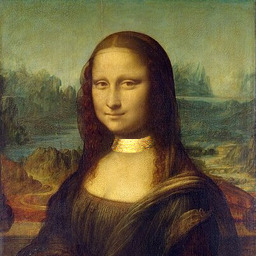} &
        \includegraphics[width=\ww,frame]{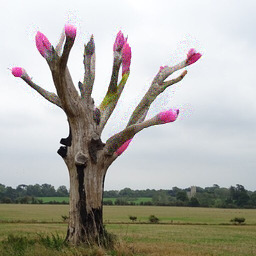} &
        \includegraphics[width=\ww,frame]{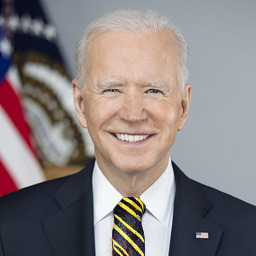} &
        \includegraphics[width=\ww,frame]{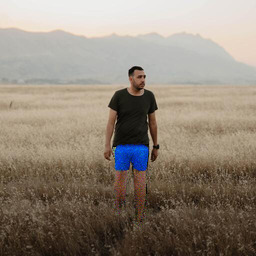} \\
        
        &
        \scriptsize{``pink yarn ball''} &
        \scriptsize{``red dog collar''} &
        \scriptsize{``dog bone''} &
        \scriptsize{``pizza''} &
        \scriptsize{``golden necklace''} &
        \scriptsize{``blooming tree''} &
        \scriptsize{``tie with black} &
        \scriptsize{``blue short pants''} \\
        
        &&&&&&& \scriptsize{and yellow stripes''} & \\
    \end{tabular}
    
    \caption{\textbf{Comparison to baselines on real images:}  A comparison with (1) Local CLIP-guided diffusion \cite{clip_guided_diffusion} and (2) $\textit{PaintByWord++}$ \cite{bau2021paint, vqgan_clip}. Both baselines fail to preserve the background and produce results that are less natural/coherent, in contrast to the results of our method.}
    \label{fig:baselines_comparison_real_images}
    \vspace{-2mm}
\end{figure*}
\begin{figure}[h]
    \centering
    \setlength{\tabcolsep}{0.5pt}
    \renewcommand{\arraystretch}{0.5}
    \setlength{\ww}{0.24\columnwidth}
  
    \begin{tabular}{cccc}
        \rotatebox{90}{\scriptsize\phantom{AAA}{{``sausages''}}} &
        \includegraphics[width=\ww,frame]{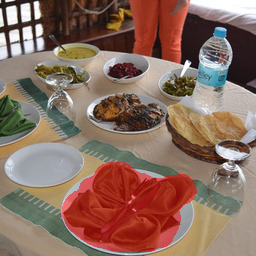} &
        \includegraphics[width=\ww,frame]{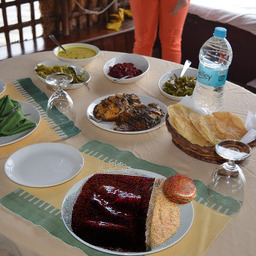} &
        \includegraphics[width=\ww,frame]{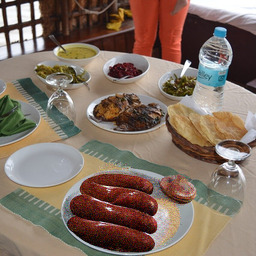} \\
        
        \rotatebox{90}{\scriptsize\phantom{AA}{{``huge avocado''}}} &
        \includegraphics[width=\ww,frame]{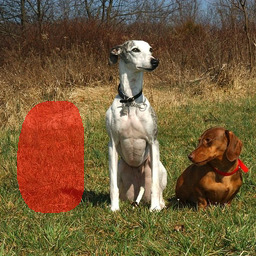} &
        \includegraphics[width=\ww,frame]{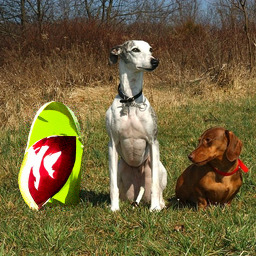} &
        \includegraphics[width=\ww,frame]{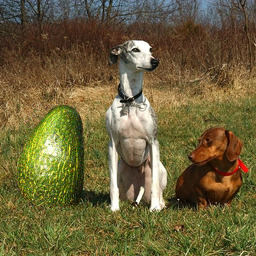} \\
        
        &
        \scriptsize{Input + mask} &
        \scriptsize{(1)} &
        \scriptsize{(2)} \\
    \end{tabular}
    
    \caption{\textbf{Extending augmentations ablation:}
    Using the same random seed and inputs, we compared the generated results (1) without extending augmentations and (2) with them. The augmentations make the resulting images more natural and coherent with the background. See supplementary material for more examples.}
    \label{fig:ablation_study_augmentations_mini}
    \vspace{-4mm}
\end{figure}

In \Cref{fig:comparison_paint_by_word_original} we compare the text-driven edits performed by our method to those performed using (1) \textit{PaintByWord}~\cite{bau2021paint}; (2) local CLIP-guided diffusion, as described in \Cref{alg:local_clip_guided_diffusion}, with $\lambda=1000$; and (3) VQGAN-CLIP + Paint By Word \cite{vqgan_clip, bau2021paint}. For the latter, we adapt VQGAN-CLIP \cite{vqgan_clip} to support masks using the same $\mathcal{D}_{\mclip}$ loss from \Cref{eqn:d_clip}. In addition, we find that results can be improved by optimizing only part of the VQGAN \cite{esser2021taming} latent space that corresponds to the edited area, similarly to the process in Bau et al.~\cite{bau2021paint}. Because VQGAN includes a pretrained decoder, we can easily use this method on real images. We denote this method \textit{PaintByWord++}.

Since the implementation of Bau et al.~\cite{bau2021paint} is not currently available, we perform this comparison using the examples included in their paper.
Note that since \textit{PaintByWord} operates only on GAN-generated images, all the input images in this comparison are synthetic and somewhat unnatural. In order to achieve better results on places, Bau et al.~\cite{bau2021paint} used two different models: one that is trained on MIT Places~\cite{zhou2014learning} and the other on ImageNet \cite{deng2009imagenet}. In contrast, our method can operate on real images and uses a single DPPM model that was trained on ImageNet.

The results shown in \Cref{fig:comparison_paint_by_word_original} demonstrate that although \textit{PaintByWord} and the other two baselines all encourage background preservation, the background is not always preserved and some global changes occur in almost all cases. Furthermore, in each of the rows (1)--(3) there are some results that appear unrealistic. In contrast, our method preserves the background perfectly, and the edits appear natural and coherent with the surrounding background.

\begin{table}
\begin{center}
\begin{adjustbox}{width=1.0\columnwidth}
 \begin{tabular}{lccc}
 \toprule
 
 \textbf{Method} & 
 Realism $\uparrow$ & 
 Background $\uparrow$ &
 Text match $\uparrow$\\
 
 \midrule
 $\textit{PaintByWord}$ \cite{bau2021paint} & $3.31\pm1.38$ & $3.25\pm1.33$ & $3.14\pm1.31$ \\
 Local CLIP GD \cite{clip_guided_diffusion} & $3.50\pm1.19$ & $3.11\pm1.24$ & $3.86\pm1.32$ \\
 $\textit{PaintByWord++}$ \cite{vqgan_clip, bau2021paint} & $1.94\pm1.36$ & $3.37\pm1.30$ & $3.01\pm1.38$ \\
 Ours & $\bm{3.93\pm1.08}$ & $\bm{4.73\pm0.61}$ & $\bm{4.63\pm0.77}$ \\
 \bottomrule
\end{tabular}
\end{adjustbox}
\caption{\textbf{User study results:} Participants were presented with the inputs and results shown in \Cref{fig:comparison_paint_by_word_original} and were asked to rate each result on a Likert scale of 1-5 according to the following criteria: overall result realism, background preservation, and correspondence between the text prompt and the outcome. The mean and standard deviation are shown for each method and criterion.}
\label{tab:user_study_paint_by_word}
\end{center}
\vspace{-6mm}
\end{table}

In order to obtain quantitative results, we conducted a preliminary user study comparing between the different results shown in \Cref{fig:comparison_paint_by_word_original}. Participants were asked to rate each result in terms of realism, background preservation, and correspondence to the text prompt. \Cref{tab:user_study_paint_by_word} shows that our method outperforms the three baselines in all of these aspects. Please see the supplementary for more details.

In \Cref{fig:baselines_comparison_real_images} we further compare our method to local CLIP-guided diffusion and \textit{PaintByWord++}, this time using real images as input.  Again, the results demonstrate the inability of the baseline methods to preserve the background, and exhibit lack of coherence between the edited region and its surroundings, in contrast to the results of our method.

\subsection{Ablation of extending augmentations}
\label{sec:ablation_study}
In order to assess the importance of the extending augmentation technique described \Cref{sec:mitigating_adversarial_examples}, we disable the extending augmentations completely from our method (\Cref{alg:final}). \Cref{fig:ablation_study_augmentations_mini} demonstrates the importance of the augmentations: the same random seed is used in two runs, one with and the other without augmentations. We can see that the images generated with the use of augmentations are more visually plausible and are more coherent than the ones generated without the augmentations.

\begin{figure}[h]
    \centering
    \setlength{\tabcolsep}{0.5pt}
    \renewcommand{\arraystretch}{0.5}
    \setlength{\ww}{0.24\columnwidth}
  
    \begin{tabular}{cccc}        
        \includegraphics[width=\ww,frame]{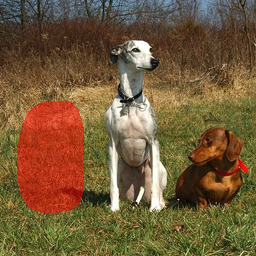} &
        \includegraphics[width=\ww,frame]{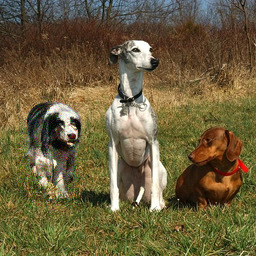} &
        \includegraphics[width=\ww,frame]{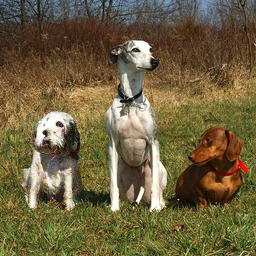} &
        \includegraphics[width=\ww,frame]{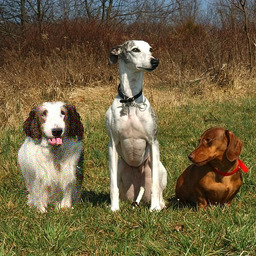} \\
        
		\includegraphics[width=\ww,frame]{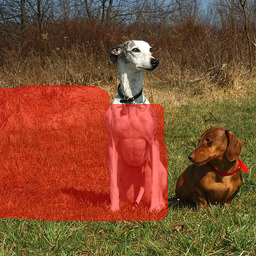} &
		\includegraphics[width=\ww,frame]{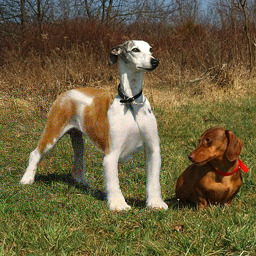} &
		\includegraphics[width=\ww,frame]{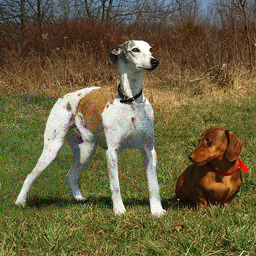} & 
		\includegraphics[width=\ww,frame]{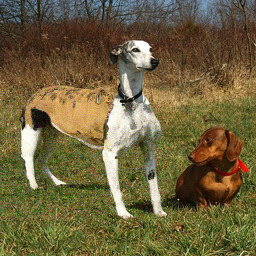} \\
		
		\scriptsize{Input + mask} & 
		\scriptsize{Result 1} & 
		\scriptsize{Result 2} & 
		\scriptsize{Result 3} \\
    \end{tabular}
    
    \caption{\textbf{Multiple outcomes:} Given the same guiding text (top row: ``a dog'', bottom row: ``body of a standing dog'') our method generates multiple plausible results.}
    \label{fig:multiple_outcomes}
    \vspace{-4mm}
\end{figure}

\subsection{Applications}
\label{sec:applications}
Our method is applicable to generic real-world images and may be used for a variety of applications. Below we demonstrate a few.

\textbf{Text-driven object editing:} we are able to add, remove or alter any of the existing objects in an image. \Cref{fig:multiple_outcomes} demonstrates the ability to add a new object to an image. Note that the method is able to generate a variety of plausible outcomes. Rather than completely replacing an object, only a part of it may be replaced, guided by a text prompt, as shown in the bottom row of \Cref{fig:multiple_outcomes}. \Cref{fig:teaser} demonstrates the ability to remove an object or replace it with a new one. Removal is achieved by not providing any text prompt, and it is equivalent to traditional image inpainting, where no text or other guidance is involved.

\textbf{Background replacement:} rather than editing the foreground object, it is also possible to replace the background using text guidance, as demonstrated in \Cref{fig:teaser}. Additional
examples for foreground and background editing are included in supplementary results.

\begin{figure}[t]
    \centering
    \setlength{\tabcolsep}{0.5pt}
    \renewcommand{\arraystretch}{0.5}
    \setlength{\ww}{0.24\columnwidth}
  
    \begin{tabular}{cccc}
        \includegraphics[width=\ww,frame]{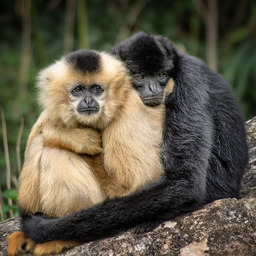} &
        \includegraphics[width=\ww,frame]{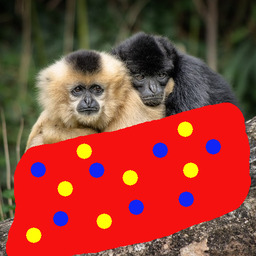} &
        \includegraphics[width=\ww,frame]{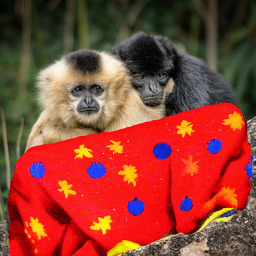} &
        \includegraphics[width=\ww,frame]{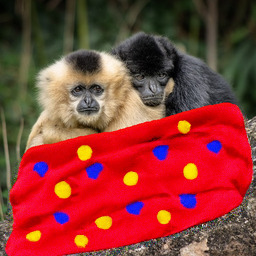} \\
        
        \scriptsize{Original image}  & 
        \scriptsize{Input scribble} & 
        \scriptsize{Result 1} & 
        \scriptsize{Result 2} \\
    \end{tabular}
    
    \caption{\textbf{Scribble-guided editing:} Users scribble a rough shape of the object they want to insert, mark the edited area, and provide a guiding text - ``blanket''. The model uses the scribble as a general shape and color reference, transforming it to match the guiding text. Note that the scribble patterns can also change.}
    \label{fig:scribble_editing_monkeys_blanket}
    \vspace{-2mm}
\end{figure}

\textbf{Scribble-guided editing:} Due to the noising process of diffusion models, another image, or a user-provided scribble, may be used as a guide. For example, the user may scribble a rough shape on a background image, provide a mask (covering the scribble) to indicate the area that is allowed to change, as well as a text prompt.
Our method will transform the scribble into a natural object while attempting to match the prompt, as demonstrated in \Cref{fig:scribble_editing_monkeys_blanket}.

\begin{figure}[tb]
    \centering
    \setlength{\tabcolsep}{0.5pt}
    \renewcommand{\arraystretch}{0.5}
    
    \includegraphics[width=\columnwidth]{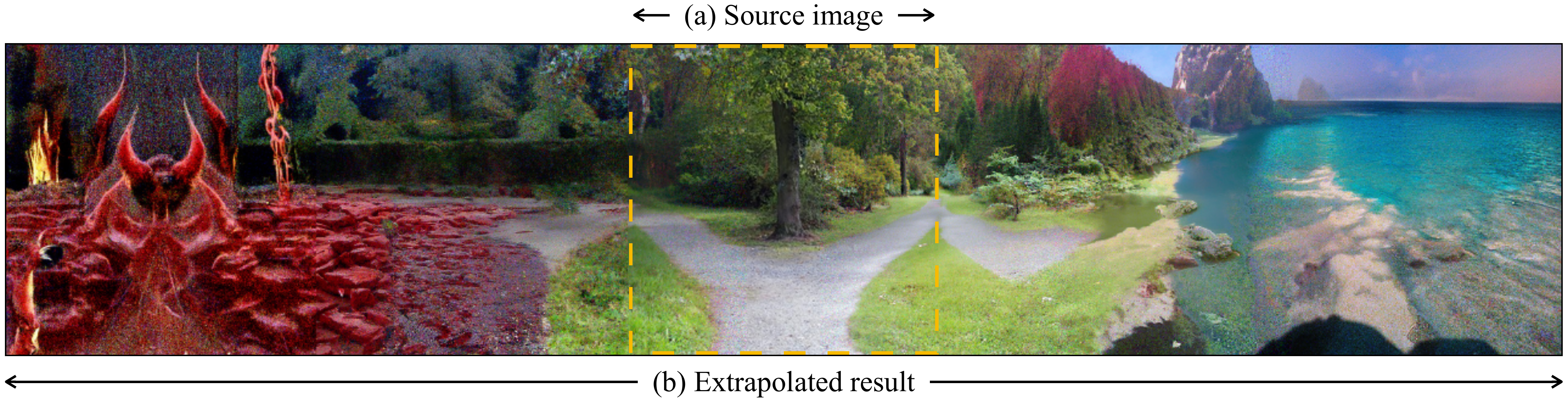}
    
    \caption{\textbf{Text-guided image extrapolation:} The user provides an input image and two text descriptions: ``hell'' and ``heaven''. The model extrapolates the image to the left using the ``hell'' prompt and to the right using the ``heaven'' prompt.}
    \label{fig:image_extrapolation_heaven_hell}
    \vspace{-5mm}
\end{figure}

\textbf{Text-guided image extrapolation} is the ability to extend an image beyond its boundaries, guided by a textual description, s.t. the resulting change is gradual. \Cref{fig:image_extrapolation_heaven_hell} demonstrates this ability: the user provides an image and two text prompts, each prompt is used to extrapolate the image in one direction. The resulting image can be arbitrarily wide (and mix multiple prompts). More details are provided in the supplementary material.

\section{Limitations and Future Work}
\label{sec:limitations}
The main limitation of our work is its inference time. Because of the sequential nature of DDPMs, generating a single image takes about 30 seconds on a modern GPU as described in the supplementary. In addition, we generate several samples and choose the top-ranked ones, as described in \Cref{sec:generation_ranking}.
This limits the applicability of our method for real-time applications and weak end-user devices (e.g. mobile devices). Further research in accelerating diffusion sampling is needed to address this problem.

\begin{figure}[t]
    \centering
    \setlength{\tabcolsep}{0.5pt}
    \renewcommand{\arraystretch}{0.5}
    \setlength{\ww}{0.24\columnwidth}
  
    \begin{tabular}{cccc}
        \includegraphics[width=\ww,frame]{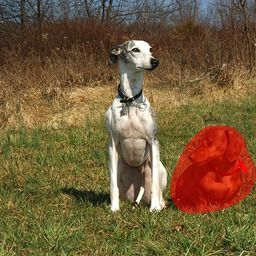} &
        \includegraphics[width=\ww,frame]{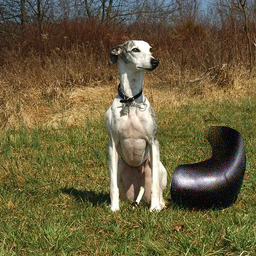} &
        \includegraphics[width=\ww,frame]{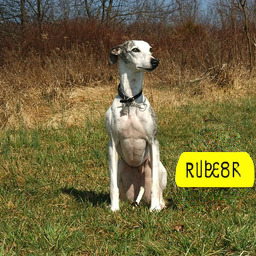} &
        \includegraphics[width=\ww,frame]{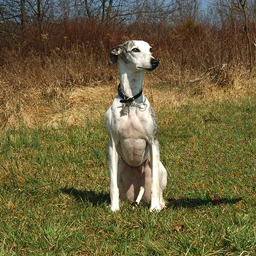} \\
        
        \scriptsize{Input + mask} & 
        \scriptsize{(1)} &
        \scriptsize{(2)} &
        \scriptsize{(3)} \\
    \end{tabular}
    
    \caption{\textbf{Failure cases:} Examples of failure cases given source image, mask and description ``rubber toy'': (1) partial object --- ranking by the edited area only may cause partial object to get a high score, (2) typographic bias --- the model can generate a sign with the word ``rubber'' on it, (3) missing object and unnatural shadows --- sometimes the method adds a shadow that is not coherent with the scene and does not correspond to the text.}
    \label{fig:failure_cases}
    \vspace{-4mm}
\end{figure}

In addition, the ranking method presented in \Cref{sec:generation_ranking} is not perfect because it takes into account only the edited area without the entire context of the image. So, bad results that contain only part of the desired object, may still get a high score, as demonstrated in \Cref{fig:failure_cases} (1). A better ranking system will enable our method to produce more compelling and coherent results.

Furthermore, because our model is based on CLIP, it inherits its weaknesses and biases. It was shown \cite{goh2021multimodal} that CLIP is susceptible to \emph{typographic attacks} - exploiting the model’s ability to read text robustly, they found that even photographs of hand-written text can often fool the model. \Cref{fig:failure_cases} (2) demonstrates that this phenomenon can occur even when generating images: instead of generating an image of a ``rubber toy'' our method generates a sign with the word ``rubber''.

One avenue for further research is training a version of CLIP that is agnostic to Gaussian noise. This may be done by training a version of CLIP that gets as an input a noisy image, a noise level, and the description text, and embeds the image and the text to a shared embedding space using contrastive loss. The noising process during training should be the same as in \Cref{eqn:fast_forward}.

Yet another avenue for research is extending our problem to other modalities such as a general-purpose text editor for 3D objects or videos.
\section{Societal Impact}
\label{sec:societal_impact}

Photo manipulations are almost as old as the photo creation process itself \cite{farid2009digital}.
Such manipulations can be used for art, entertainment, aesthetics, storytelling, and other legitimate use cases, but at the same time can also be used to tell lies via photos, for bullying, harassment, extortion, and may have psychological consequences \cite{10.1145/3326601}.
Indeed, our method can be used for all of the above. For example, it can be misused to add credibility to fake news, which is a growing concern in the current media climate.
It may also erode trust in photographic evidence and allow real events and real evidence to be brushed off as fake \cite{access_hollywood}. 

While our work does not enable anything that was out of reach for professional image editors, it certainly adds to the ease-of-use of the manipulation process, thus allowing users with limited technical capabilities to manipulate photos. We are passionate about our research, not only due to the legitimate use-cases, but also because we believe such research must be conducted openly in academia and not kept secret. We will provide our code for the benefit of the academic community, and we are actively working on the complement of this work: image and video forensic methods.

\section{Conclusions}

We introduced a novel solution to the problem of text-driven editing of natural images and demonstrated its superiority over the baselines. We believe that editing natural images using free text is a highly intuitive interaction, that will be further developed to 
a level which will make it an indispensable tool in the arsenal of every content creator.

\smallskip
\textbf{Acknowledgments} This work was supported in part by Lightricks Ltd and by the Israel Science Foundation (grants No. 2492/20 and 1574/21).

\clearpage
{\small
\bibliographystyle{ieee_fullname}
\bibliography{egbib}
}

\clearpage
\appendix
\section{Additional Examples}
In this section we provide additional examples of the applications and the failure cases that were mentioned in the main paper. 
In addition, we show that our method naturally supports an iterative editing process.
Lastly, we demonstrate the \naive{} blending approach (main paper, \Cref{sec:background_preservation_blending}).

\subsection{Applications --- Additional Examples}
\begin{figure*}[ht]
    \centering
    \setlength{\tabcolsep}{0.5pt}
    \renewcommand{\arraystretch}{0.5}
    \setlength{\ww}{0.33\columnwidth}
  
    \begin{tabular}{cccccc}        
        \includegraphics[width=\ww,frame]{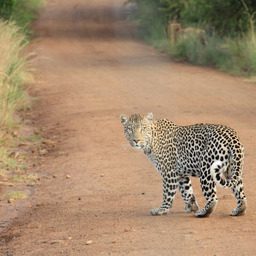} &
        \includegraphics[width=\ww,frame]{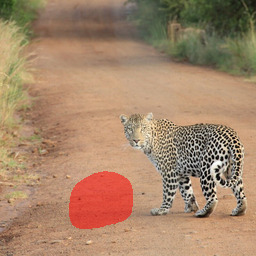} &
        \includegraphics[width=\ww,frame]{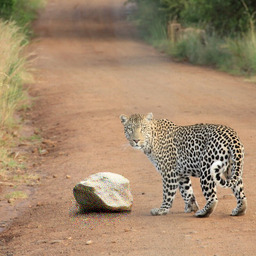} &
        \includegraphics[width=\ww,frame]{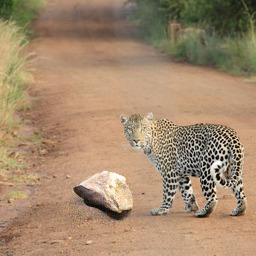} &
        \includegraphics[width=\ww,frame]{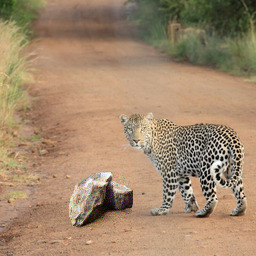} &
        \includegraphics[width=\ww,frame]{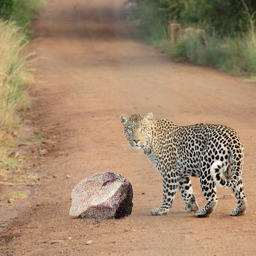}
        \\

        Input image &
        Image mask &
        Result 1 &
        Result 2 &
        Result 3 &
        Result 4 \\

        \includegraphics[width=\ww,frame]{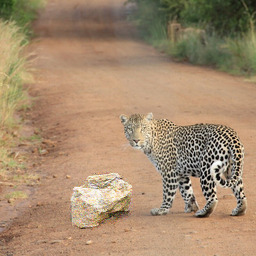} &
        \includegraphics[width=\ww,frame]{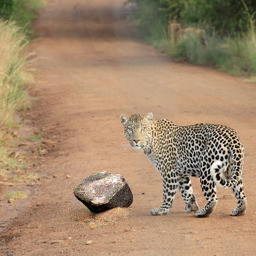} &
        \includegraphics[width=\ww,frame]{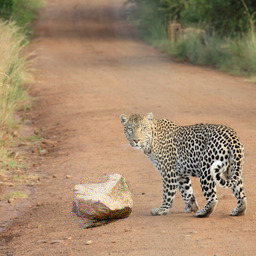} &
        \includegraphics[width=\ww,frame]{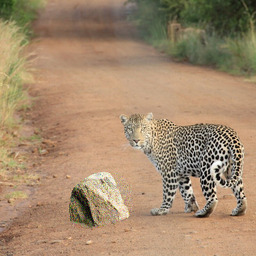} &
        \includegraphics[width=\ww,frame]{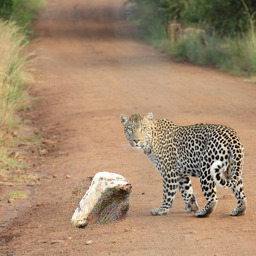} &
        \includegraphics[width=\ww,frame]{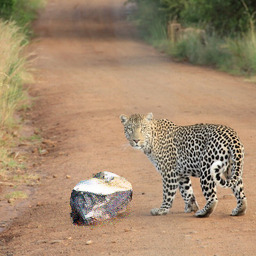}
        \\

        Result 5 &
        Result 6 &
        Result 7 &
        Result 8 &
        Result 9 &
        Result 10 
        \\
    \end{tabular}
    
    \caption{\textbf{Adding a new object (multiple results for the same input):} Given the input image, mask and text description ``rock'', our model is able to generate multiple plausible results.}
    \label{fig:leopard_multiple_results}
\end{figure*}

\begin{figure*}[ht]
    \centering
    \setlength{\tabcolsep}{0.5pt}
    \renewcommand{\arraystretch}{0.5}
    \setlength{\ww}{0.33\columnwidth}
  
    \begin{tabular}{cccccc}        
        \includegraphics[width=\ww,frame]{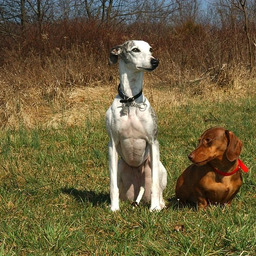} &
        \includegraphics[width=\ww,frame]{figures/foreground_editing/foreground_add_dogs/assets/mask_overlay.jpg} &
        \includegraphics[width=\ww,frame]{figures/foreground_editing/foreground_add_dogs/assets/pred5.jpg} &
        \includegraphics[width=\ww,frame]{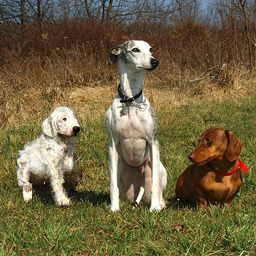} &
        \includegraphics[width=\ww,frame]{figures/foreground_editing/foreground_add_dogs/assets/pred3.jpg} &
        \includegraphics[width=\ww,frame]{figures/foreground_editing/foreground_add_dogs/assets/pred4.jpg} \\

        Input image &
        Image mask &
        Result 1 &
        Result 2 &
        Result 3 &
        Result 4 \\

        \includegraphics[width=\ww,frame]{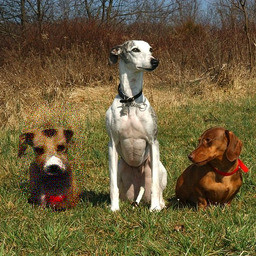} &
        \includegraphics[width=\ww,frame]{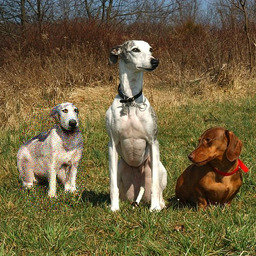} &
        \includegraphics[width=\ww,frame]{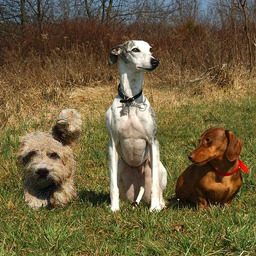} &
        \includegraphics[width=\ww,frame]{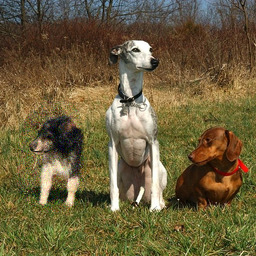} &
        \includegraphics[width=\ww,frame]{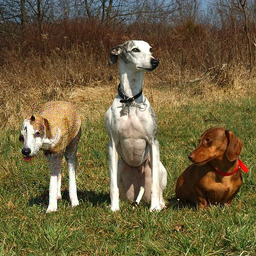} &
        \includegraphics[width=\ww,frame]{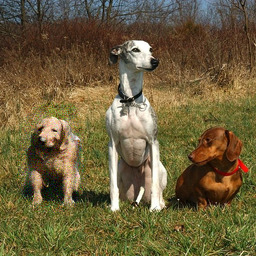} \\

        Result 5 &
        Result 6 &
        Result 7 &
        Result 8 &
        Result 9 &
        Result 10 \\
    \end{tabular}
    
    \caption{\textbf{Adding a new object (multiple results for the same input):} Given the input image, mask, and text description ``a dog'', our model is able to generate multiple plausible results. Some results are better (first row) than others (second row).
    }
    \label{fig:foreground_add_dogs}
\end{figure*}

\begin{figure*}[ht]
    \centering
    \setlength{\tabcolsep}{0.5pt}
    \renewcommand{\arraystretch}{0.5}
    \setlength{\ww}{0.16\textwidth}
  
    \begin{tabular}{cccccc}
        \includegraphics[width=\ww,frame]{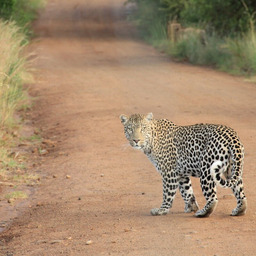} &
        \includegraphics[width=\ww,frame]{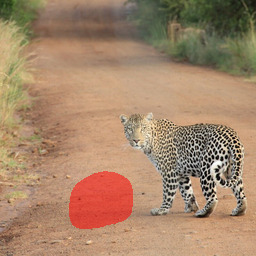} &
        \includegraphics[width=\ww,frame]{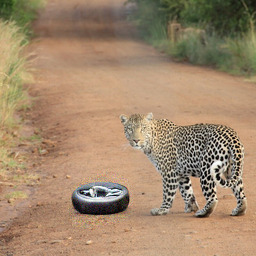} &
        \includegraphics[width=\ww,frame]{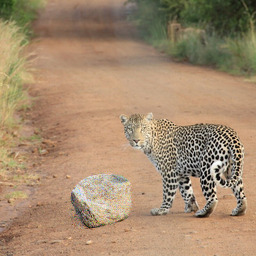} &
        \includegraphics[width=\ww,frame]{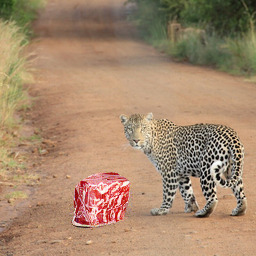} &
        \includegraphics[width=\ww,frame]{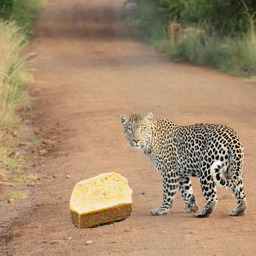}
        \\
        
        Input image & 
        Input mask & 
        ``car tire'' &
        ``big stone'' &
        ``meat'' &
        ``tofu''
        \\

        \includegraphics[width=\ww,frame]{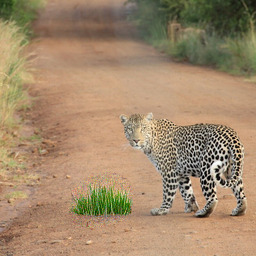} &
        \includegraphics[width=\ww,frame]{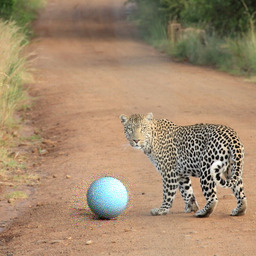} &
        \includegraphics[width=\ww,frame]{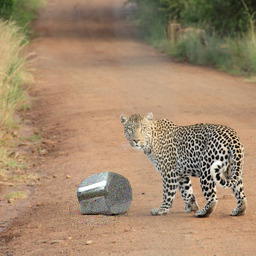} &
        \includegraphics[width=\ww,frame]{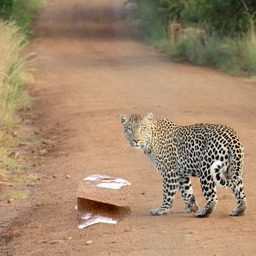} &
        \includegraphics[width=\ww,frame]{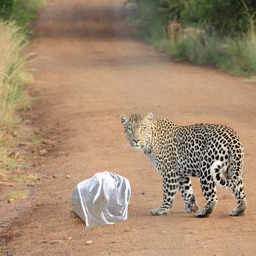} &
        \includegraphics[width=\ww,frame]{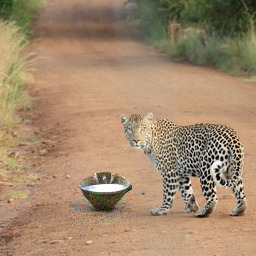}
        \\

        ``grass'' &
        ``blue ball'' &
        ``silver brick'' &
        ``ice cube'' &
        ``plastic bag'' &
        ``bowl of water'' 
        \\

        \includegraphics[width=\ww,frame]{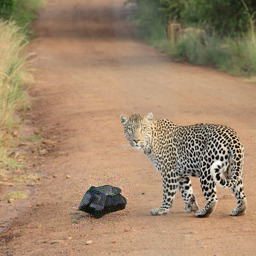} &
        \includegraphics[width=\ww,frame]{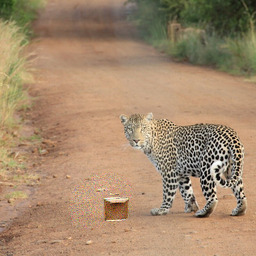} &
        \includegraphics[width=\ww,frame]{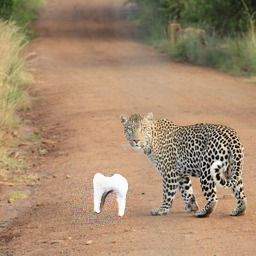} &
        \includegraphics[width=\ww,frame]{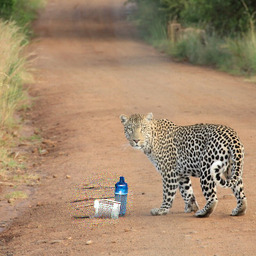} &
        \includegraphics[width=\ww,frame]{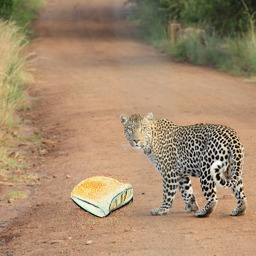} &
        \includegraphics[width=\ww,frame]{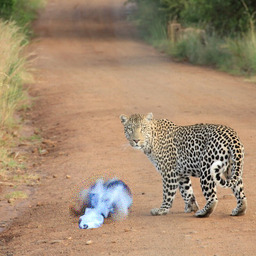}
        \\

        ``black rock'' &
        ``cardboard'' &
        ``tooth'' &
        ``water bottle'' &
        ``bread'' &
        ``smoke''
        \\

        \includegraphics[width=\ww,frame]{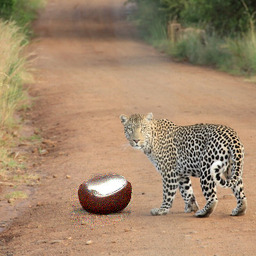} &
        \includegraphics[width=\ww,frame]{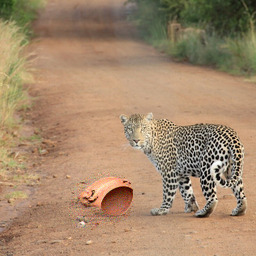} &
        \includegraphics[width=\ww,frame]{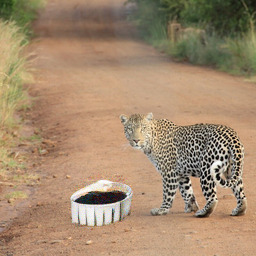} &
        \includegraphics[width=\ww,frame]{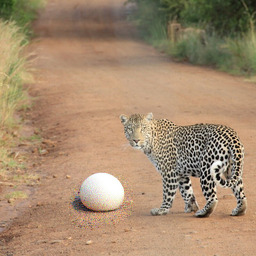} &
        \includegraphics[width=\ww,frame]{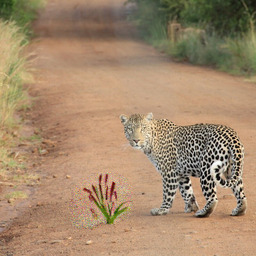} &
        \includegraphics[width=\ww,frame]{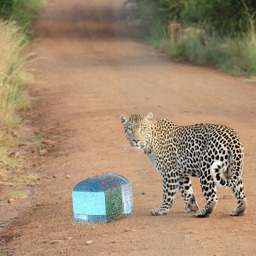}
        \\

        ``chocolate'' &
        ``clay pot'' &
        ``cola'' &
        ``egg'' &
        ``flower'' &
        ``glass''
        \\

        \includegraphics[width=\ww,frame]{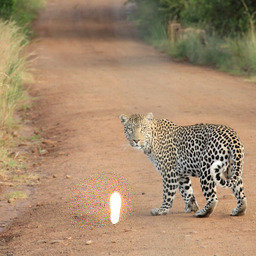} &
        \includegraphics[width=\ww,frame]{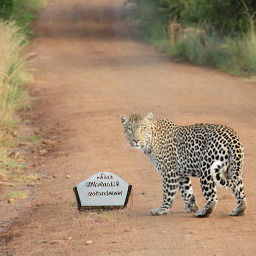} &
        \includegraphics[width=\ww,frame]{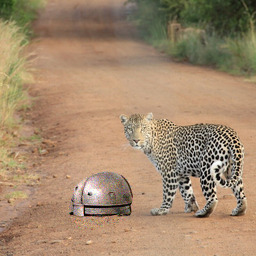} &
        \includegraphics[width=\ww,frame]{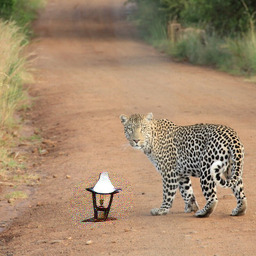} &
        \includegraphics[width=\ww,frame]{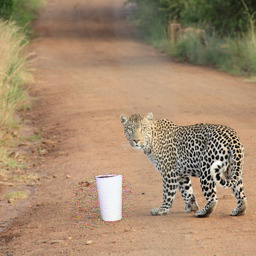} &
        \includegraphics[width=\ww,frame]{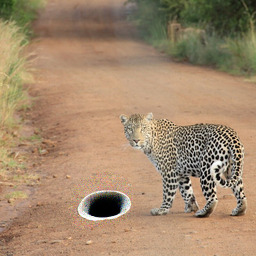}
        \\

        ``glow stick'' &
        ``gravestone'' &
        ``helmet'' &
        ``lamp'' &
        ``milk'' &
        ``hole''
        \\

    \end{tabular}
    
    \caption{\textbf{Adding a new object (different prompts):}  Given an input image and mask, our model is able to generate different objects corresponding to different text descriptions.}
    \label{fig:foreground_leopard1_add}
\end{figure*}

\begin{figure*}[ht]
    \centering
    \setlength{\tabcolsep}{0.5pt}
    \renewcommand{\arraystretch}{0.5}
    \setlength{\ww}{0.16\textwidth}
  
    \begin{tabular}{cccccc}
        \includegraphics[width=\ww,frame]{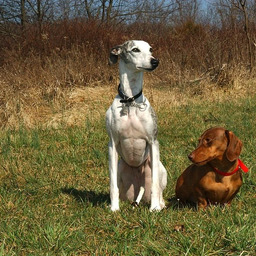} &
        \includegraphics[width=\ww,frame]{figures/foreground_editing/foreground_remove_dogs/assets/mask_overlay.jpg} &
        \includegraphics[width=\ww,frame]{figures/foreground_editing/foreground_remove_dogs/assets/pred1.jpg} &
        \includegraphics[width=\ww,frame]{figures/foreground_editing/foreground_remove_dogs/assets/pred2.jpg} &
        \includegraphics[width=\ww,frame]{figures/foreground_editing/foreground_remove_dogs/assets/pred3.jpg} &
        \includegraphics[width=\ww,frame]{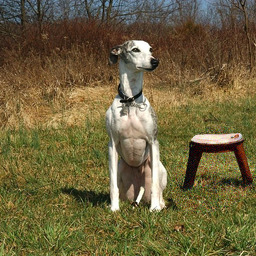} \\
        
        Input image & 
        Input mask & 
        No prompt & 
        ``white ball'' & 
        ``bowl of water''  & 
        ``stool'' \\

        \includegraphics[width=\ww,frame]{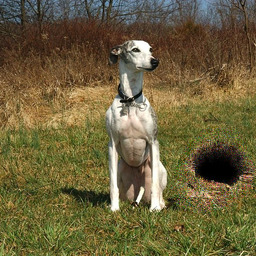} &
        \includegraphics[width=\ww,frame]{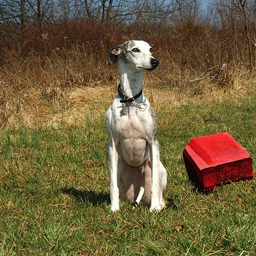} &
        \includegraphics[width=\ww,frame]{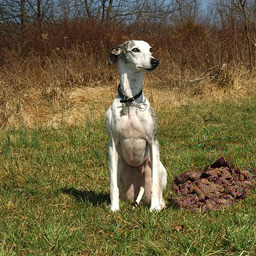} &
        \includegraphics[width=\ww,frame]{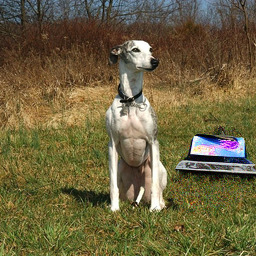} &
        \includegraphics[width=\ww,frame]{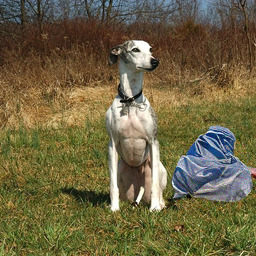} &
        \includegraphics[width=\ww,frame]{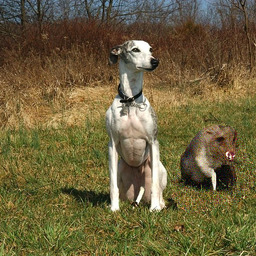} \\

        ``hole'' &
        ``red brick'' &
        ``pile of dirt'' &
        ``laptop'' &
        ``plastic bag'' &
        ``rat'' \\

        \includegraphics[width=\ww,frame]{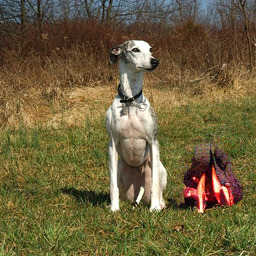} &
        \includegraphics[width=\ww,frame]{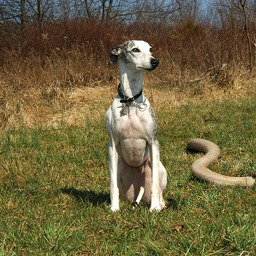} &
        \includegraphics[width=\ww,frame]{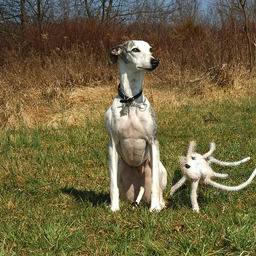} & 
        \includegraphics[width=\ww,frame]{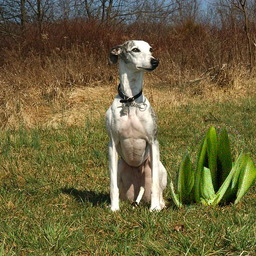} & 
        \includegraphics[width=\ww,frame]{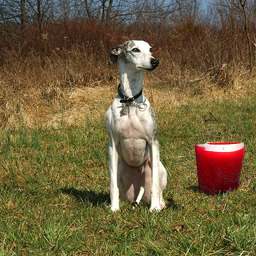} & 
        \includegraphics[width=\ww,frame]{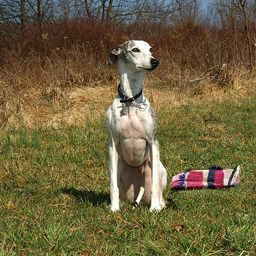}
        \\
        
        ``bonfire'' &
        ``snake'' &
        ``spider'' &
        ``plant'' &
        ``candle'' &
        ``blanket''
        \\

        \includegraphics[width=\ww,frame]{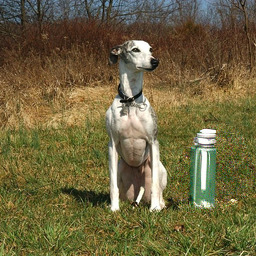} &
        \includegraphics[width=\ww,frame]{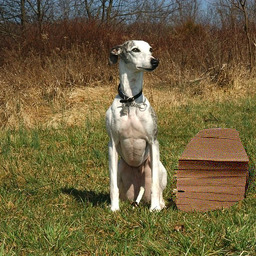} &
        \includegraphics[width=\ww,frame]{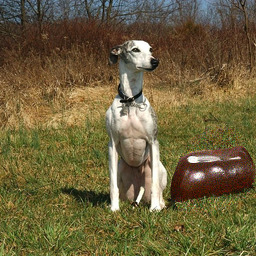} &
        \includegraphics[width=\ww,frame]{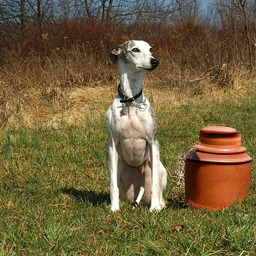} &
        \includegraphics[width=\ww,frame]{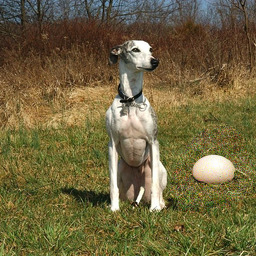} &
        \includegraphics[width=\ww,frame]{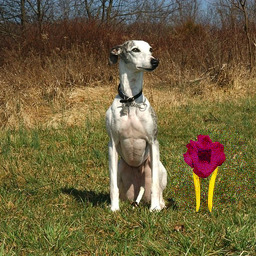}
        \\

        ``bottle'' &
        ``cardboard'' &
        ``chocolate'' &
        ``clay pot'' &
        ``egg'' &
        ``flower''
        \\

        \includegraphics[width=\ww,frame]{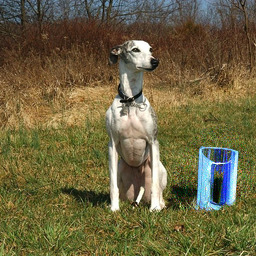} &
        \includegraphics[width=\ww,frame]{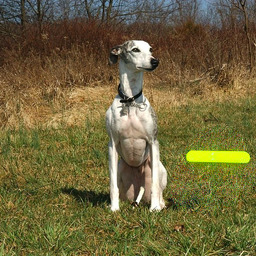} &
        \includegraphics[width=\ww,frame]{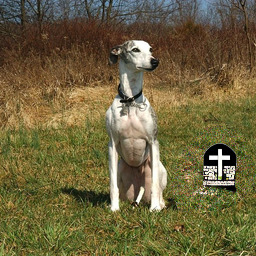} &
        \includegraphics[width=\ww,frame]{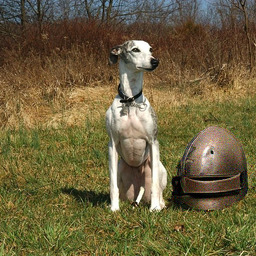} &
        \includegraphics[width=\ww,frame]{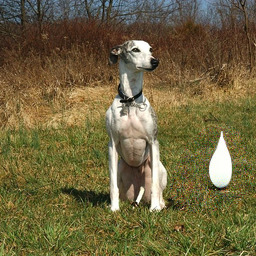} &
        \includegraphics[width=\ww,frame]{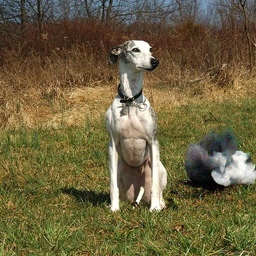}
        \\

        ``glass'' &
        ``glow stick'' &
        ``gravestone'' &
        ``helmet'' &
        ``milk'' &
        ``smoke''
        \\

    \end{tabular}
    
    \caption{\textbf{Removing/replacing a foreground object:}  Given an input image and a mask, we demonstrate inpainting of the masked region using different guiding texts. When no prompt is given, the result is similar to traditional image inpainting.}
    \label{fig:foreground_remove_dogs}
\end{figure*}

\begin{figure*}[ht]
    \centering
    \setlength{\tabcolsep}{1pt}
    \renewcommand{\arraystretch}{0.5}
    \setlength{\ww}{0.33\columnwidth}
  
    \begin{tabular}{cccccc}
        \includegraphics[width=\ww,frame]{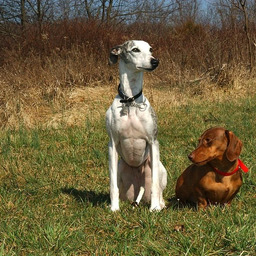} &
        \includegraphics[width=\ww,frame]{figures/foreground_editing/foreground_alter_dogs/assets/mask_overlay.jpg} &
        \includegraphics[width=\ww,frame]{figures/foreground_editing/foreground_alter_dogs/assets/pred4.jpg} &
        \includegraphics[width=\ww,frame]{figures/foreground_editing/foreground_alter_dogs/assets/pred1.jpg} &
        \includegraphics[width=\ww,frame]{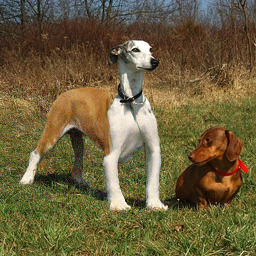} &
        \includegraphics[width=\ww,frame]{figures/foreground_editing/foreground_alter_dogs/assets/pred2.jpg} \\
        
        Input image & 
        Input mask & 
        Result 1 & 
        Result 2 & 
        Result 3 & 
        Result 4 \\

        \includegraphics[width=\ww,frame]{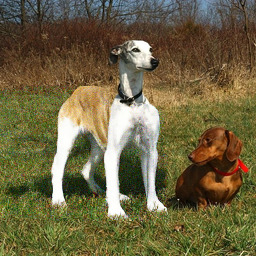} &
        \includegraphics[width=\ww,frame]{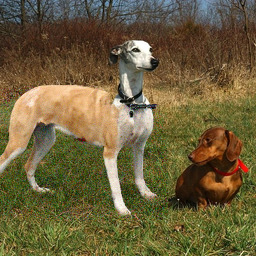} &
        \includegraphics[width=\ww,frame]{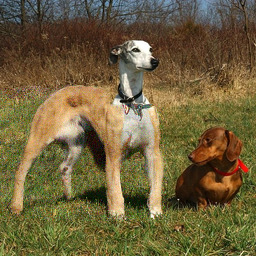} &
        \includegraphics[width=\ww,frame]{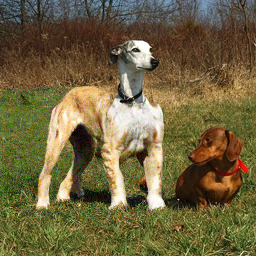} &
        \includegraphics[width=\ww,frame]{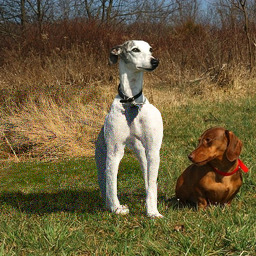} &
        \includegraphics[width=\ww,frame]{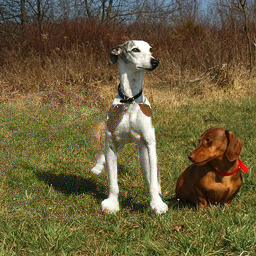} \\

        Result 5 & 
        Result 6 & 
        Result 7 &
        Result 8 & 
        Result 9 & 
        Result 10 \\

        \includegraphics[width=\ww,frame]{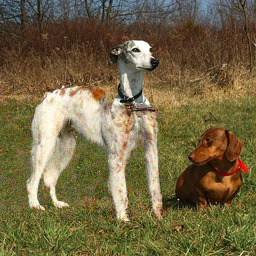} &
        \includegraphics[width=\ww,frame]{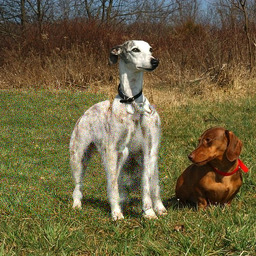} &
        \includegraphics[width=\ww,frame]{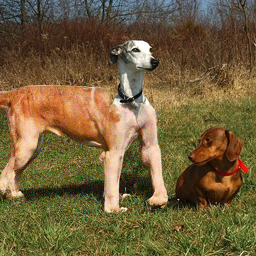} &
        \includegraphics[width=\ww,frame]{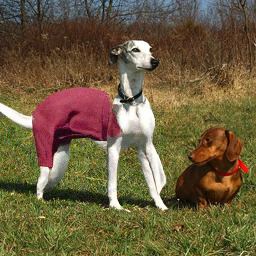} &
        \includegraphics[width=\ww,frame]{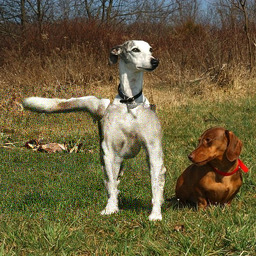} &
        \includegraphics[width=\ww,frame]{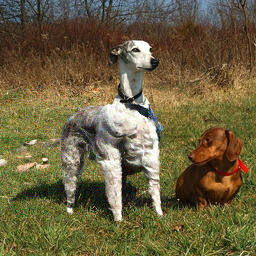} \\

        Result 11 & 
        Result 12 & 
        Result 13 &
        Result 14 & 
        Result 15 & 
        Result 16 \\

        \includegraphics[width=\ww,frame]{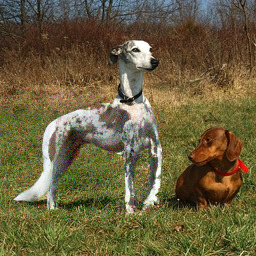} &
        \includegraphics[width=\ww,frame]{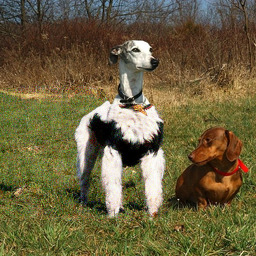} &
        \includegraphics[width=\ww,frame]{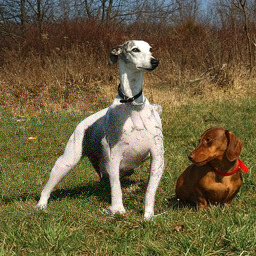} &
        \includegraphics[width=\ww,frame]{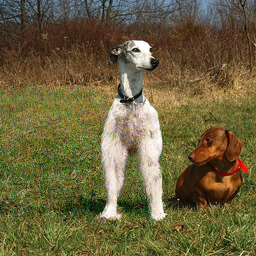} &
        \includegraphics[width=\ww,frame]{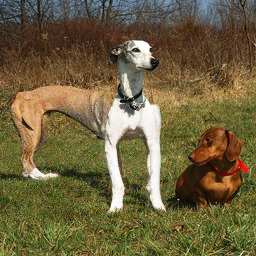} &
        \includegraphics[width=\ww,frame]{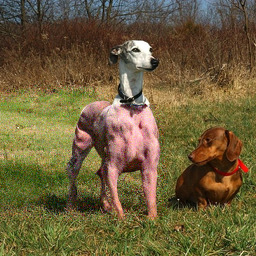} \\

        Result 17 & 
        Result 18 & 
        Result 19 &
        Result 20 & 
        Result 21 & 
        Result 22 \\

    \end{tabular}
    
    \caption{\textbf{Altering a part of an existing foreground object:}  Given an input image and a mask, we aim to alter the foreground object corresponding to the guiding text ``body of a standing dog''. Multiple plausible results are generated, some more plausible than others. (The first two rows are better than the bottom two rows.)}
    \label{fig:foreground_alter_dogs}
\end{figure*}

\begin{figure*}[ht]
    \centering
    \setlength{\tabcolsep}{1pt}
    \renewcommand{\arraystretch}{0.5}
    \setlength{\ww}{0.33\columnwidth}
  
    \begin{tabular}{cccccc}
        \includegraphics[width=\ww,frame]{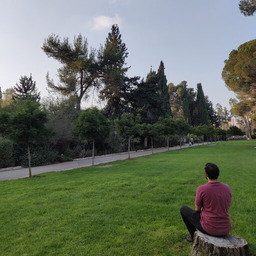} &
        \includegraphics[width=\ww,frame]{figures/background_editing/yossi_grass/assets/mask_overlay.jpg} &
        \includegraphics[width=\ww,frame]{figures/background_editing/yossi_grass/assets/pred1.jpg} &
        \includegraphics[width=\ww,frame]{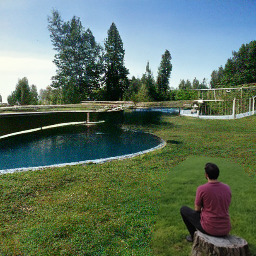} &
        \includegraphics[width=\ww,frame]{figures/background_editing/yossi_grass/assets/pred4.jpg} &
        \includegraphics[width=\ww,frame]{figures/background_editing/yossi_grass/assets/pred3.jpg} \\
        
        \scriptsize{Input image} & 
        \scriptsize{Input mask} & 
        \scriptsize{``big mountain''} & 
        \scriptsize{``swimming pool''} & 
        \scriptsize{``big wall''} & 
        \scriptsize{``New York City''}\\

        \includegraphics[width=\ww,frame]{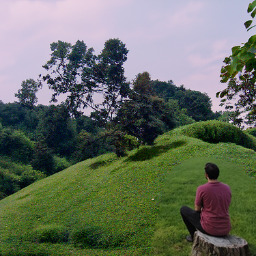} &
        \includegraphics[width=\ww,frame]{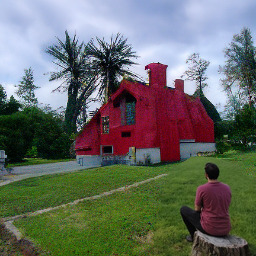} & 
        \includegraphics[width=\ww,frame]{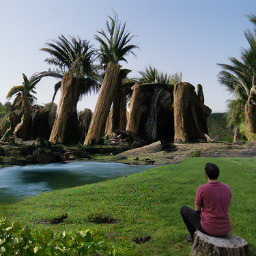} &
        \includegraphics[width=\ww,frame]{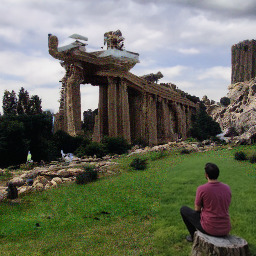} &
        \includegraphics[width=\ww,frame]{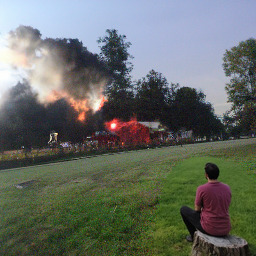} &
        \includegraphics[width=\ww,frame]{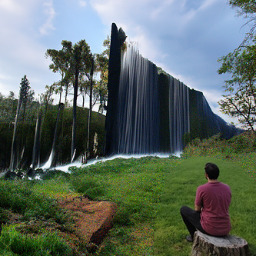} \\

        \scriptsize{``green hills''} &
        \scriptsize{``red house''} & 
        \scriptsize{``oasis''} &
        \scriptsize{``Acropolis''} &
        \scriptsize{``fire''} &
        \scriptsize{``big waterfall''} \\

        \includegraphics[width=\ww,frame]{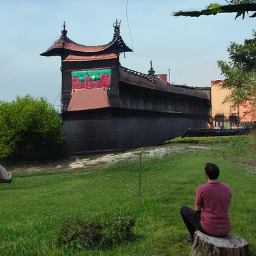} &
        \includegraphics[width=\ww,frame]{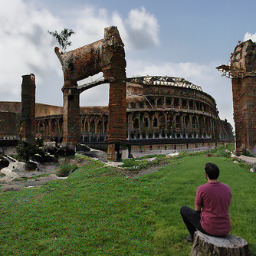} &
        \includegraphics[width=\ww,frame]{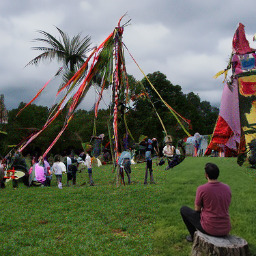} &
        \includegraphics[width=\ww,frame]{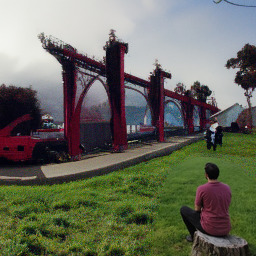} &
        \includegraphics[width=\ww,frame]{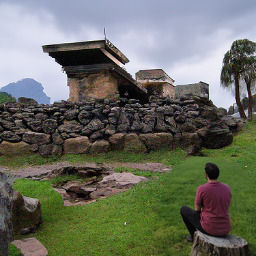} &
        \includegraphics[width=\ww,frame]{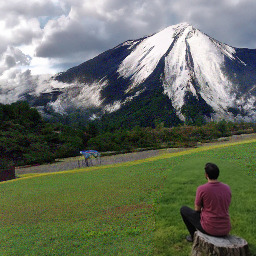} \\

        \scriptsize{``China''} &
        \scriptsize{``Colosseum''} &
        \scriptsize{``festival''} &
        \scriptsize{``Golden Gate Bridge''} & 
        \scriptsize{``Machu Picchu''} &
        \scriptsize{``Mount Fuji''}\\

        \includegraphics[width=\ww,frame]{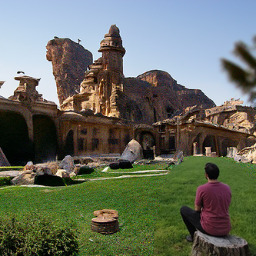} &
        \includegraphics[width=\ww,frame]{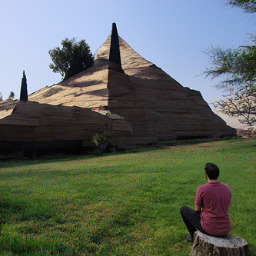} &
        \includegraphics[width=\ww,frame]{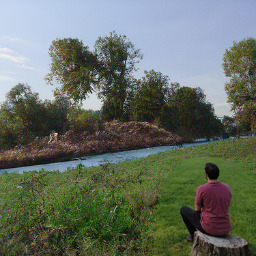} &
        \includegraphics[width=\ww,frame]{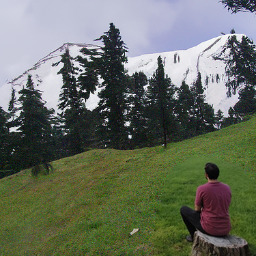} &
        \includegraphics[width=\ww,frame]{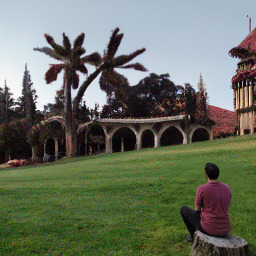} &
        \includegraphics[width=\ww,frame]{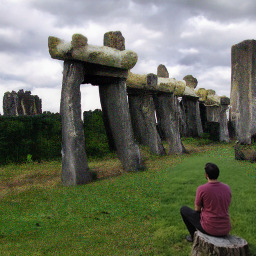} \\

        \scriptsize{``Petra''} &
        \scriptsize{``The Great Pyramid of Giza''} &
        \scriptsize{``river''} &
        \scriptsize{``snowy mountain''} &
        \scriptsize{``Stanford University''} & 
        \scriptsize{``Stonehenge''} \\

        \includegraphics[width=\ww,frame]{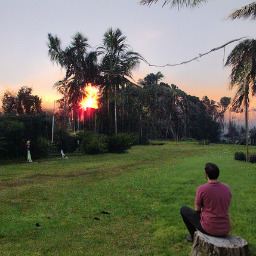} &
        \includegraphics[width=\ww,frame]{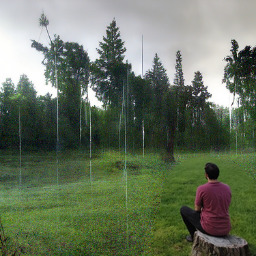} &
        \includegraphics[width=\ww,frame]{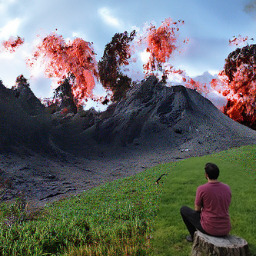} &
        \includegraphics[width=\ww,frame]{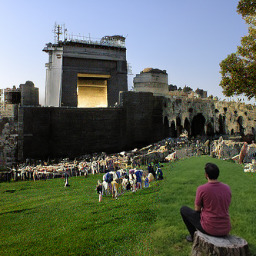} & 
        \includegraphics[width=\ww,frame]{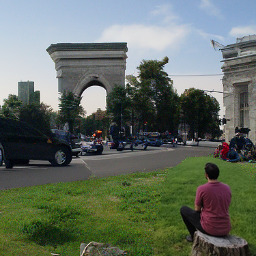} &
        \includegraphics[width=\ww,frame]{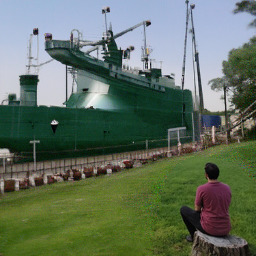} \\

        \scriptsize{``sunrise''} &
        \scriptsize{``rainy''} &
        \scriptsize{``volcanic eruption''} &
        \scriptsize{``The Western Wall''} & 
        \scriptsize{``Arc de Triomphe''} &
        \scriptsize{``big ship''} \\
    \end{tabular}
    
    \caption{\textbf{Background replacement:}  Given a source image and a mask of the background, the model is able to replace the background according to the text description. Note that the famous landmarks are not meant to accurately appear in the new background, but serve as an inspiration for the image completion.}
    \label{fig:background_editing_yossi_grass}
\end{figure*}

\begin{figure*}[ht]
    \centering
    \setlength{\tabcolsep}{1pt}
    \renewcommand{\arraystretch}{0.5}
    \setlength{\ww}{0.33\columnwidth}
  
    \begin{tabular}{cccccc}
        \includegraphics[width=\ww,frame]{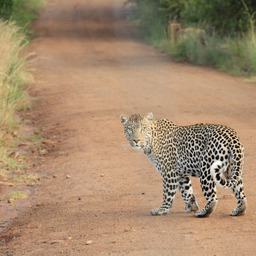} &
        \includegraphics[width=\ww,frame]{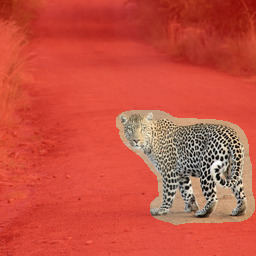} &
        \includegraphics[width=\ww,frame]{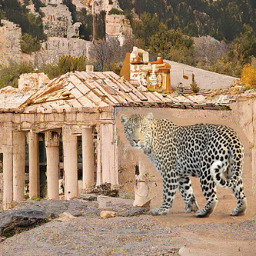} &
        \includegraphics[width=\ww,frame]{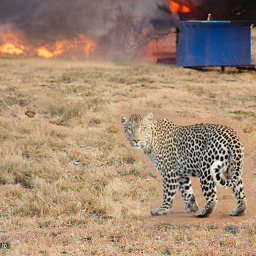} &
        \includegraphics[width=\ww,frame]{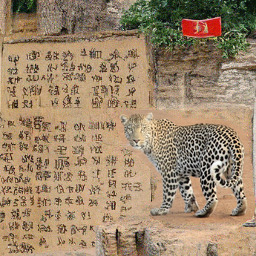} &
        \includegraphics[width=\ww,frame]{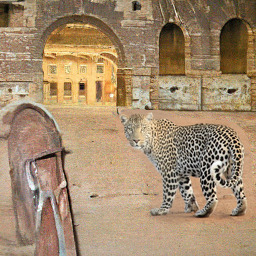} 
        \\
        
        \scriptsize{Input image} & 
        \scriptsize{Input mask} & 
        \scriptsize{``Acropolis''} &
        \scriptsize{``fire''} &
        \scriptsize{``China''} &
        \scriptsize{``Colosseum''} 
        \\

        \includegraphics[width=\ww,frame]{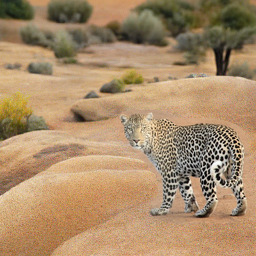} &
        \includegraphics[width=\ww,frame]{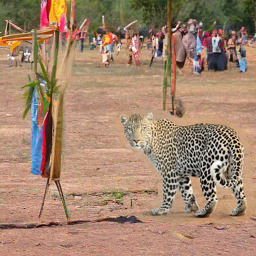} &
        \includegraphics[width=\ww,frame]{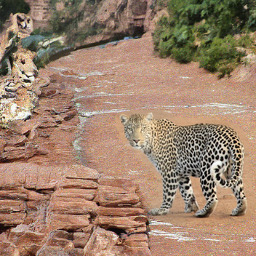} &
        \includegraphics[width=\ww,frame]{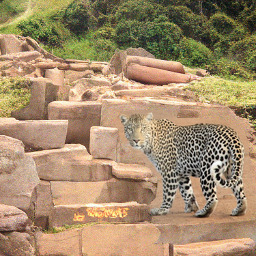} &
        \includegraphics[width=\ww,frame]{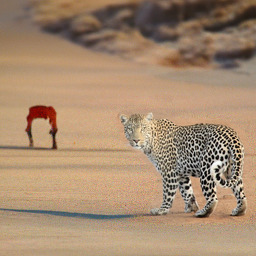} &
        \includegraphics[width=\ww,frame]{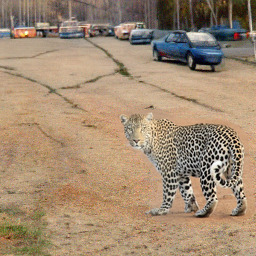}
        \\

        \scriptsize{``desert''} &
        \scriptsize{``festival''} &
        \scriptsize{``Grand Canyon''} &
        \scriptsize{``Machu Pichu''} &
        \scriptsize{``north pole''} &
        \scriptsize{``parking lot''}
        \\

        \includegraphics[width=\ww,frame]{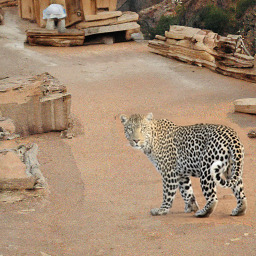} &
        \includegraphics[width=\ww,frame]{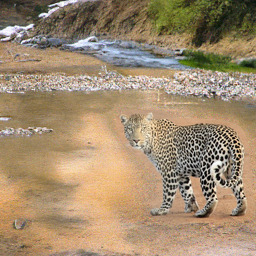} &
        \includegraphics[width=\ww,frame]{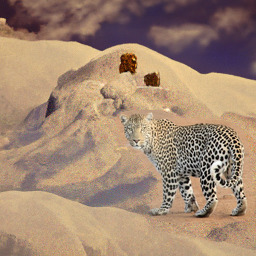} &
        \includegraphics[width=\ww,frame]{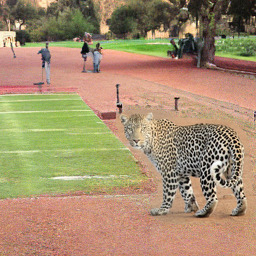} &
        \includegraphics[width=\ww,frame]{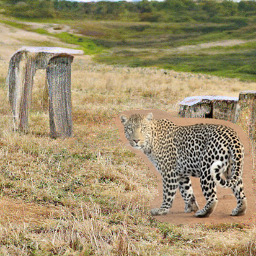} &
        \includegraphics[width=\ww,frame]{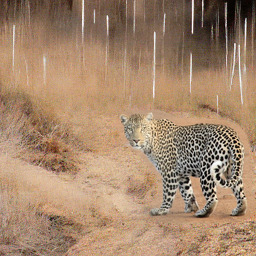}
        \\

        \scriptsize{``Petra''} &
        \scriptsize{``river''} &
        \scriptsize{``snowy mountain''} &
        \scriptsize{``Stanford University''} &
        \scriptsize{``Stonehenge''} &
        \scriptsize{``rainy''}
        \\

        \includegraphics[width=\ww,frame]{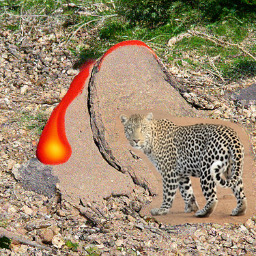} & 
        \includegraphics[width=\ww,frame]{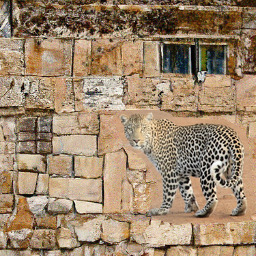} & 
        \includegraphics[width=\ww,frame]{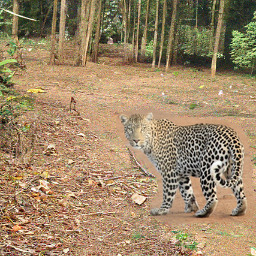} & 
        \includegraphics[width=\ww,frame]{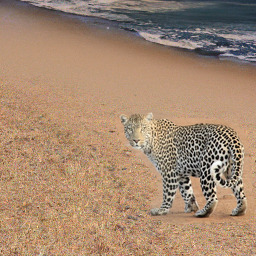} & 
        \includegraphics[width=\ww,frame]{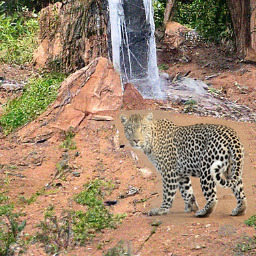} & 
        \includegraphics[width=\ww,frame]{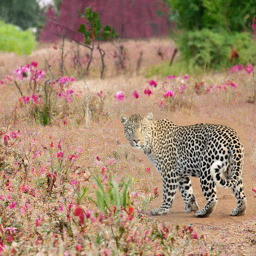}
        \\

        \scriptsize{``volcanic eruption''} &
        \scriptsize{``The Western Wall''} &
        \scriptsize{``in the woods''} &
        \scriptsize{``beach''} &
        \scriptsize{``big waterfall''} &
        \scriptsize{``flower field''}
        \\

        \includegraphics[width=\ww,frame]{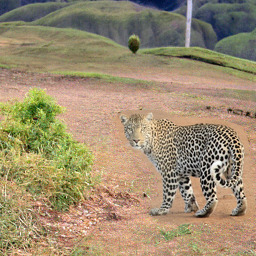} &
        \includegraphics[width=\ww,frame]{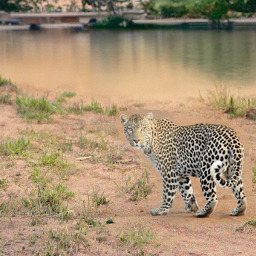} &
        \includegraphics[width=\ww,frame]{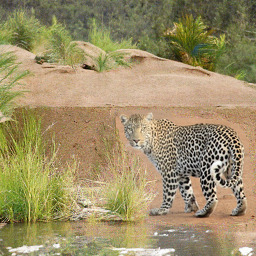} &
        \includegraphics[width=\ww,frame]{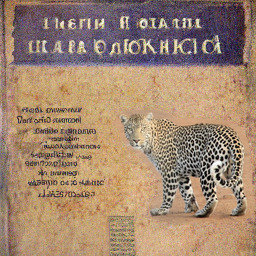} &
        \includegraphics[width=\ww,frame]{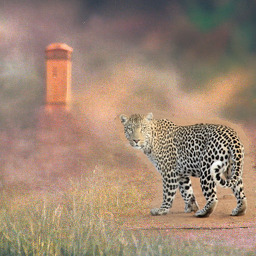} &
        \includegraphics[width=\ww,frame]{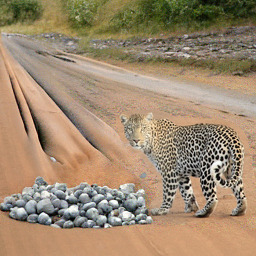}
        \\

        \scriptsize{``green hills''} &
        \scriptsize{``lake''} &
        \scriptsize{``oasis''} &
        \scriptsize{``book cover''} &
        \scriptsize{``fog''} &
        \scriptsize{``gravel''}
        \\

    \end{tabular}
    
    \caption{\textbf{Background replacement:}  Given a source image and a mask of the background, the model is able to replace the background corresponding to the text description. Note that the famous landmarks are not meant to accurately appear in the new background, but serve as an inspiration for the image completion.}
    \label{fig:background_editing_leopard}
\end{figure*}

\begin{figure*}[t]
    \centering
    \setlength{\tabcolsep}{0.5pt}
    \renewcommand{\arraystretch}{0.5}
    \setlength{\ww}{0.33\columnwidth}
  
    \begin{tabular}{cccccc}
        \rotatebox{90}{\phantom{AAA}``leopard''}
        \includegraphics[width=\ww,frame]{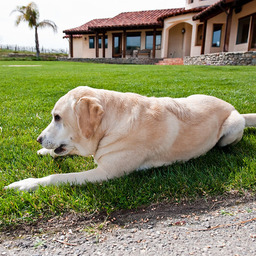} &
        \includegraphics[width=\ww,frame]{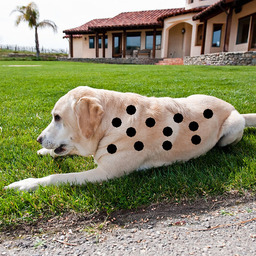} &
        \includegraphics[width=\ww,frame]{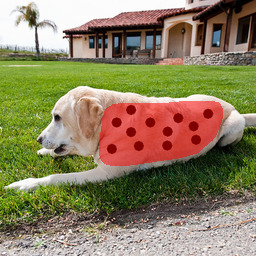} &
        \includegraphics[width=\ww,frame]{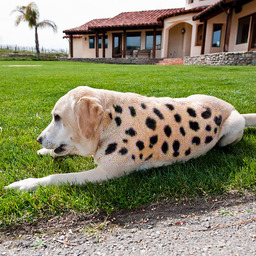} &
        \includegraphics[width=\ww,frame]{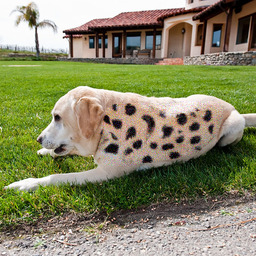} &
        \includegraphics[width=\ww,frame]{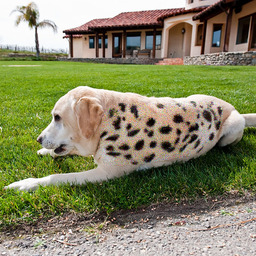}
        \\

        \rotatebox{90}{\phantom{AA}``straw chair''}
        \includegraphics[width=\ww,frame]{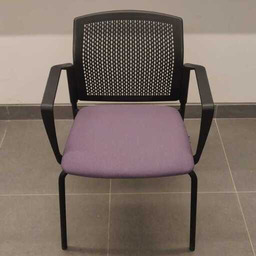} &
        \includegraphics[width=\ww,frame]{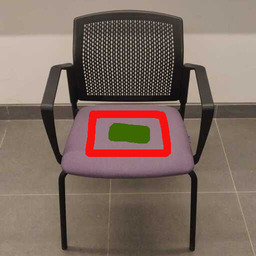} &
        \includegraphics[width=\ww,frame]{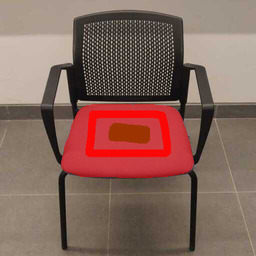} &
        \includegraphics[width=\ww,frame]{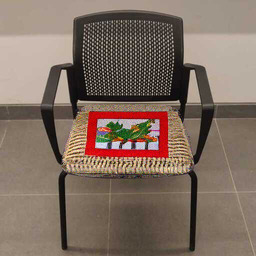} &
        \includegraphics[width=\ww,frame]{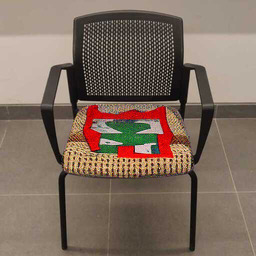} &
        \includegraphics[width=\ww,frame]{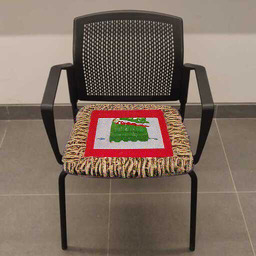}
        \\

        \rotatebox{90}{\phantom{AA}``floral carpet''}
        \includegraphics[width=\ww,frame]{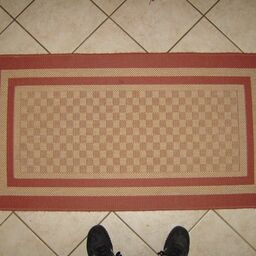} &
        \includegraphics[width=\ww,frame]{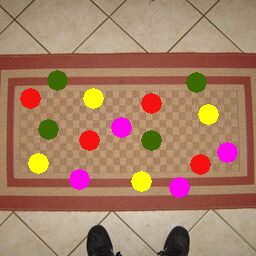} &
        \includegraphics[width=\ww,frame]{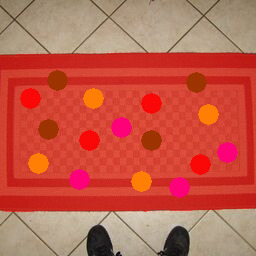} &
        \includegraphics[width=\ww,frame]{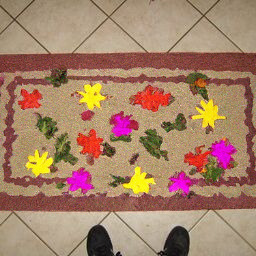} &
        \includegraphics[width=\ww,frame]{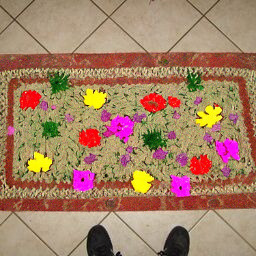} &
        \includegraphics[width=\ww,frame]{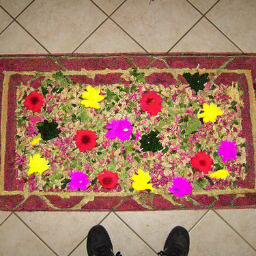}
        \\

        \rotatebox{90}{\phantom{AAA}``graffiti''}
        \includegraphics[width=\ww,frame]{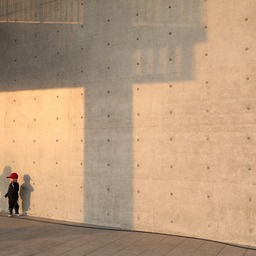} &
        \includegraphics[width=\ww,frame]{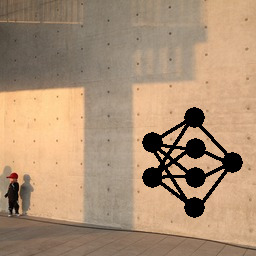} &
        \includegraphics[width=\ww,frame]{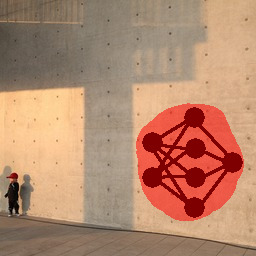} &
        \includegraphics[width=\ww,frame]{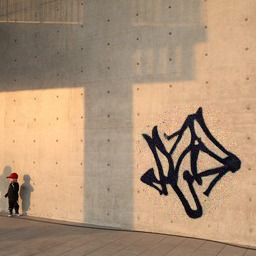} &
        \includegraphics[width=\ww,frame]{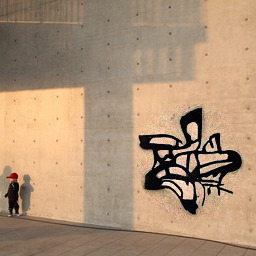} &
        \includegraphics[width=\ww,frame]{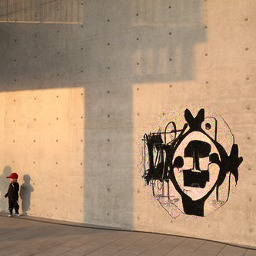}
        \\

        \rotatebox{90}{\phantom{AAA}``table''}
        \includegraphics[width=\ww,frame]{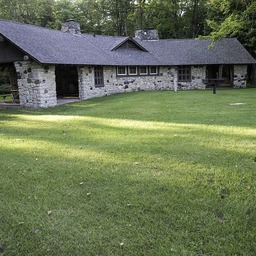} &
        \includegraphics[width=\ww,frame]{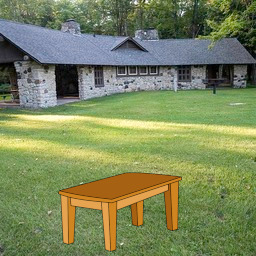} &
        \includegraphics[width=\ww,frame]{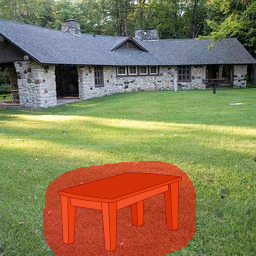} &
        \includegraphics[width=\ww,frame]{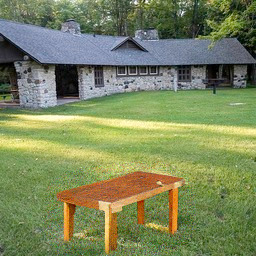} &
        \includegraphics[width=\ww,frame]{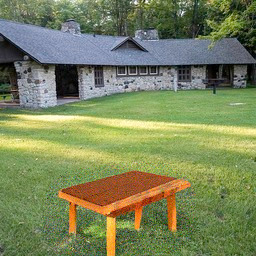} &
        \includegraphics[width=\ww,frame]{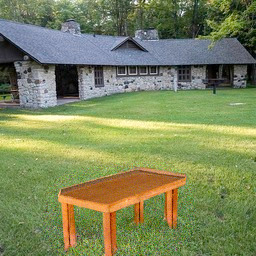}
        \\
        
        Input image &
        Image + scribble &
        Mask &
        Result 1 &
        Result 2 &
        Result 3 
        \\
    \end{tabular}
    
    \caption{\textbf{Scribble-guided editing:} Users scribble a rough shape of the object they want to insert, mark the edited area, and provide a guiding text. The model uses the scribble as a general shape and color reference, transforming it to match the guiding text. Note that the scribble patterns can also change. In the last example, we embedded a clip art of a table instead of a manual scribble, it shows the effectiveness of our model to transform unnatural clip arts into real-looking objects.}
    \label{fig:scribble_editing_additional_examples}
\end{figure*}

\begin{figure*}[ht]
    \centering
    \setlength{\tabcolsep}{0.5pt}
    \renewcommand{\arraystretch}{0.5}
    
    \centering
    \includegraphics[width=\textwidth]{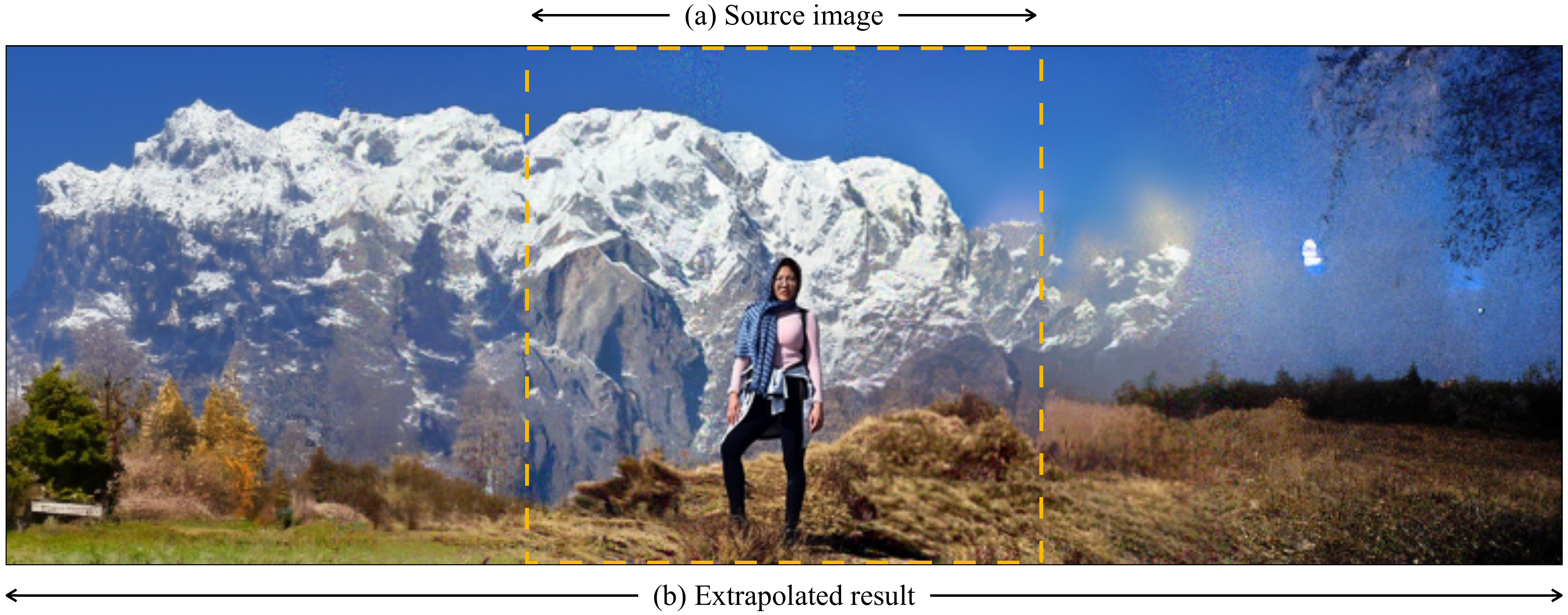}

    \caption{\textbf{Text-guided image extrapolation:} The user provides an image and two text descriptions that guide the extrapolation to the left (``sunny day'' in this example) and to the right (``dark night'').}
    \label{fig:image_extrapolation_himalaya}
\end{figure*}

We provide additional examples for the applications mentioned in the paper: \Cref{fig:leopard_multiple_results,fig:foreground_add_dogs,fig:foreground_leopard1_add} demonstrate the ability of our method to add new objects to an existing image, where \Cref{fig:leopard_multiple_results,fig:foreground_add_dogs} show that different results can be obtained for the same text prompt, while \Cref{fig:foreground_leopard1_add} shows results obtained using a variety of prompts. \Cref{fig:foreground_remove_dogs} demonstrates the ability to remove or replace objects in an exiting image, while \Cref{fig:foreground_alter_dogs} demonstrates the ability to alter an existing object in an image. \Cref{fig:background_editing_yossi_grass,fig:background_editing_leopard} demonstrate the ability to replace the background of an image. \Cref{fig:scribble_editing_additional_examples} demonstrates more examples of scribble-guided editing, and \Cref{fig:image_extrapolation_himalaya} demonstrates text-guided image extrapolation.



\subsection{Iterative Editing}
The synthesis results that are given by our method 
are at times exactly what the user envisioned, but they can also be different from the user's intent or might include unwanted artifacts.
Unlike other text-driven image editing techniques that operate 
on the entire image (e.g., StyleCLIP \cite{patashnik2021styleclip}),
our method is region-based, thus allowing the user to progressively refine their result in an \emph{incremental} editing session.

\begin{figure*}[t]
    \centering
    \setlength{\tabcolsep}{0.5pt}
    \renewcommand{\arraystretch}{0.5}
    \setlength{\ww}{0.4\columnwidth}
  
    \begin{tabular}{cccccc}
        \includegraphics[width=\ww,frame]{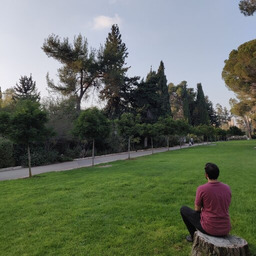} &
        \includegraphics[width=\ww,frame]{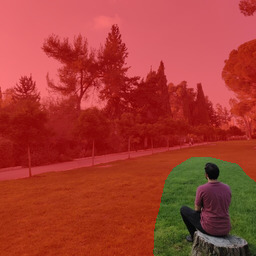} &
        \includegraphics[width=\ww,frame]{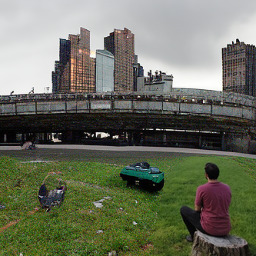} &
        \includegraphics[width=\ww,frame]{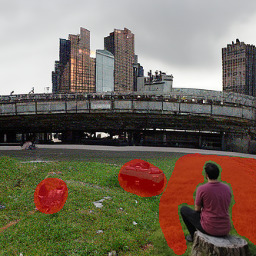} &
        \includegraphics[width=\ww,frame]{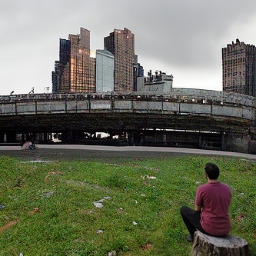} &
        \\
        
        Input image &
        Input mask &
        First result &
        Refinement mask &
        Final result
        \\
    \end{tabular}
    
    \caption{\textbf{Result refinement:} The initial synthesis result of our model 
    can be further refined. For example, here the user first masks a rough area in the source image and replaces the background using the prompt ``New York City''. Next, they wish to remove two unwanted objects from the generated result and to further refine the rough mask that was used in the first stage. They provide additional masks and no guiding text in this case (to perform inpainting) in order to obtain the final result.}
    \label{fig:editing_session_background_refinement}
\end{figure*}

\Cref{fig:editing_session_background_refinement} demonstrates such an editing session. At first, the user starts by replacing the background, as described in \Cref{sec:applications} in the main paper,
and obtains a result that is mostly satisfactory, but is not perfect: there are two unwanted generated objects on the grass that the user wishes to remove. In addition, the user decides that the initial mask used in the previous step was not accurate enough, causing a mismatch between the generated grass and the grass from the original scene. The user then provides additional masks, without a text prompt, causing our method to inpaint these areas, yielding the final result. 

\begin{figure*}[t]
    \centering
    \setlength{\tabcolsep}{1pt}
    \renewcommand{\arraystretch}{1}
    \setlength{\ww}{0.55\columnwidth}
  
    \begin{tabular}{ccc}
        \rotatebox{90}{\phantom{AA} Step 1: ``curly blond hair''}
        \includegraphics[width=\ww,frame]{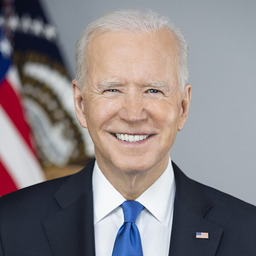} &
        \includegraphics[width=\ww,frame]{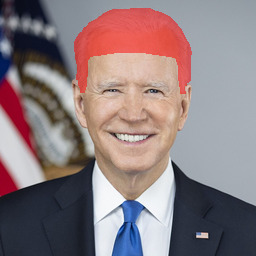} &
        \includegraphics[width=\ww,frame]{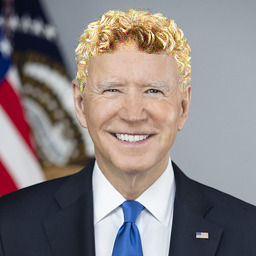}
        \\

        \rotatebox{90}{\phantom{AA} Step 2: ``shiny purple tie''}
        \includegraphics[width=\ww,frame]{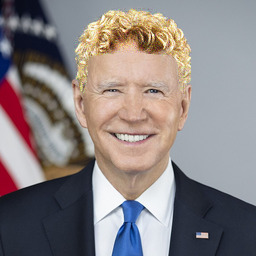} &
        \includegraphics[width=\ww,frame]{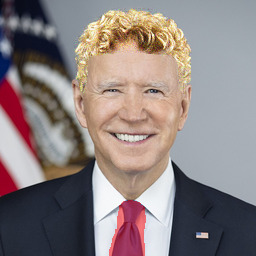} &
        \includegraphics[width=\ww,frame]{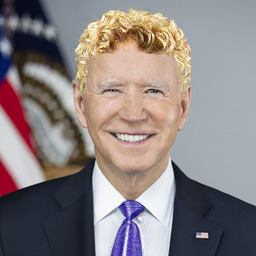}
        \\

        \rotatebox{90}{Step 3: scribble + ``floral jacket''}
        \includegraphics[width=\ww,frame]{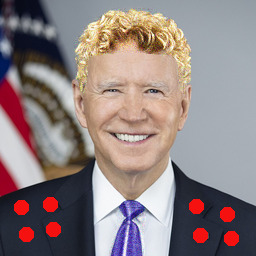} &
        \includegraphics[width=\ww,frame]{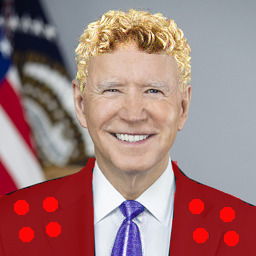} &
        \includegraphics[width=\ww,frame]{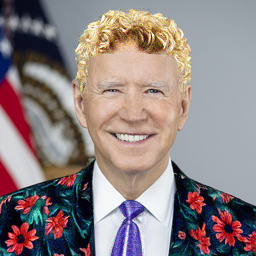}
        \\
        
        Input image &
        Mask &
        Result
        \\
    \end{tabular}
    
    \caption{\textbf{Editing session mix example:} The user can use several editing operations consecutively. For example, as the first step, the user masks the hair of the person and provides the guiding text ``curly blond hair''. As the second step, the user masks the tie and provides the guiding text ``shiny purple tie''. At the last step, the user scribbles red dots on the jacket, masks the jacket, and provides the guiding text ``floral jacket''.}
    \label{fig:mix_biden}
\end{figure*}

\begin{figure*}[t]
    \centering
    \setlength{\tabcolsep}{0.5pt}
    \renewcommand{\arraystretch}{0.5}
    \setlength{\ww}{0.55\columnwidth}
  
    \begin{tabular}{ccc}
        \rotatebox{90}{ Step 1: paste clip art + ``table''}
        \includegraphics[width=\ww,frame]{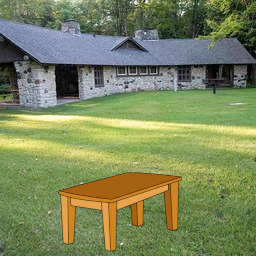} &
        \includegraphics[width=\ww,frame]{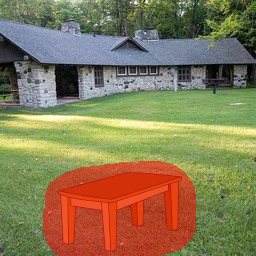} &
        \includegraphics[width=\ww,frame]{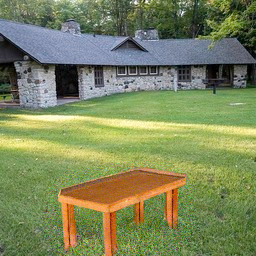}
        \\

        \rotatebox{90}{\phantom{AAAAA} Step 2: ``orange''}
        \includegraphics[width=\ww,frame]{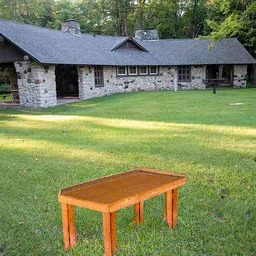} &
        \includegraphics[width=\ww,frame]{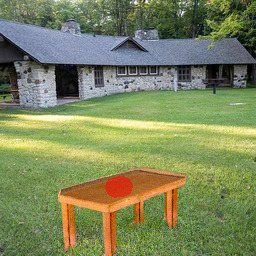} &
        \includegraphics[width=\ww,frame]{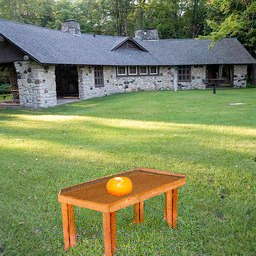}
        \\

        \rotatebox{90}{\phantom{AAA} Step 3: ``river bank''}
        \includegraphics[width=\ww,frame]{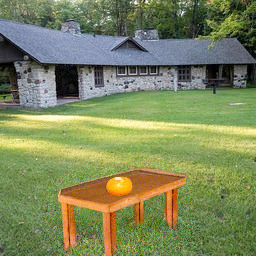} &
        \includegraphics[width=\ww,frame]{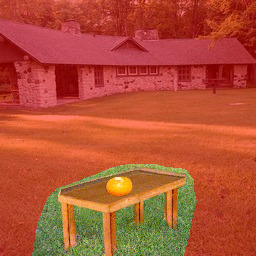} &
        \includegraphics[width=\ww,frame]{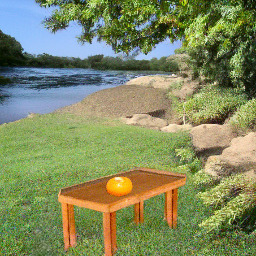}
        \\
        
        Input image &
        Mask &
        Result
        \\
    \end{tabular}
    
    \caption{\textbf{Editing session mix example:} The user can use several editing operations consecutively. For example, here the user starts by pasting a clip art of a table on the image, then masks the relevant area and provides the guiding text ``table'' to get a more natural looking table. In the second stage, the user masks an area on the previous synthesis result and provides the guiding text ``orange''. In the last stage, the user masks the background of the previous synthesis result and provides the guiding text ``river bank'' to get the final synthesis result.}
    \label{fig:mix_table}
\end{figure*}

\begin{figure*}[t]
    \centering
    \setlength{\tabcolsep}{0.5pt}
    \renewcommand{\arraystretch}{0.5}
    \setlength{\ww}{0.55\columnwidth}
  
    \begin{tabular}{ccc}
        \rotatebox{90}{\phantom{AAAAA} Step 1: ``dresser''}
        \includegraphics[width=\ww,frame]{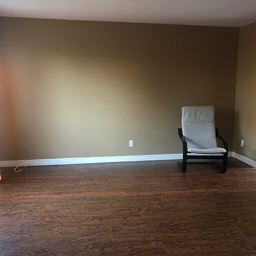} &
        \includegraphics[width=\ww,frame]{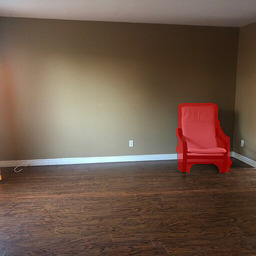} &
        \includegraphics[width=\ww,frame]{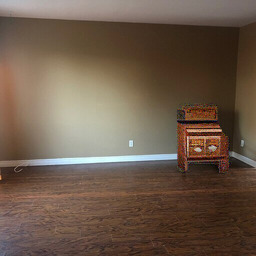}
        \\

        \rotatebox{90}{Step 2: scribble + ``ceiling lamp''}
        \includegraphics[width=\ww,frame]{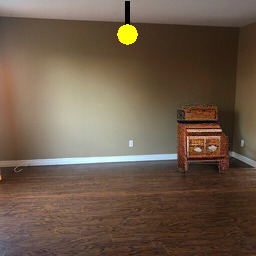} &
        \includegraphics[width=\ww,frame]{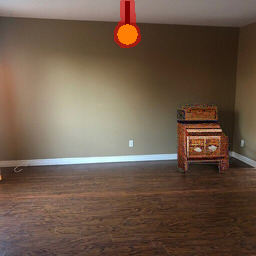} &
        \includegraphics[width=\ww,frame]{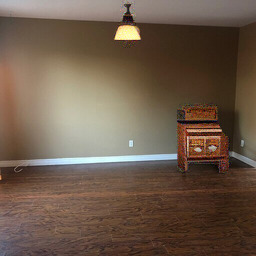}
        \\

        \rotatebox{90}{\phantom{AAAAA} Step 3: ``window''}
        \includegraphics[width=\ww,frame]{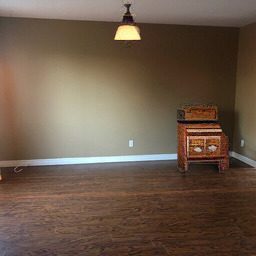} &
        \includegraphics[width=\ww,frame]{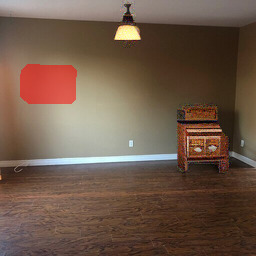} &
        \includegraphics[width=\ww,frame]{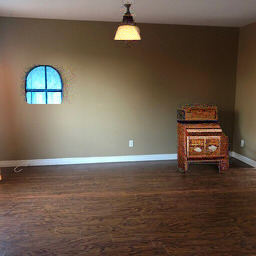}
        \\
        
        Input image &
        Mask &
        Result
        \\
    \end{tabular}
    
    \caption{\textbf{Editing session mix example:} The user can use several editing operations consecutively. As a first step, the user masks the chair and provides the guiding text ``dresser''. Next, the user scribbles a rough shape of a lamp on the result of the previous step, masks the area of the lamp, and provides the guiding text ``ceiling lamp''. Finally, the user masks an area over the wall in the previous result, and provides the guiding text ``window'' to obtain the final result.}
    \label{fig:mix_living_room}
\end{figure*}

\Cref{fig:mix_biden,fig:mix_table,fig:mix_living_room} demonstrate more editing sessions. Each of the sessions utilizes a variety of editing types: adding, changing and removing objects and backgrounds, scribble-guided edits, and clip-art-guided edits. Out method is compositional by design, and does not require any modifications to support such mixed editing sessions.

Unless stated otherwise, all the results in the main paper and in this supplemental document are \emph{without} such incremental refinements --- we show the raw results with no further user interaction.


\subsection{Failure Cases}
\begin{figure*}[t]
    \centering
    \setlength{\tabcolsep}{1pt}
    \renewcommand{\arraystretch}{1}
    \setlength{\ww}{0.5\columnwidth}
  
    \begin{tabular}{ccc}
        \rotatebox{90}{\phantom{AAAAA} ``rubber toy''}
        \includegraphics[width=\ww,frame]{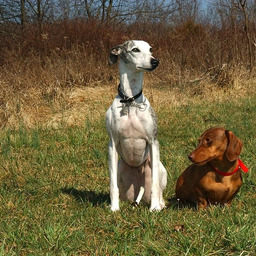} &
        \includegraphics[width=\ww,frame]{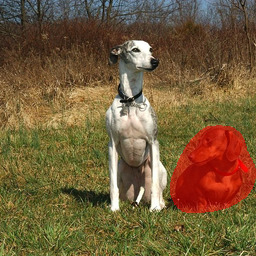} &
        \includegraphics[width=\ww,frame]{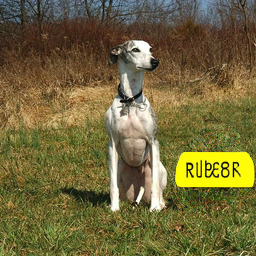}
        \\

        \rotatebox{90}{\phantom{AAAAA} ``Colosseum''}
        \includegraphics[width=\ww,frame]{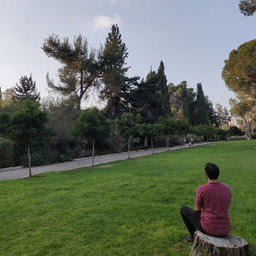} &
        \includegraphics[width=\ww,frame]{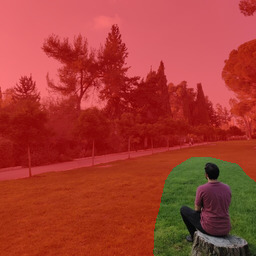} &
        \includegraphics[width=\ww,frame]{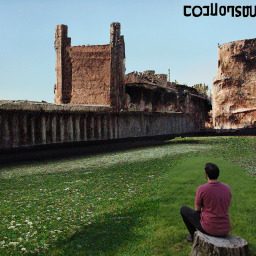}
        \\

        \rotatebox{90}{\phantom{AAAAA} ``Acropolis''}
        \includegraphics[width=\ww,frame]{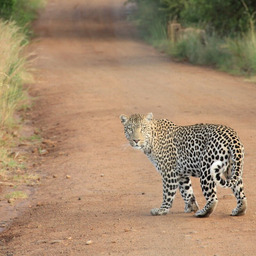} &
        \includegraphics[width=\ww,frame]{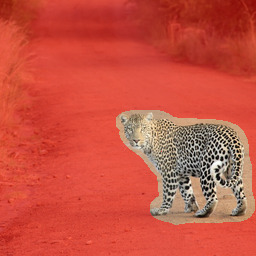} &
        \includegraphics[width=\ww,frame]{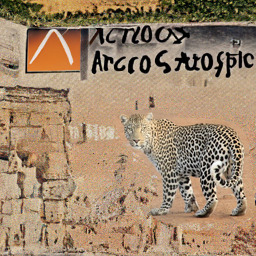}
        \\

        \rotatebox{90}{\phantom{AAAAA} ``rubber toy''}
        \includegraphics[width=\ww,frame]{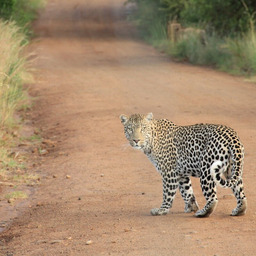} &
        \includegraphics[width=\ww,frame]{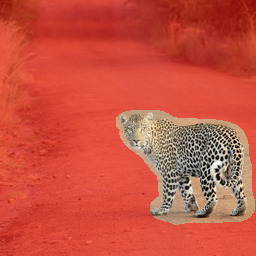} &
        \includegraphics[width=\ww,frame]{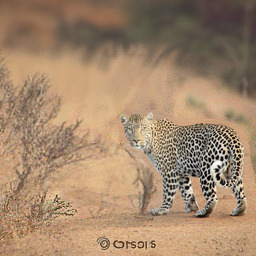}
        \\
        
        Input image &
        Mask &
        Result
        \\
    \end{tabular}
    
    \caption{\textbf{Typographic failure:} Our model inherits CLIP \cite{radford2021learning} susceptibility to typographic attacks \cite{goh2021multimodal}. Instead of generating an object or a scene, the model might generate a textual description.}
    \label{fig:fail_case_typographic_failure}
\end{figure*}

\begin{figure*}[ht]
    \centering
    \setlength{\tabcolsep}{0.5pt}
    \renewcommand{\arraystretch}{0.5}
    \setlength{\ww}{0.5\columnwidth}
  
    \begin{tabular}{ccc}
        \includegraphics[width=\ww,frame]{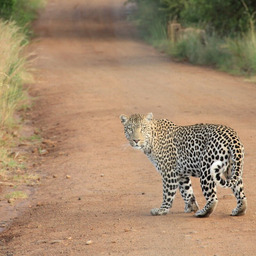} &
        \includegraphics[width=\ww,frame]{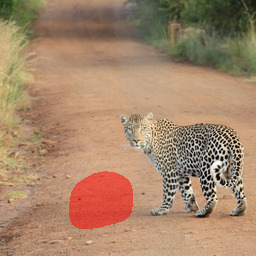} &
        \includegraphics[width=\ww,frame]{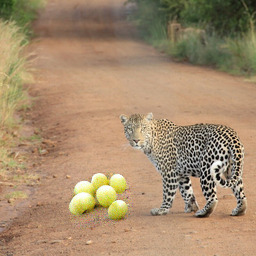}
        \\
        
        Input image &
        Mask &
        Synthesis result
        \\
    \end{tabular}
    
    \caption{\textbf{Out of proportion synthesis:} We show a failure case in which our method generates objects the look natural by themselves, but with the wrong proportion to the rest of the scene. For the guiding text ``grapes'', the synthesized result contains grapes which are huge compared to the leopard and to the rest of the scene.}
    \label{fig:fail_case_unproportional_synthesis}
\end{figure*}

\Cref{fig:fail_case_typographic_failure} demonstrates the susceptibility of our model to typographic attacks \cite{goh2021multimodal}. \Cref{fig:fail_case_unproportional_synthesis} demonstrates synthesis of objects which appear natural on their own, but possess the wrong size compared to the rest of the photo.


\subsection{\Naive{} blending example}
\begin{figure*}[h]
    \centering
    \setlength{\tabcolsep}{0.5pt}
    \renewcommand{\arraystretch}{0.5}
    \setlength{\ww}{0.5\columnwidth}
  
    \begin{tabular}{cccc}
        \includegraphics[width=\ww,frame]{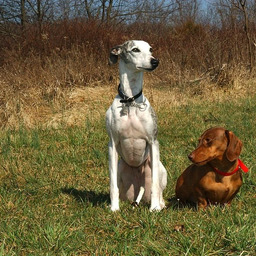} &
        \includegraphics[width=\ww,frame]{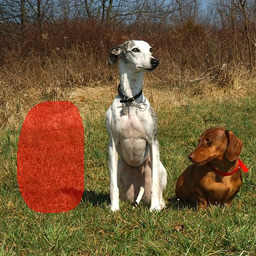} &
        \includegraphics[width=\ww,frame]{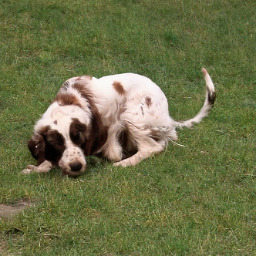} &
        \includegraphics[width=\ww,frame]{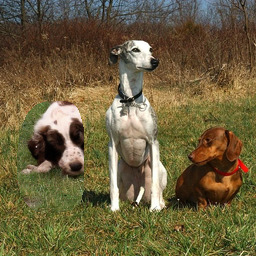} \\
        
        \scriptsize{Input image} & 
        \scriptsize{Input mask} & 
        \scriptsize{Initial prediction} & 
        \scriptsize{\Naive{} blend} \\
    \end{tabular}
    
    \caption{\textbf{\Naive{} Blending:} When providing the model the input image and mask with the text prompt ``a dog'', and without using the background preservation loss --- the result is a dog whose head is inside the mask, but most of the dog's body is outside the mask. Blending such a result with the input image using the input mask we obtain an unnatural result.}
    \label{fig:naive_blending}
\end{figure*}

As discussed in \Cref{sec:background_preservation_blending} of the paper,
\naive{} blending of the input image and the diffusion-synthesized result inside the masked area yields an unnatural result, as can be seen in \Cref{fig:naive_blending}.

\subsection{High-resolution generation}
\begin{figure*}[ht]
    \centering
    \setlength{\tabcolsep}{0.5pt}
    \renewcommand{\arraystretch}{0.5}
    \setlength{\ww}{0.16\textwidth}
  
    \begin{tabular}{cccccc}
        \includegraphics[width=\ww,frame]{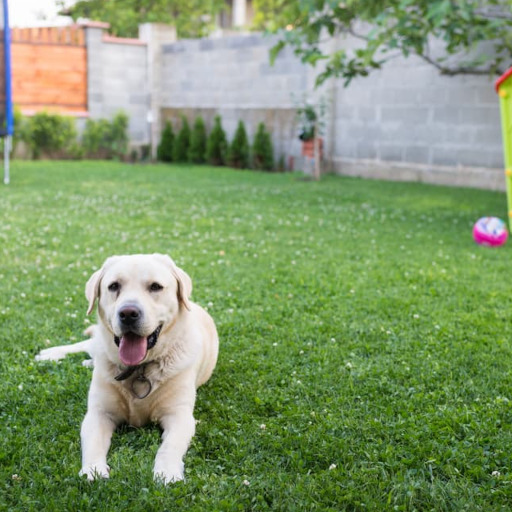} &
        \includegraphics[width=\ww,frame]{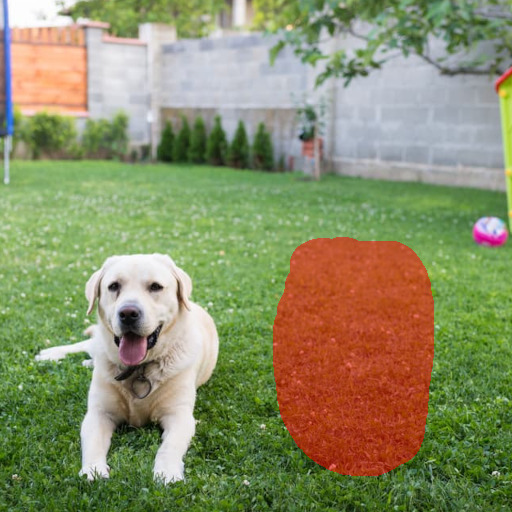} &
        \includegraphics[width=\ww,frame]{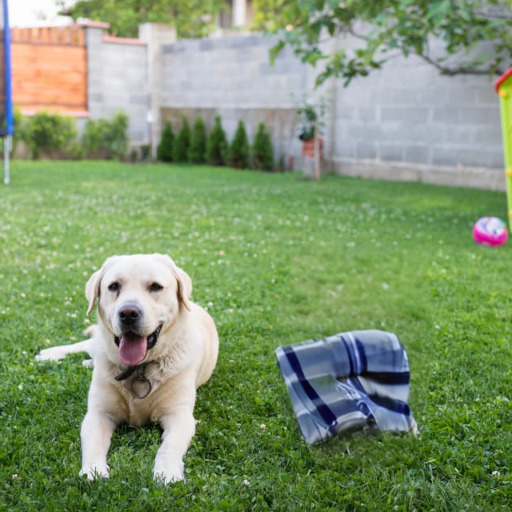} &
        \includegraphics[width=\ww,frame]{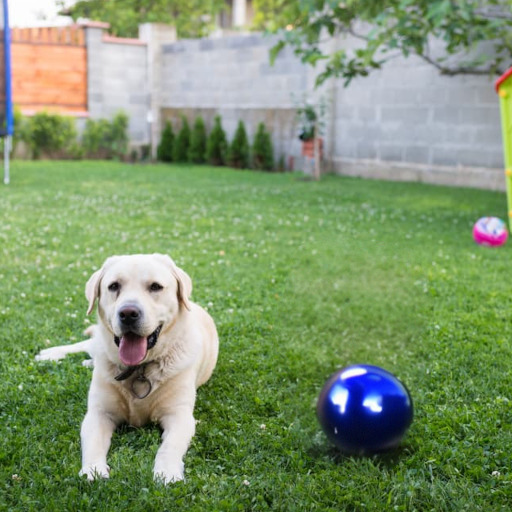} &
        \includegraphics[width=\ww,frame]{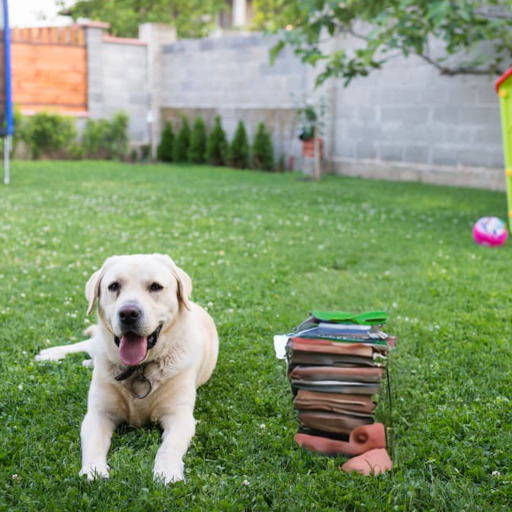} &
        \includegraphics[width=\ww,frame]{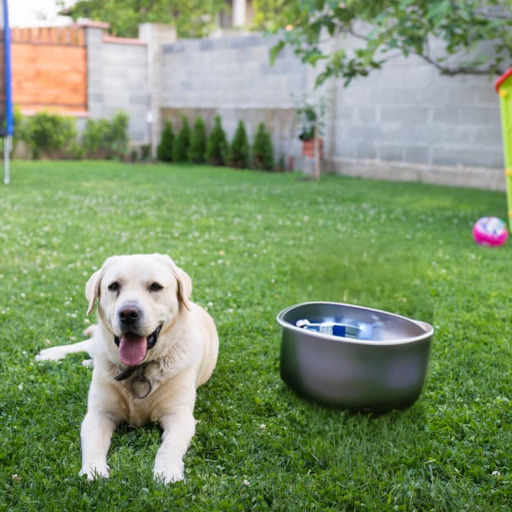}
        \\
        
        Input image & 
        Input mask & 
        ``blanket'' &
        ``blue ball'' &
        ``pile of books'' &
        ``bowl of water''
        \\

        \includegraphics[width=\ww,frame]{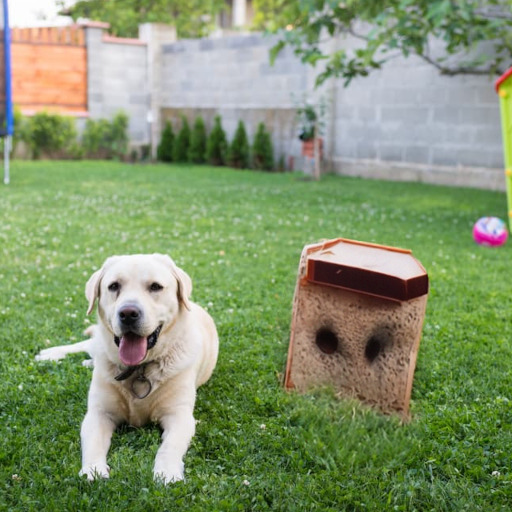} &
        \includegraphics[width=\ww,frame]{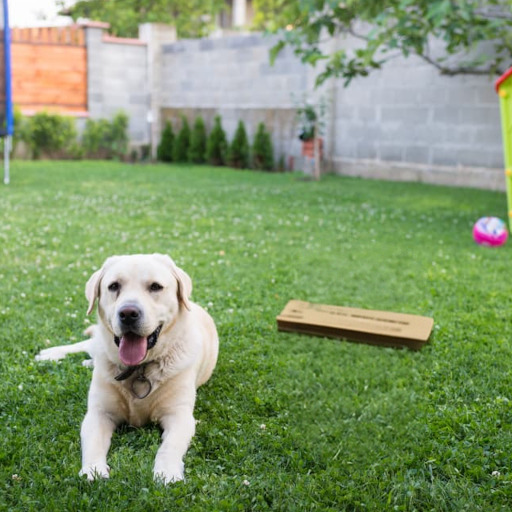} &
        \includegraphics[width=\ww,frame]{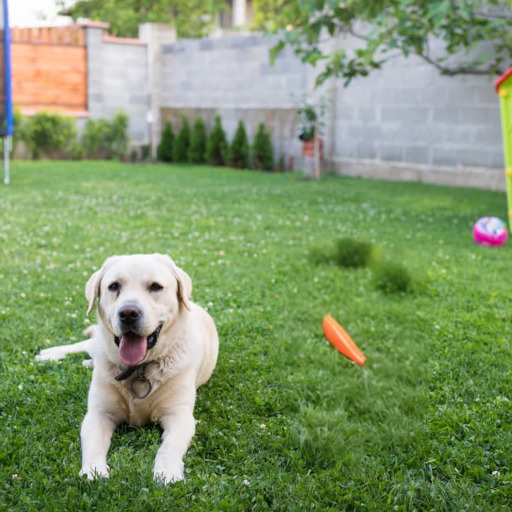} &
        \includegraphics[width=\ww,frame]{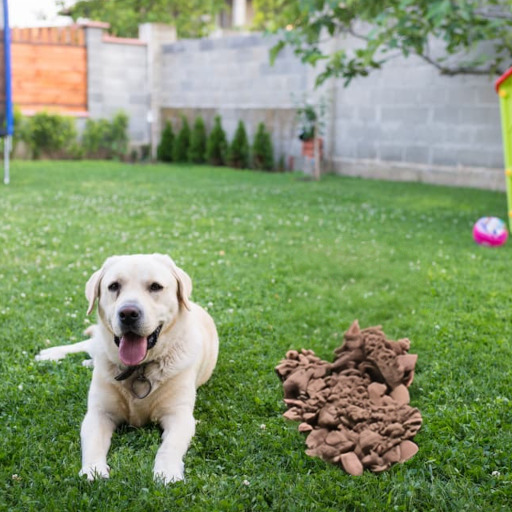} &
        \includegraphics[width=\ww,frame]{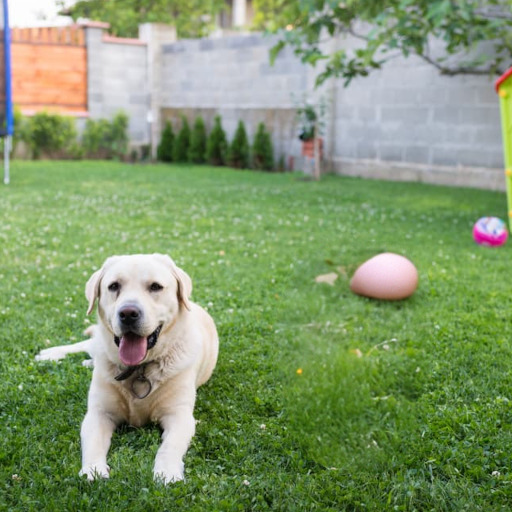} &
        \includegraphics[width=\ww,frame]{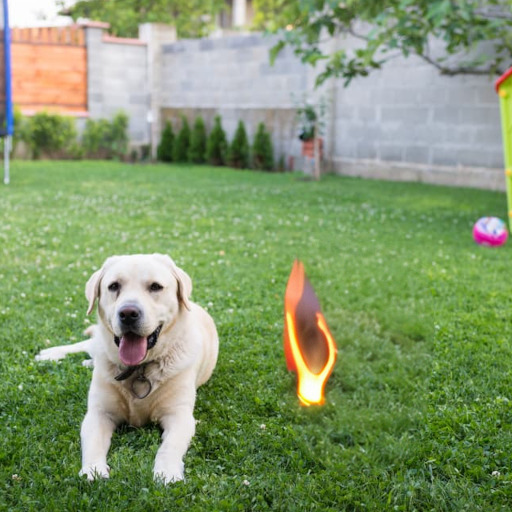}
        \\
        
        ``bread'' & 
        ``cardboard'' & 
        ``carrot'' &
        ``pile of dirt'' &
        ``egg'' &
        ``fire''
        \\

        \includegraphics[width=\ww,frame]{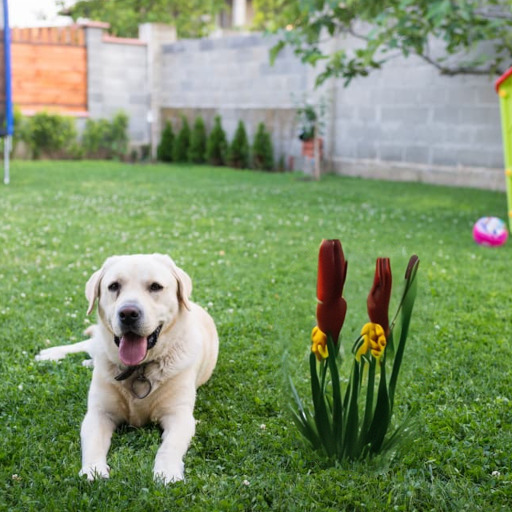} &
        \includegraphics[width=\ww,frame]{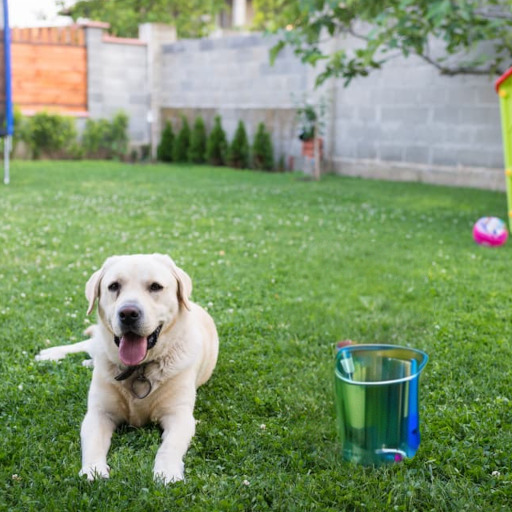} &
        \includegraphics[width=\ww,frame]{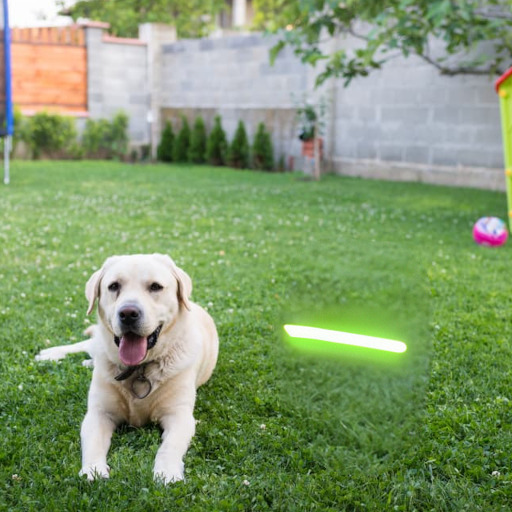} &
        \includegraphics[width=\ww,frame]{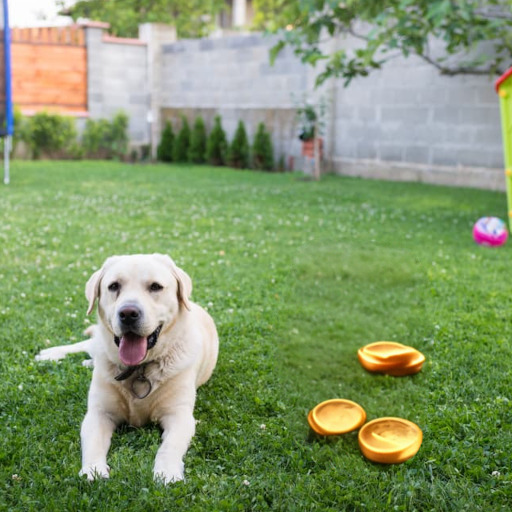} &
        \includegraphics[width=\ww,frame]{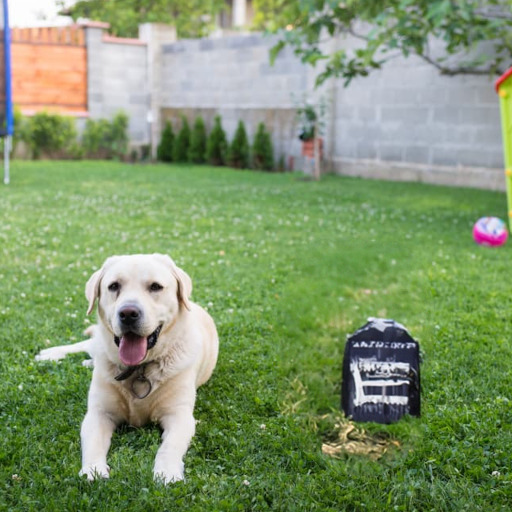} &
        \includegraphics[width=\ww,frame]{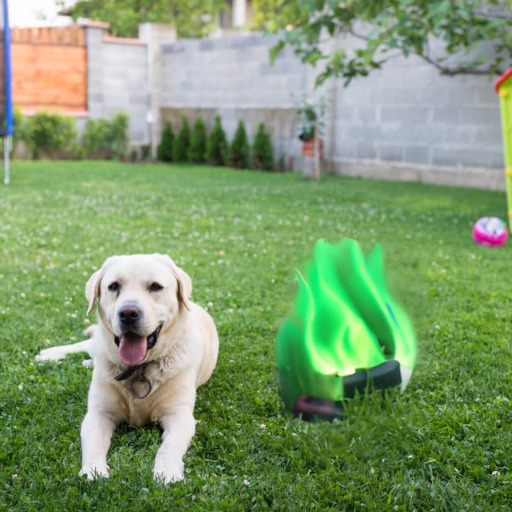}
        \\
        
        ``flower'' & 
        ``glass'' & 
        ``glow stick'' &
        ``golden coins'' &
        ``grave'' &
        ``green flame''
        \\

        \includegraphics[width=\ww,frame]{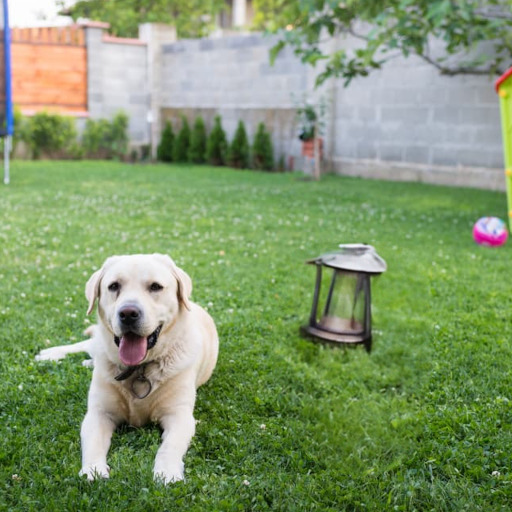} &
        \includegraphics[width=\ww,frame]{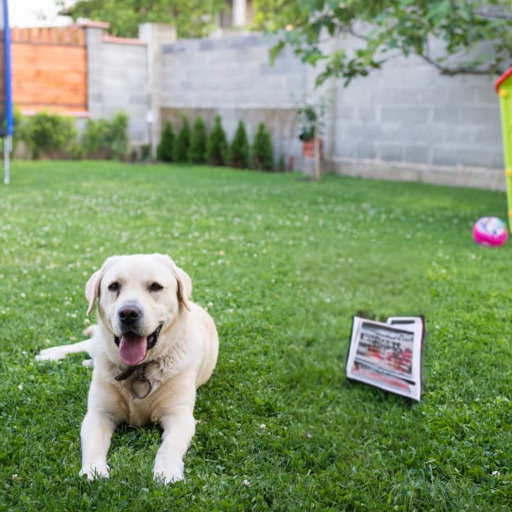} &
        \includegraphics[width=\ww,frame]{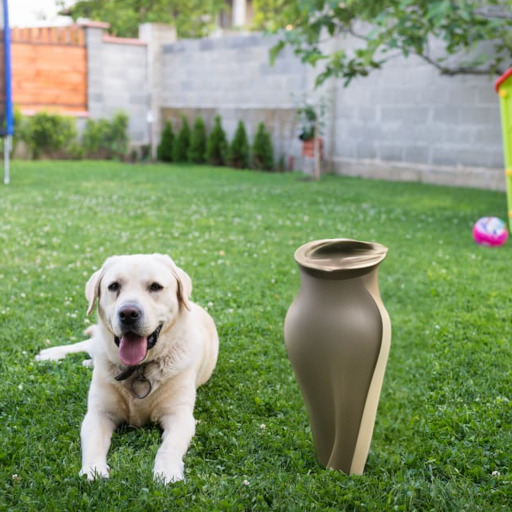} &
        \includegraphics[width=\ww,frame]{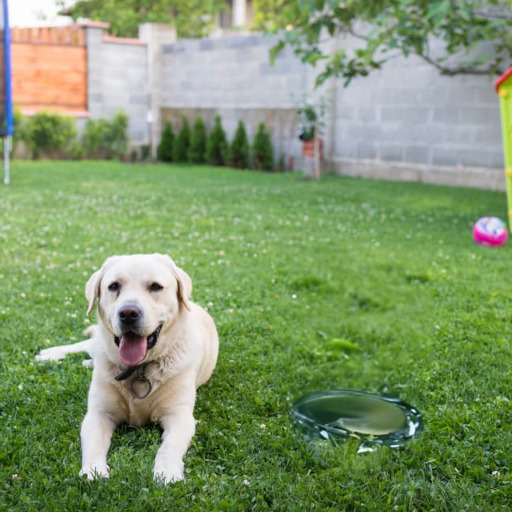} &
        \includegraphics[width=\ww,frame]{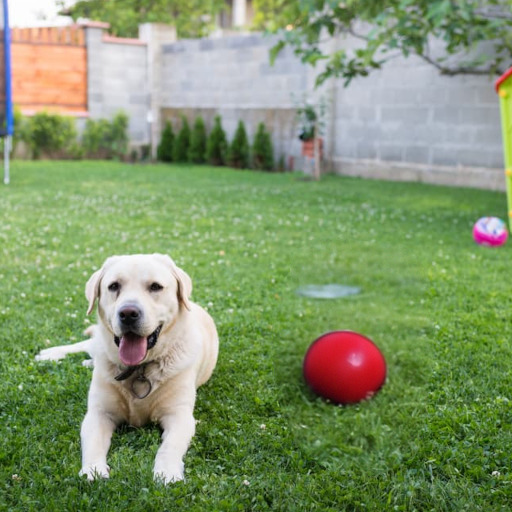} &
        \includegraphics[width=\ww,frame]{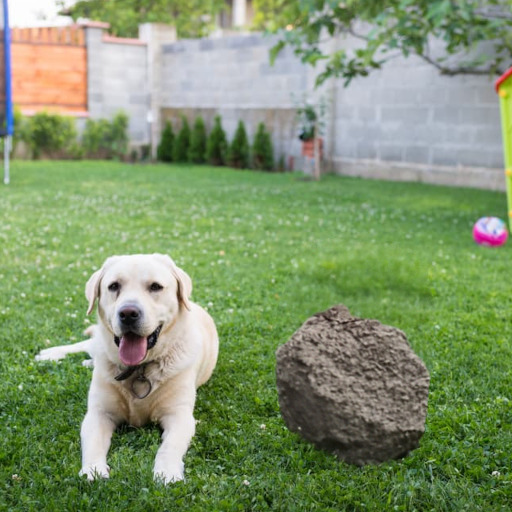}
        \\
        
        ``lamp'' & 
        ``newspaper'' & 
        ``clay pot'' &
        ``puddle'' &
        ``red ball'' &
        ``rock''
        \\

        \includegraphics[width=\ww,frame]{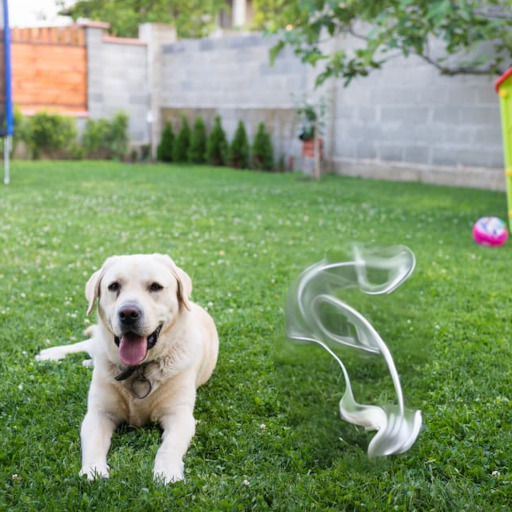} &
        \includegraphics[width=\ww,frame]{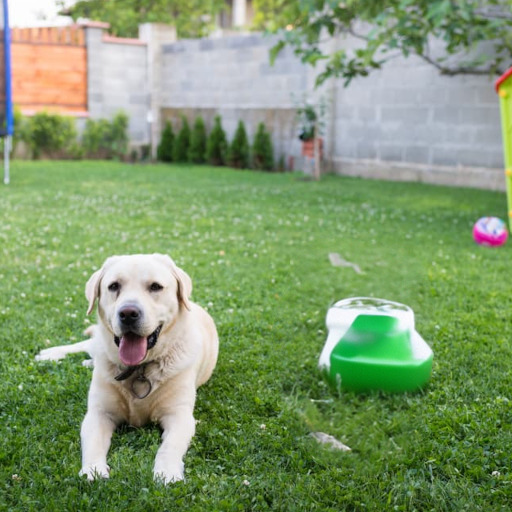} &
        \includegraphics[width=\ww,frame]{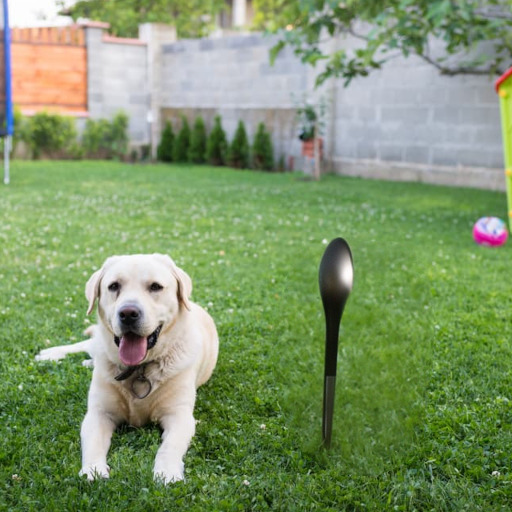} &
        \includegraphics[width=\ww,frame]{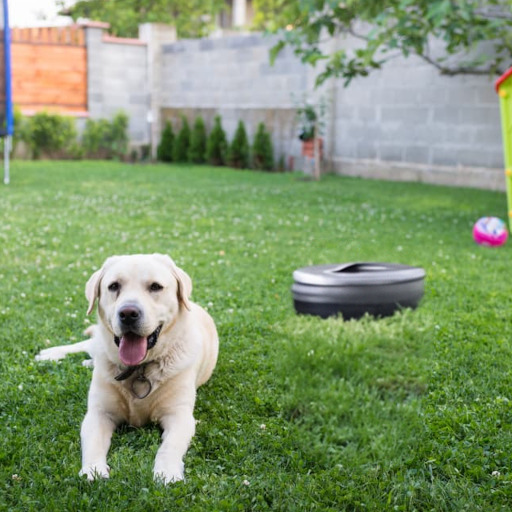} &
        \includegraphics[width=\ww,frame]{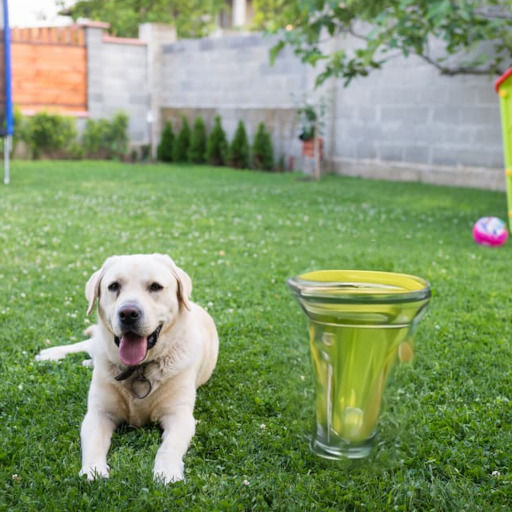} &
        \includegraphics[width=\ww,frame]{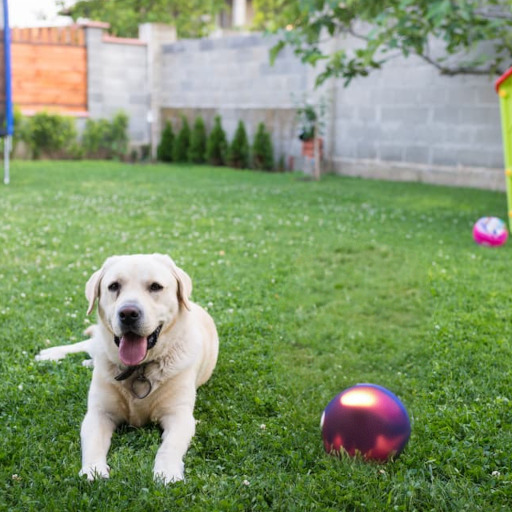}
        \\
        
        ``smoke'' & 
        ``soap'' & 
        ``spoon'' &
        ``car tire'' &
        ``vase'' &
        ``shiny ball''
        \\

    \end{tabular}
    
    \caption{\textbf{High resolution results:}  Given an input image of and mask, our model is able to generate different objects corresponding to different text descriptions. Results were produced using $512 \times 512$ DDPM model.}
    \label{fig:high_resolution_results}
\end{figure*}

Most results presented in the paper use an unconditional DDPM model of resolution $256 \times 256$, producing generated images of that resolution. 
Nevertheless, we are not constrained to this resolution, as can be seen in
\Cref{fig:image_extrapolation_heaven_hell} in the main paper
and in \Cref{fig:image_extrapolation_himalaya} in this supplementary document (for more details read \Cref{sec:tex-guided-extrapolation-details}).
We can also use OpenAI's unconditional $512 \times 512$ version of the model \cite{guided_diffusion_github}, by feeding the one-hot encoding with zeroes vector (similarly to \cite{clip_guided_diffusion}). Demonstration of using the higher resolution model for blended diffusion can be seen in \Cref{fig:high_resolution_results}.

\subsection{Comparison to DDIM}
\begin{figure*}[ht]
    \centering
    \setlength{\tabcolsep}{1pt}
    \renewcommand{\arraystretch}{0.5}
    \setlength{\ww}{0.33\columnwidth}
  
    \begin{tabular}{ccccccc}
        \rotatebox{90}{\phantom{AAA}{{DDPM}}} &
        \includegraphics[width=\ww,frame]{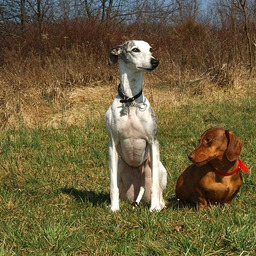} & 
        \includegraphics[width=\ww,frame]{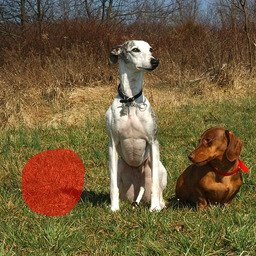} & 
        \includegraphics[width=\ww,frame]{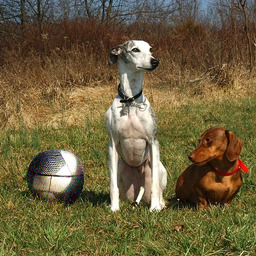} & 
        \includegraphics[width=\ww,frame]{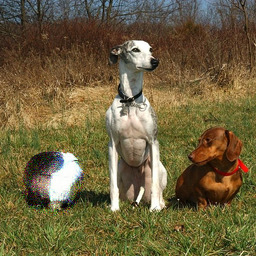} & 
        \includegraphics[width=\ww,frame]{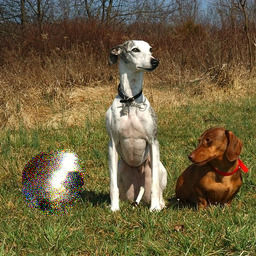} & 
        \includegraphics[width=\ww,frame]{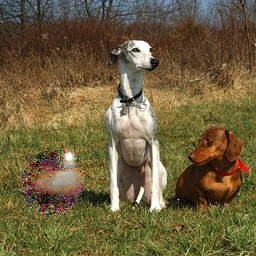}
        \\

        \rotatebox{90}{\phantom{AAA}{{DDIM}}} &
        \includegraphics[width=\ww,frame]{figures/ddpm_vs_ddim/assets/dogs/img.jpg} & 
        \includegraphics[width=\ww,frame]{figures/ddpm_vs_ddim/assets/dogs/mask_overlay.jpg} & 
        \includegraphics[width=\ww,frame]{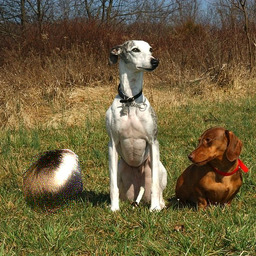} & 
        \includegraphics[width=\ww,frame]{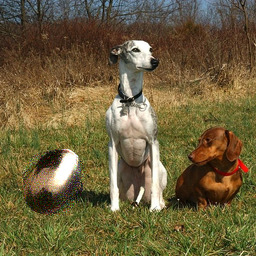} & 
        \includegraphics[width=\ww,frame]{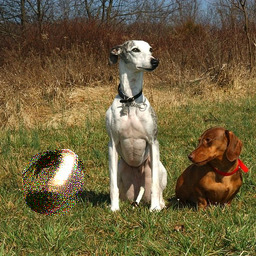} & 
        \includegraphics[width=\ww,frame]{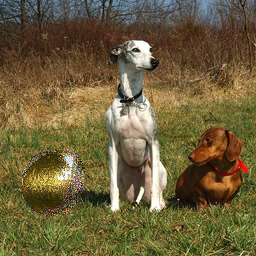}
        \\

        \midrule
        \\

        \rotatebox{90}{\phantom{AAA}{{DDPM}}} &
        \includegraphics[width=\ww,frame]{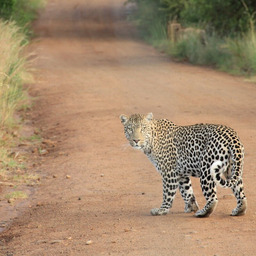} & 
        \includegraphics[width=\ww,frame]{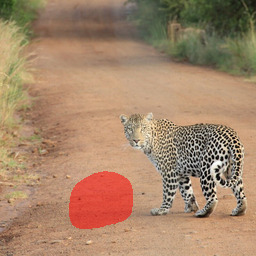} & 
        \includegraphics[width=\ww,frame]{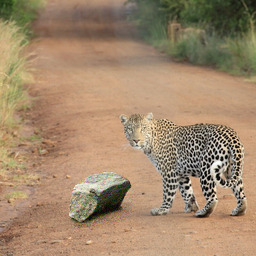} & 
        \includegraphics[width=\ww,frame]{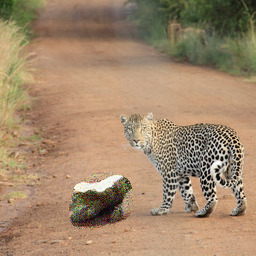} & 
        \includegraphics[width=\ww,frame]{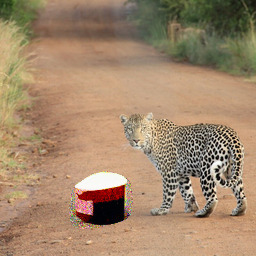} & 
        \includegraphics[width=\ww,frame]{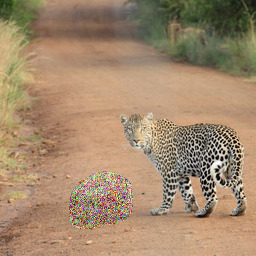}
        \\

        \rotatebox{90}{\phantom{AAA}{{DDIM}}} &
        \includegraphics[width=\ww,frame]{figures/ddpm_vs_ddim/assets/tiger/img.jpg} & 
        \includegraphics[width=\ww,frame]{figures/ddpm_vs_ddim/assets/tiger/mask_overlay.jpg} & 
        \includegraphics[width=\ww,frame]{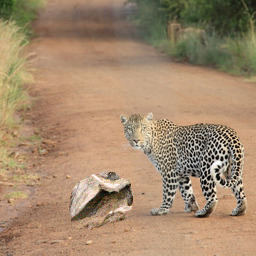} & 
        \includegraphics[width=\ww,frame]{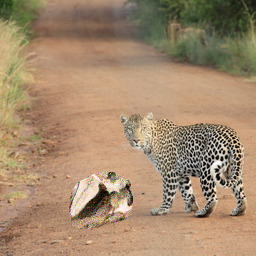} & 
        \includegraphics[width=\ww,frame]{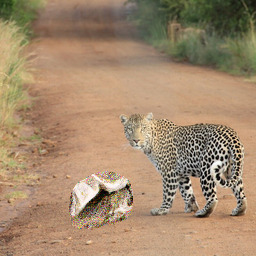} & 
        \includegraphics[width=\ww,frame]{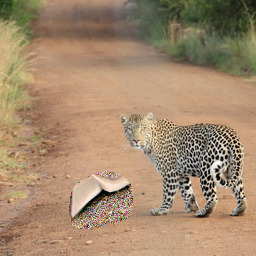}
        \\

        \midrule
        \\

        \rotatebox{90}{\phantom{AAA}{{DDPM}}} &
        \includegraphics[width=\ww,frame]{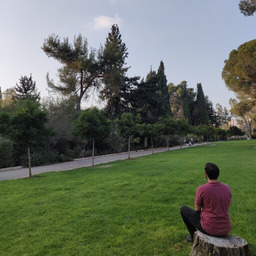} & 
        \includegraphics[width=\ww,frame]{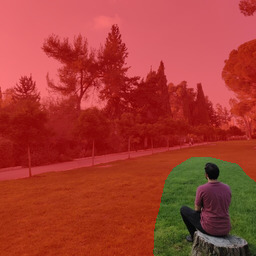} & 
        \includegraphics[width=\ww,frame]{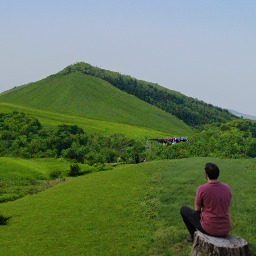} & 
        \includegraphics[width=\ww,frame]{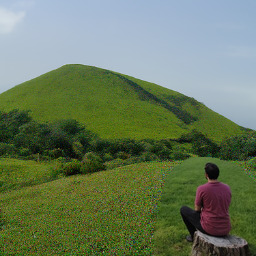} & 
        \includegraphics[width=\ww,frame]{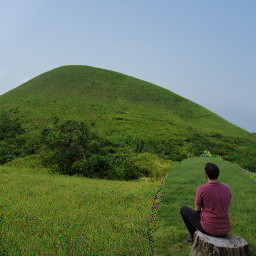} & 
        \includegraphics[width=\ww,frame]{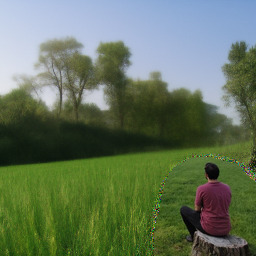}
        \\

        \rotatebox{90}{\phantom{AAA}{{DDIM}}} &
        \includegraphics[width=\ww,frame]{figures/ddpm_vs_ddim/assets/hills/img.jpg} & 
        \includegraphics[width=\ww,frame]{figures/ddpm_vs_ddim/assets/hills/mask_overlay.jpg} & 
        \includegraphics[width=\ww,frame]{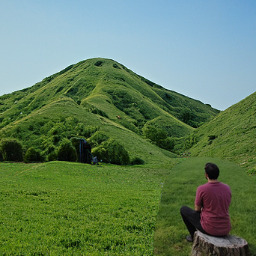} & 
        \includegraphics[width=\ww,frame]{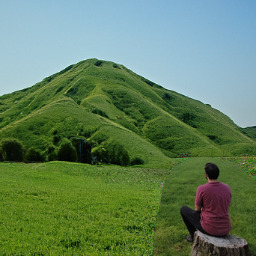} & 
        \includegraphics[width=\ww,frame]{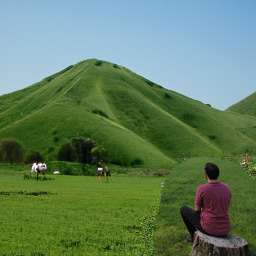} & 
        \includegraphics[width=\ww,frame]{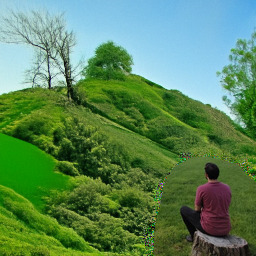}
        \\
        
        &
        Input image & 
        Input mask & 
        100 steps & 
        50 steps & 
        25 steps & 
        12 steps \\

    \end{tabular}
    
    \caption{\textbf{Blended Diffusion DDPM VS Blended Diffusion DDIM comparison:} The part corresponds to the editing text ``a shiny ball'', the middle part to ``a rock'' and the bottom part to ``green hills''. As we can see, DDPM produces better results when using 100 diffusion steps, whereas it produces worse results in less than 50 diffusion steps.}
    \label{fig:ddpm_vs_ddim}
\end{figure*}

Our method uses Denoising Diffusion Probabilistic Models (DDPMs). Recently, Song et al. propose  Denoising Diffusion Implicit Models (DDIMs) \cite{song2020denoising}, a fast
sampling algorithm for DDPMs that produces a new implicit model with the same marginal noise distributions, but deterministically maps noise to images. 
Nichol et al. \cite{nichol2021improved} showed that DDIMs produce better samples than DDPMs with fewer than 50 sampling steps, but worse samples when using 50 or more steps. In order to check the effect of using DDIM instead of DDPM we first adjusted the DDIM version of the guided-diffusion algorithm \cite{dhariwal2021diffusion} with Blended Diffusion in \Cref{alg:ddim_blended_diffusion}. As we can see experimentally in \Cref{fig:ddpm_vs_ddim}, the same holds for image generation using Blended Diffusion: DDPMs produce better results than DDIMs when using 100 diffusion steps, but worse results when using less than 50 diffusion steps.

\begin{algorithm}[t]
    \footnotesize
    \caption{DDIM blended diffusion: given a diffusion model $(\mu_{\theta}(x_t), \Sigma_{\theta}(x_t))$, and $\mclip$ model}
    \label{alg:ddim_blended_diffusion}
    \begin{algorithmic}
        \STATE \textbf{Input:} source image $x$, target text description $d$, input mask $m$, diffusion steps $k$, number of extending augmentations $N$
        \STATE \textbf{Output:} edited image $\widehat{x}$ that differs from input image $x$ inside area $m$ according to text description $d$ 
        \STATE $x_k \sim \mathcal{N}(\sqrt{\bar{\alpha}_k} x_0, (1-\bar{\alpha}_k) \mathbf{I})$
        \FORALL{$t$ from $k$ to 0}
            \STATE $\mu, \Sigma \gets \mu_{\theta}(x_t), \Sigma_{\theta}(x_t)$
            \STATE $\widehat{x}_0 \gets \frac{x_t}{\sqrt{\bar{\alpha}_t}} - \frac{\sqrt{1-\bar{\alpha}_t} \epsilon_{\theta}(x_t, t)}{\sqrt{\bar{\alpha}_t}}$
            \STATE $\widehat{x}_{0,{\textit{aug}}} \gets {\textit{ExtendingAugmentations}(\widehat{x}_0, N)}$
            \STATE $\nabla_{\textit{text}} \gets \frac{1}{N} \sum_{i=1}^{N} \nabla_{\widehat{x}_{0,{\textit{aug}}}} \mathcal{D}_{\textit{CLIP}}(\widehat{x}_{0,{\textit{aug}}}, d, m)$
            \STATE $\hat \epsilon \gets \epsilon_{\theta}(x_t) - \sqrt{1-\bar{\alpha}_t} \nabla_{\textit{text}}$
            \STATE $x_{\textit{fg}} \gets \sqrt{\bar{\alpha}_{t-1}} \left( \frac{x_t - \sqrt{1-\bar{\alpha}_t} \hat{\epsilon}}{\sqrt{\bar{\alpha}_t}} \right) + \sqrt{1-\bar{\alpha}_{t-1}} \hat{\epsilon}$
            \STATE $x_{\textit{bg}} \sim \mathcal{N}(\sqrt{\bar{\alpha}_t} x_0, (1-\bar{\alpha}_t) \mathbf{I})$
            \STATE $x_{t-1} \gets x_{\textit{fg}} \odot m + x_{\textit{bg}} \odot (1 - m)$
        \ENDFOR
        \RETURN $x_{-1}$
    \end{algorithmic}
\end{algorithm}
\section{Implementation Details}
\label{sec:implementation_details}
For all the experiments reported in this paper we used a pre-trained CLIP model \cite{radford2021learning} and a pre-trained guided-diffusion model \cite{dhariwal2021diffusion}:

\begin{itemize}
    \item For the CLIP model we used ViT-B/16 as a backbone for the Vision Transformer \cite{dosovitskiy2020image} that was released by OpenAI~\cite{clip_github}.
    \item For the diffusion model we used an unconditional model of resolution $256\times256$ \cite{guided_diffusion_github}.
\end{itemize}

Both of these models were released under MIT license and were developed using PyTorch \cite{paszke2019pytorch}.

All the input images in this paper are real images (i.e., not synthesized), except the ones in
\Cref{fig:comparison_paint_by_word_original} of the main paper,
which were generated by Bau et al. \cite{bau2021paint}.
All images were released freely under a Creative Commons license.

\subsection{Hyperparamters}
We used the CLIP model as-is, without changing any parameters. In addition, we did not utilize any prompt engineering techniques as described by Radford et al.~\cite{radford2021learning}.

We used the following hyperparameters in the guided-diffusion model across the different experiments (both in our model and in the baselines):
\begin{itemize}
    \item \textbf{Fast sampling speed:} We follow the fast sampling speed from \cite{nichol2021improved} which showed that 100 sampling steps are sufficient to achieve near-optimal FID score \cite{heusel2017gans} on ImageNet \cite{deng2009imagenet}. This scheme reduces the sampling time to 27 seconds, for more details see \Cref{sec:inference_time}.
    \item \textbf{Number of diffusion steps:} In most of our experiments we set the number of diffusion steps to $k=75$, allowing the model to change the input image in a sufficient manner. Exceptions are scribble-based editing ($k=60$) and background editing ($k=67$).
\end{itemize}

In \Cref{alg:final} we use the following hyperparameters:
\begin{itemize}
    \item \textbf{Number of extending augmentations:} We found that setting this to $N=16$ was sufficient to mitigate the adversarial example phenomena.
    \item \textbf{Number of total repetitions:} As explained in
            \Cref{sec:generation_ranking},
          we generate several results and rank them using the CLIP model. In our experiments, we generate 64 samples and choose the best ones. For more details on inference time see \Cref{sec:inference_time}.
\end{itemize}

\subsection{Extending Augmentations}
\begin{figure*}[ht]
    \centering
    \setlength{\tabcolsep}{0.5pt}
    \renewcommand{\arraystretch}{0.5}
    \setlength{\ww}{0.5\columnwidth}
  
    \begin{tabular}{cccc}
        \includegraphics[width=\ww,frame]{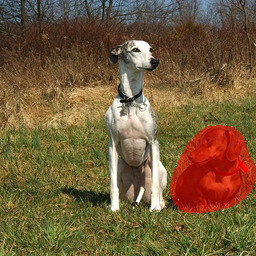} &
        \includegraphics[width=\ww,frame]{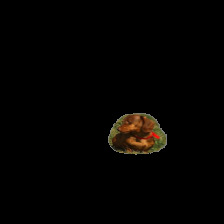} &
        \includegraphics[width=\ww,frame]{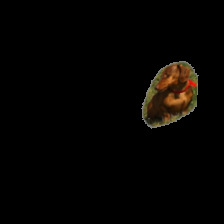} &
        \includegraphics[width=\ww,frame]{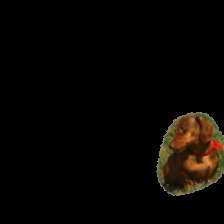} \\
        
        \scriptsize{Input + mask} & 
        \scriptsize{Augmentation 1} & 
        \scriptsize{Augmentation 2} & 
        \scriptsize{Augmentation 3} \\
    \end{tabular}
    
    \caption{\textbf{Extending augmentation example:} Given an input image and mask, we augment the masked area in the image using various projective transformations.}
    \label{fig:extending_augmentation_example}
\end{figure*}

Given an input image $x$, in the resolution of the diffusion model ($256 \times 256$ in our case), we first resize it to the input size of the CLIP model ($224 \times 224$) along with its input mask. Next, we create $N$ copies of this image and perform a different random projective transformation on each copy, along with the same transformation on the corresponding mask (see \Cref{fig:extending_augmentation_example}). Finally, we calculate the gradients using the CLIP loss w.r.t each one of the transformed copies and average all the gradients. This way, an adversarial manipulation is much less likely, as it would have to ``fool'' CLIP under multiple transformations.

\begin{figure*}[ht]
    \centering
    \setlength{\tabcolsep}{0.5pt}
    \renewcommand{\arraystretch}{0.5}
    \setlength{\ww}{0.39\columnwidth}
  
    \begin{tabular}{cccc}
        \rotatebox{90}{\phantom{AAA} ``big red pillow''}
        \includegraphics[width=\ww,frame]{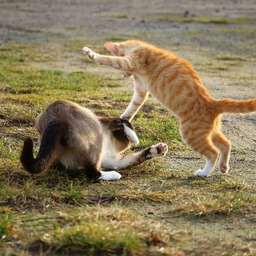} &
        \includegraphics[width=\ww,frame]{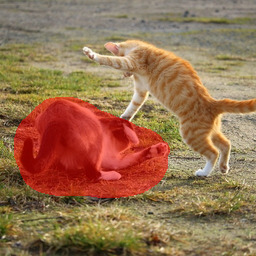} &
        \includegraphics[width=\ww,frame]{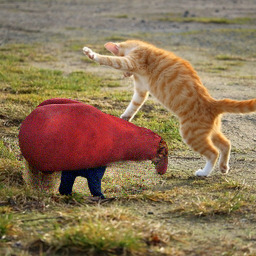} &
        \includegraphics[width=\ww,frame]{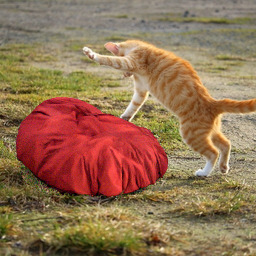} \\

        \rotatebox{90}{\phantom{Aa} ``happy dog mouth''}
        \includegraphics[width=\ww,frame]{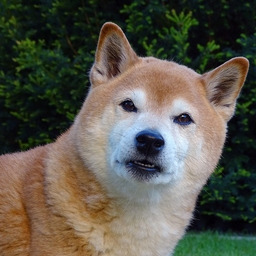} &
        \includegraphics[width=\ww,frame]{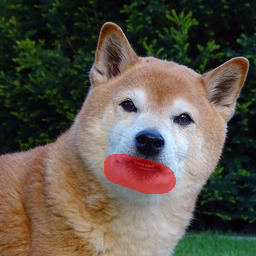} &
        \includegraphics[width=\ww,frame]{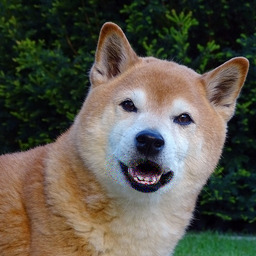} &
        \includegraphics[width=\ww,frame]{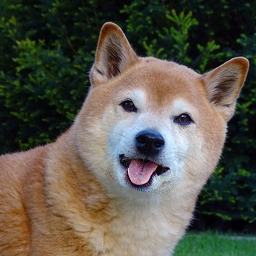} \\

        \rotatebox{90}{\phantom{AAAA} ``sausages''}
        \includegraphics[width=\ww,frame]{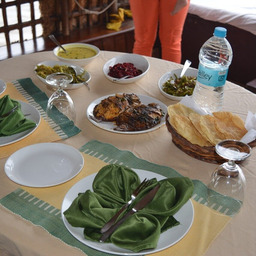} &
        \includegraphics[width=\ww,frame]{figures/ablation_study_augmentations/assets/mask_overlay3.jpg} &
        \includegraphics[width=\ww,frame]{figures/ablation_study_augmentations/assets/pred3_no_aug.jpg} &
        \includegraphics[width=\ww,frame]{figures/ablation_study_augmentations/assets/pred3_aug.jpg} \\

        \rotatebox{90}{\phantom{AAA} ``huge avocado''}
        \includegraphics[width=\ww,frame]{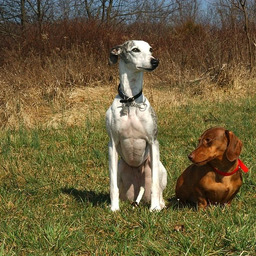} &
        \includegraphics[width=\ww,frame]{figures/ablation_study_augmentations/assets/mask_overlay4.jpg} &
        \includegraphics[width=\ww,frame]{figures/ablation_study_augmentations/assets/pred4_no_aug.jpg} &
        \includegraphics[width=\ww,frame]{figures/ablation_study_augmentations/assets/pred4_aug.jpg} \\

        \rotatebox{90}{\phantom{AAA} ``knit beanie''}
        \includegraphics[width=\ww,frame]{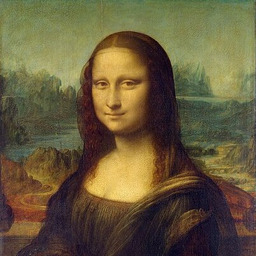} &
        \includegraphics[width=\ww,frame]{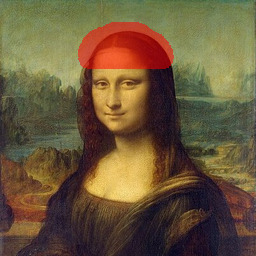} &
        \includegraphics[width=\ww,frame]{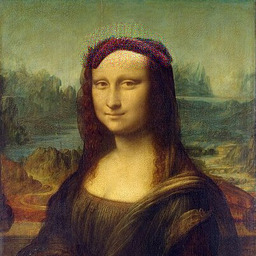} &
        \includegraphics[width=\ww,frame]{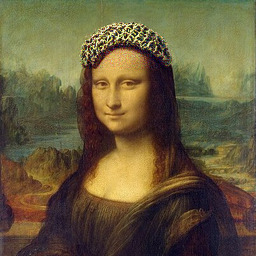} \\

        \rotatebox{90}{\phantom{AAA}``burning tree''}
        \includegraphics[width=\ww,frame]{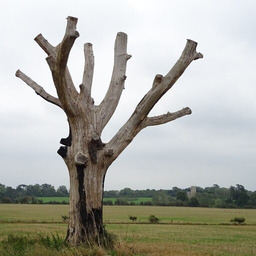} &
        \includegraphics[width=\ww,frame]{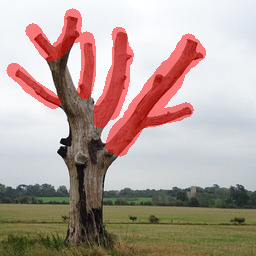} &
        \includegraphics[width=\ww,frame]{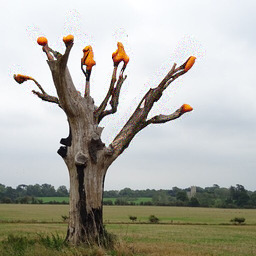} &
        \includegraphics[width=\ww,frame]{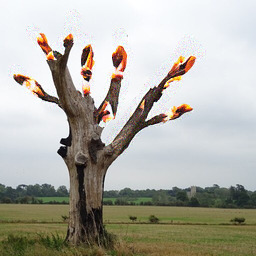} \\

        Input image &
        Input image + mask &
        (1) &
        (2)
    \end{tabular}
    
    \caption{\textbf{Extending augmentations ablation:} In order to assess the importance of the extending augmentation technique, we used the same random seeds for the same inputs to ensure that the results would differ in the use of augmentations. As we can see, (2) using extending augmentations makes the resulting images more natural and coherent with the background in comparison to (1) not using extending augmentations.}
    \label{fig:ablation_study_augmentations}
\end{figure*}

As mentioned in
\Cref{sec:ablation_study}
we performed an ablation study for the extending augmentations. \Cref{fig:ablation_study_augmentations} demonstrates the importance of the augmentations: the same random seed is used in two runs, one with and the other without augmentations. We can see that the images generated with the use of augmentations are more visually plausible and are more coherent than the ones generated without the augmentations. (This is an extended version of
\Cref{fig:ablation_study_augmentations_mini} from the main paper.)

\subsection{Inference Time}
\label{sec:inference_time}
We report synthesis time for a single image using one NVIDIA A10 GPU:
\begin{itemize}
    \item Our method (\Cref{alg:final})
          \& Local CLIP-guided diffusion (\Cref{alg:local_clip_guided_diffusion}): 27 seconds.
    \item $\textit{PaintByWord++}$: 78 seconds.
\end{itemize}

Original paint by word \cite{bau2021paint} did not release their code and did not mention the run-time.

In practice, as described in \Cref{sec:generation_ranking},
we generate several results for the same inputs and use the best ones. Instead of generating them sequentially, we accelerate the generation process using two techniques:
\begin{enumerate}
    \item \textbf{Batch generation:} Instead of generating a single image in each diffusion pass, we multiplied the input several times and generated several instances on the same pass. Because of the stochasticity of the diffusion process, each result is different.
    \item \textbf{Parallel generation:} Because each of the generation processes is independent, we can distribute the generation across multiple GPUs. In our experiments, we concurrently used 4 NVIDIA A10 GPUs.
\end{enumerate}
Using the above accelerations, we generate 64 synthesis results in about 6 minutes --- less than 6 seconds per image.

\subsection{Comparison with Baselines}
\paragraph{$\textit{PaintByWord}$} Because the models and code that was used by Bau et al.~\cite{bau2021paint} are currently unavailable, we used as input the images and masks extracted from their paper.

\paragraph{$\textit{PaintByWord++}$} We adapted the VQGAN+CLIP \cite{vqgan_clip} implementation to support masks using the same $\mathcal{D}_{\textit{CLIP}}$ loss from
\Cref{eqn:d_clip}.
We used the VQGAN \cite{esser2021taming} model that was trained on ImageNet with reduction factor $f=16$.
For the latent optimization, we used the Adam optimizer with a learning rate of 0.1 for 500 steps. We found that constraining the optimization of the latent space $z$ only to the corresponding mask area, the same way it was done by Bau et al.~\cite{bau2021paint}, improved the background preservation.

\subsection{Implementation Details for Applications}
In this section, we provide the implementation details for scribble-guided editing and text-guided image extrapolation applications.

\subsubsection{Scribble-guided editing}
In order to create the results that are demonstrated in
\Cref{fig:scribble_editing_monkeys_blanket} of the main paper,
the user first scribbles on the input image, then masks the scribble area (the masking can also be done automatically by taking the scribbles area and dilating it by morphological operations), then provides a text prompt and uses the same algorithm as for object altering.

\begin{figure*}[t]
    \centering
    \setlength{\tabcolsep}{1pt}
    \renewcommand{\arraystretch}{2}
    \setlength{\ww}{0.45\columnwidth}

    \begin{tabular}{ccc}
        Original image &
        Input scribble &
        Mask
        \\

        \includegraphics[width=\ww,frame]{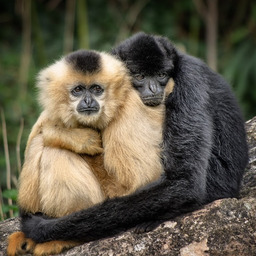} &
        \includegraphics[width=\ww,frame]{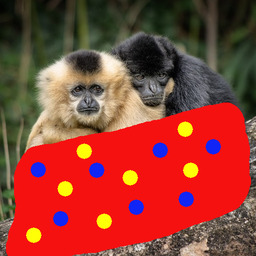} &
        \includegraphics[width=\ww,frame]{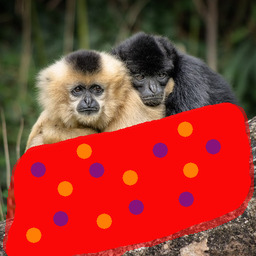}
        \\
    \end{tabular}
  
    \renewcommand{\arraystretch}{0.5}
    \begin{tabular}{cccc}
        \rotatebox{90}{\phantom{A}Diffusion steps $k = 20$}
        \includegraphics[width=\ww,frame]{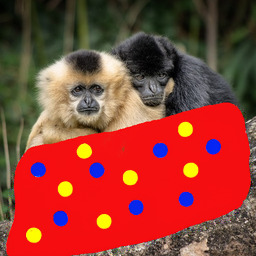} &
        \includegraphics[width=\ww,frame]{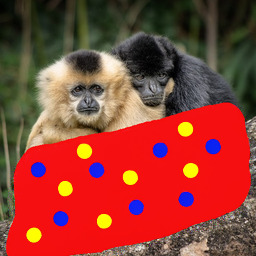} &
        \includegraphics[width=\ww,frame]{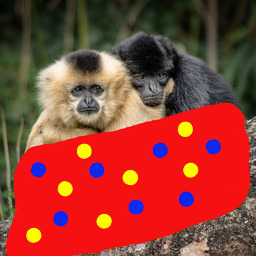} &
        \includegraphics[width=\ww,frame]{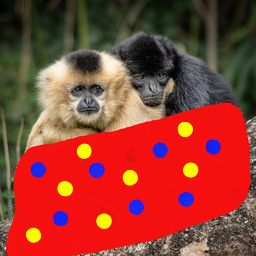}
        \\

        \rotatebox{90}{\phantom{A}Diffusion steps $k = 40$}
        \includegraphics[width=\ww,frame]{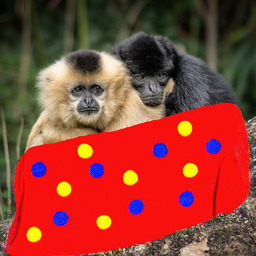} &
        \includegraphics[width=\ww,frame]{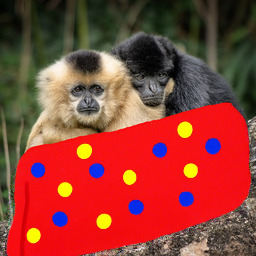} &
        \includegraphics[width=\ww,frame]{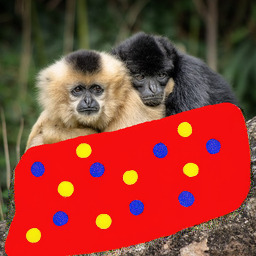} &
        \includegraphics[width=\ww,frame]{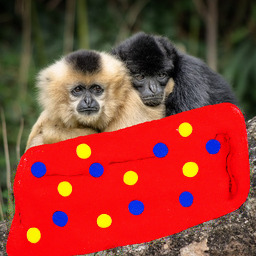}
        \\

        \rotatebox{90}{\phantom{A}Diffusion steps $k = 60$}
        \includegraphics[width=\ww,frame]{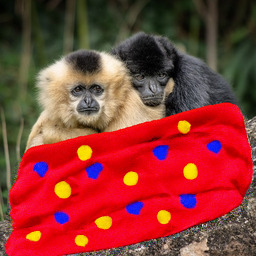} &
        \includegraphics[width=\ww,frame]{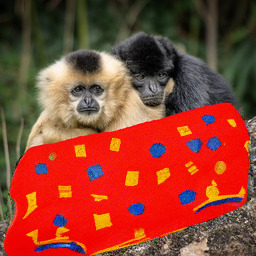} &
        \includegraphics[width=\ww,frame]{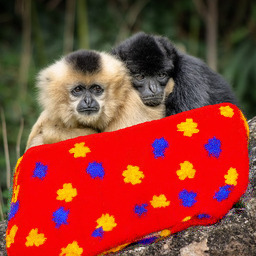} &
        \includegraphics[width=\ww,frame]{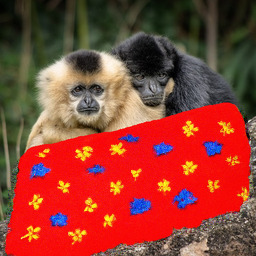}
        \\

        \rotatebox{90}{\phantom{A}Diffusion steps $k = 80$}
        \includegraphics[width=\ww,frame]{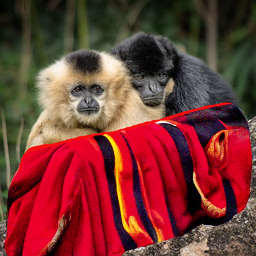} &
        \includegraphics[width=\ww,frame]{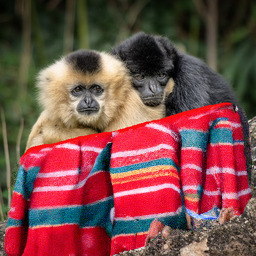} &
        \includegraphics[width=\ww,frame]{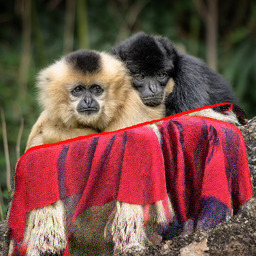} &
        \includegraphics[width=\ww,frame]{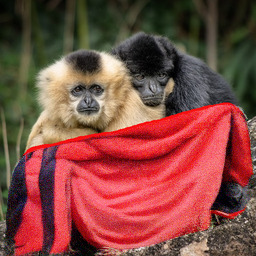}
        \\

        Result 1 &
        Result 2 &
        Result 3 &
        Result 4
        \\

    \end{tabular}
    
    \caption{\textbf{Scribble-guided editing diffusion steps effect:} when the diffusion steps are large (e.g. $k=80$), the resulting images are more realistic and diverse but do not preserve the colors of the input scribble, on the other hand, when the diffusions steps are low (e.g. $k=20$), the resulting images are almost identical to the input scribble.}
    \label{fig:scribble_editing_diffusion_steps_effect}
\end{figure*}

An important hyper-parameter for this application is the number of target diffusion steps $k$ in
\Cref{alg:final}.
\Cref{fig:scribble_editing_diffusion_steps_effect} demonstrates the effect of changing this parameter: when diffusing for a longer period (e.g., 80 diffusion steps out of 100), only the main red color of the blanket is kept, the blanket shading is more realistic, and the results are more diverse. When diffusing for a shorter period (e.g., 20 diffusion steps out of 100), the scribble is hardly modified.

\subsubsection{Text-guided image extrapolation}
\label{sec:tex-guided-extrapolation-details}
\begin{figure*}[p]
    \centering
    \setlength{\tabcolsep}{2pt}
    \renewcommand{\arraystretch}{0.5}
    \setlength{\ww}{0.5\columnwidth}

    \begin{tabular}{cccc}
        \includegraphics[width=\ww,frame]{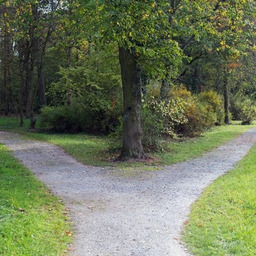}       &
        \includegraphics[width=\ww,frame]{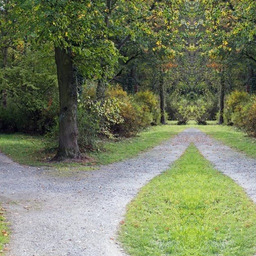} &
        \includegraphics[width=\ww,frame]{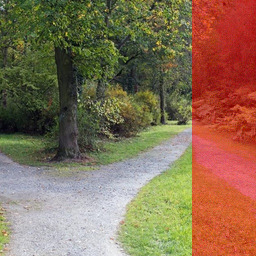}      &
        \includegraphics[width=\ww,frame]{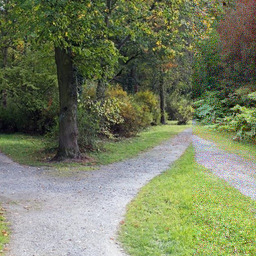}           \\

        (1) Source image &
        (2) Translated and reflected &
        (3) Mask &
        (4) 1st prediction \\

        \includegraphics[width=\ww,frame]{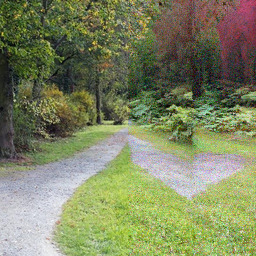}         &
        \includegraphics[width=\ww,frame]{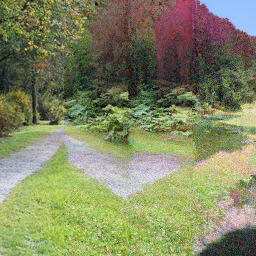}         &
        \includegraphics[width=\ww,frame]{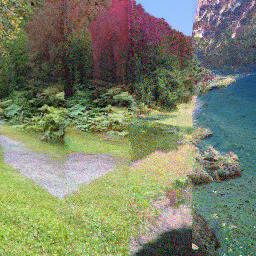}         &
        \includegraphics[width=\ww,frame]{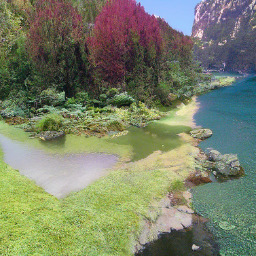}  \\

        (5) 2nd prediction &
        (6) 3rd prediction &
        (7) 4th prediction &
        (8) Denoised prediction
    \end{tabular}

    \caption{\textbf{Text-guided image extrapolation:} We aim to extrapolate the source image (1) to the right according to the guiding text ``heaven''. We start by (2) translating the image to the left by $\frac{1}{4}$ of the input resolution, and filling the missing area with reflection padding. Then we mask the new area (3) and predict the missing part (4) using the foreground altering algorithm. We perform this process 3 more times (5-7) to get a noisy prediction (7). In order to denoise it, we do the same process with a mask that covers the entire image and get the denoised result (8) that we can stitch to the source image. Notice that we can reach an arbitrary resolution using this method.}
    \label{fig:image_extrapolation_heaven_hell_explanation}
\end{figure*}

In order to extend the image beyond its original resolution, we gradually predict the unknown parts of the image in a sequential manner. \Cref{fig:image_extrapolation_heaven_hell_explanation} demonstrates the building process: at each stage, (2) we translate the image $\frac{1}{4}$ to the opposite of the desired direction and fill the missing area using standard reflection padding, (4) then we inpaint the new area guided by the text description, using the regular algorithm for foreground editing. (5-7) We repeat the process 3 times until we have a new image. The new image is still a bit noisy --- due of the gradual inpainting, each synthesis result is noisier than the previous one because of the chaining of the natural image statistics. In order to mitigate it, (8) we denoise this image using the diffusion process again. We repeat the same process in the other direction. Our output can have an arbitrarily large image resolution.

We also notice that gradual diffusion steps are beneficial: we diffuse the first quarter for a small number of diffusion steps, and then in each step, we enlarge the number of diffusion steps.

\subsection{Ranking Implementation Details}
\begin{figure*}[p]
    \centering
    \setlength{\tabcolsep}{0.5pt}
    \renewcommand{\arraystretch}{0.5}
    \setlength{\ww}{0.33\columnwidth}
  
    \begin{tabular}{cccccc}
        \rotatebox{90}{\phantom{Aa}``pile of books''}
        \includegraphics[width=\ww,frame]{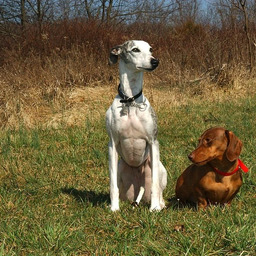} &
        \includegraphics[width=\ww,frame]{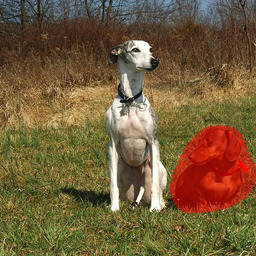} &
        \includegraphics[width=\ww,frame]{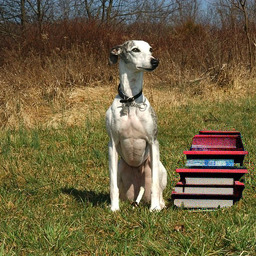} &
        \includegraphics[width=\ww,frame]{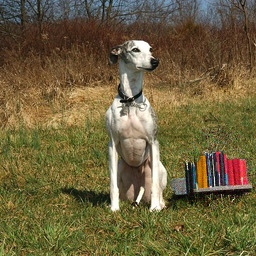} &
        \includegraphics[width=\ww,frame]{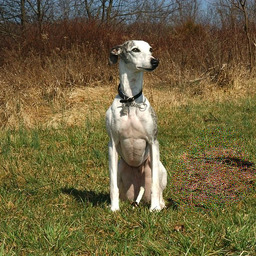} &
        \includegraphics[width=\ww,frame]{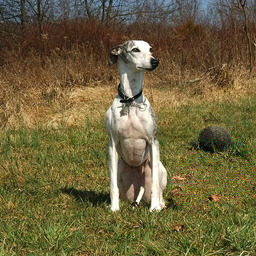}
        \\

        \rotatebox{90}{\phantom{AA}``car tire''}
        \includegraphics[width=\ww,frame]{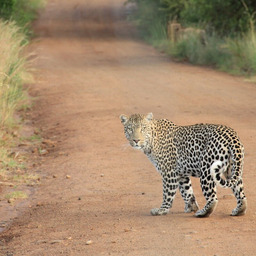} &
        \includegraphics[width=\ww,frame]{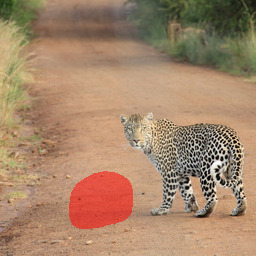} &
        \includegraphics[width=\ww,frame]{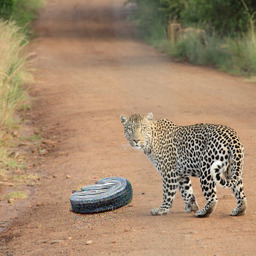} &
        \includegraphics[width=\ww,frame]{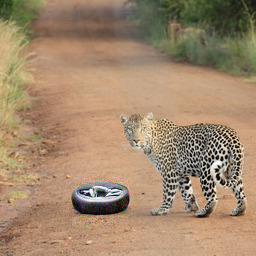} &
        \includegraphics[width=\ww,frame]{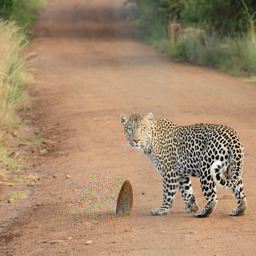} &
        \includegraphics[width=\ww,frame]{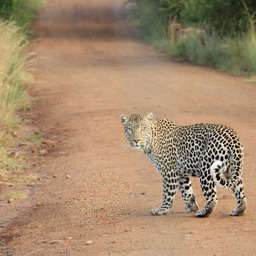}
        \\

        \rotatebox{90}{\phantom{AAA}``sunrise''}
        \includegraphics[width=\ww,frame]{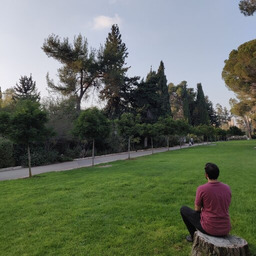} &
        \includegraphics[width=\ww,frame]{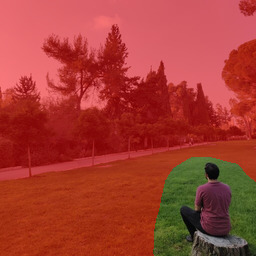} &
        \includegraphics[width=\ww,frame]{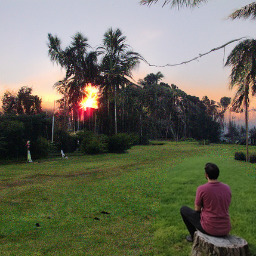} &
        \includegraphics[width=\ww,frame]{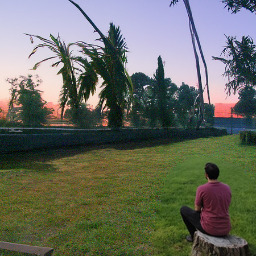} &
        \includegraphics[width=\ww,frame]{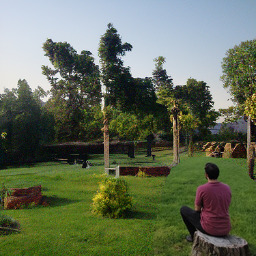} &
        \includegraphics[width=\ww,frame]{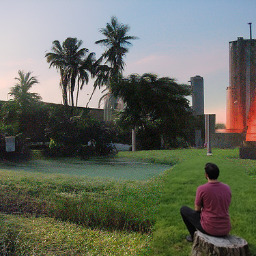}
        \\

        \rotatebox{90}{\phantom{AAA}``gravel''}
        \includegraphics[width=\ww,frame]{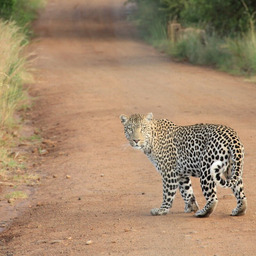} &
        \includegraphics[width=\ww,frame]{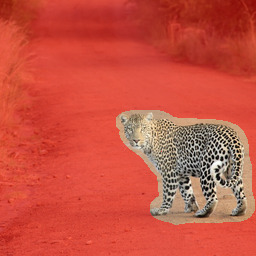} &
        \includegraphics[width=\ww,frame]{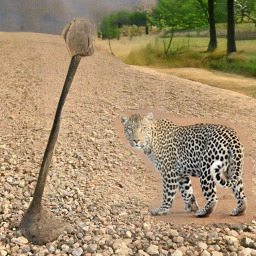} &
        \includegraphics[width=\ww,frame]{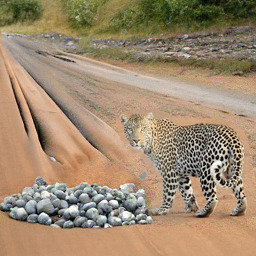} &
        \includegraphics[width=\ww,frame]{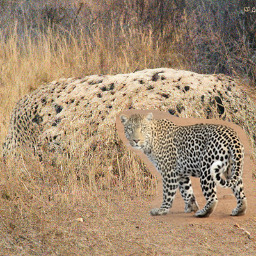} &
        \includegraphics[width=\ww,frame]{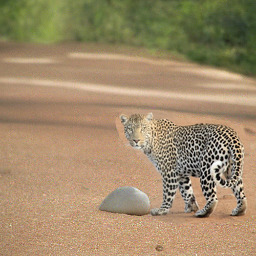}
        \\
        
        Input image &
        Input mask &
        1st ranked result &
        2nd ranked result &
        63th ranked result &
        64th ranked result
        \\
    \end{tabular}
    
    \caption{\textbf{Ranking algorithm effectiveness:} We generate 64 synthesis results and rank them using CLIP. We found that this method only roughly ranks the results: the top 20\% are consistently better than the bottom 20\%, but in the resolution of a single image, this is not the case --- the first result isn't always better than the second one.}
    \label{fig:ranking_algorithm}
\end{figure*}

We utilized the ranking algorithm that is explained in
\Cref{sec:generation_ranking} in the main paper
using 64 synthesis results. As described in
\Cref{sec:limitations} in the main paper,
the ranking is not perfect because it takes into account only the generated area. In addition, the ranking is not accurate enough in the resolution of single images: the top-ranked image isn't always better than the second one, etc. Nonetheless, the top 20\% of the images are almost always better than the bottom 20\%. In practice, we generate 64 results and choose manually from the top 10 images ordered by their ranking (in both the baselines and our method). \Cref{fig:ranking_algorithm} demonstrates the effectiveness of the ranking algorithm.

\section{User Study}
In order to evaluate our model quantitatively, we conducted a user study. The only results of the Paint By Word model on general images (albeit GAN-generated) that were available are the ones in their paper. Hence, we chose to conduct the user study on these images (along with their corresponding masks). The study was conducted on 35 participants.

The participants were shown each time the inputs to the model (image, mask and text description) along with the model prediction, and were asked to rate the prediction, on a scale of 1--5, for one of the following criteria:
\begin{enumerate}
    \item The overall realism of the prediction.
    \item The amount of background preservation of the prediction in the unedited area.
    \item The correspondence of the edited image to the guiding text description.
\end{enumerate}

The questions were randomly ordered, and the participant had the ability to go back and edit their previous ratings until submission.

Mean user study scores are presented in Table 1 of the main paper. 
The difference between conditions is statistically significant (Kruskal-Wallis test, $p<10^{-130}$). 
Further analysis using Tukey's honestly significant difference procedure \cite{tukey1949comparing} shows that the improvement shown by our method is statistically significant vs.~all other conditions (\Cref{tab:user_study_stats}).

\begin{table*}
\begin{center}
 \begin{tabular}{ccccc}
 \toprule
 
 \textbf{Method 1} & 
 \textbf{Method 2} & 
 Realism &
 Background preservation &
 Text match \\
 
 & & p-value & p-value & p-value \\
 
 \midrule
 
 Local CLIP GD \cite{clip_guided_diffusion} &   Ours &                                                      \textbf{0.003}  & \textbf{<0.001} & \textbf{<0.001} \\
 Local CLIP GD \cite{clip_guided_diffusion} &   \textit{PaintByWord} \cite{bau2021paint} &                  0.435           & 0.578           & \textbf{<0.001} \\
 Local CLIP GD \cite{clip_guided_diffusion} &   \textit{PaintByWord++} \cite{vqgan_clip, bau2021paint} &    \textbf{<0.001} & 0.106           & \textbf{<0.001} \\
 Ours &                                         \textit{PaintByWord} \cite{bau2021paint} &                  \textbf{<0.001} & \textbf{<0.001} & \textbf{<0.001} \\
 Ours &                                         \textit{PaintByWord++} \cite{vqgan_clip, bau2021paint} &    \textbf{<0.001} & \textbf{<0.001} & \textbf{<0.001} \\
 \textit{PaintByWord} \cite{bau2021paint} &     \textit{PaintByWord++} \cite{vqgan_clip, bau2021paint} &    \textbf{<0.001} & 0.719           & 0.704 \\
 
 \bottomrule
\end{tabular}
\caption{\textbf{User study statistical analysis:} We use Tukey’s honestly significant difference procedure \cite{tukey1949comparing} to test whether the differences between mean scores in our user study are statistically significant. Significant results in bold. Our results are statistically better than all other methods on all the measured conditions.}
\label{tab:user_study_stats}
\end{center}
\end{table*}

\end{document}